\documentclass[10pt,twocolumn,letterpaper]{article}
\usepackage[accsupp]{axessibility}  %

\usepackage{iccv}              %

\usepackage{stfloats}
\usepackage{multirow}
\usepackage{cuted}
\usepackage{capt-of}
\usepackage{float}%
\usepackage{caption}%
\usepackage{graphicx}

\definecolor{iccvblue}{rgb}{0.21,0.49,0.74}
\usepackage[pagebackref,breaklinks,colorlinks,allcolors=iccvblue]{hyperref}

\def\paperID{6861} %
\def\confName{ICCV}
\def\confYear{2025}

\title{BokehDiff: Neural Lens Blur with One-Step Diffusion}

\author{Chengxuan~Zhu$^{1,2\dagger}$$\,\,$ Qingnan~Fan$^{2\ast}$$\,\,$Qi~Zhang$^{2}$$\,\,$Jinwei~Chen$^{2}$$\,\,$Huaqi~Zhang$^{2}$$\,\,$Chao~Xu$^{1}$$\,\,$Boxin~Shi$^{3,4\ast}$\\
\small{$^{1}$National Key Lab of General AI, School of Intelligence Science and Technology, Peking University}\\
\small{$^{2}$Vivo Mobile Communication Co., Ltd.}\\
\small{$^{3}$State Key Laboratory of Multimedia Information Processing, School of Computer Science, Peking University}\\
\small{$^{4}$National Engineering Research Center of Visual Technology, School of Computer Science, Peking University}\\
{\tt\small \{peterzhu,shiboxin\}@pku.edu.cn, qingnanfan@vivo.com} \\
{\small\href{https://github.com/FreeButUselessSoul/bokehdiff}{\url{https://github.com/FreeButUselessSoul/bokehdiff}}}
}

\begin{document}
\maketitle
\def\thefootnote{$\dagger$}{\footnotetext{Work done during an internship at Vivo.}} %
\def\thefootnote{$\ast$}{\footnotetext{Corresponding authors.}}
\begin{strip}
    \centering
    \vspace{-6em}
    \includegraphics[width=\linewidth]{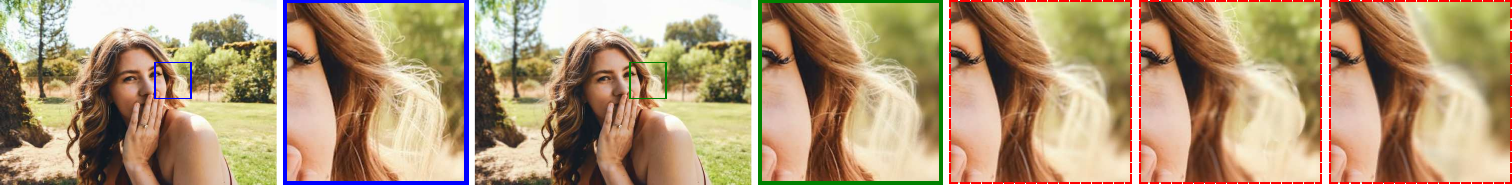}
    \includegraphics[width=\linewidth]{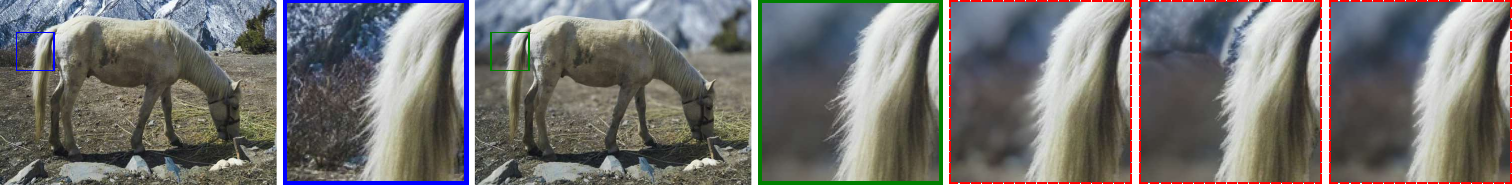}
    \includegraphics[width=\linewidth]{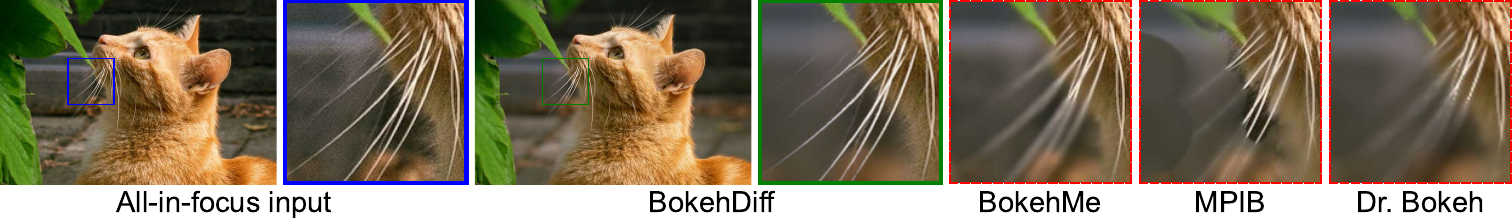}
    \vspace{-2em}
    \captionof{figure}{BokehDiff bridges the gap between physics and diffusion priors, and is able to synthesize photorealistic lens blur effects even when inaccurate depth estimation causes previous methods (BokehMe~\cite{peng2022bokehme}, MPIB~\cite{peng2022mpib}, and Dr. Bokeh~\cite{sheng2024dr}) to fail, especially at the depth discontinuities. The examples show previous methods over-blur the horse's tail, the person's hair, and the whiskers of the cat.}
    \label{fig:teaser}
\end{strip}

\begin{abstract}
    We introduce \textbf{BokehDiff}, a novel lens blur rendering method that achieves physically accurate and visually appealing outcomes, with the help of generative diffusion prior. Previous methods are bounded by the accuracy of depth estimation, generating artifacts in depth discontinuities. Our method employs a physics-inspired self-attention module that aligns with the image formation process, incorporating depth-dependent circle of confusion constraint and self-occlusion effects. We adapt the diffusion model to the one-step inference scheme without introducing additional noise, and achieve results of high quality and fidelity. To address the lack of scalable paired data, we propose to synthesize photorealistic foregrounds with transparency with diffusion models, balancing authenticity and scene diversity.
    \vspace{-4em}
\end{abstract}

\section{Introduction}
\label{sec:intro}
The bokeh effect is the out-of-focus blurriness observed in photos, physically caused by using a lens with a large aperture, and is often used in portrait photography to emphasize the subject. Due to the cost of large aperture lenses, bokeh rendering has become a hot topic in the computational photography community. Previous works~\cite{peng2022mpib,peng2022bokehme,deeplens2018,zhang2019synthetic} mostly aim to simulate the blurriness accurately with a pixel-level accurate depth estimation. However, since depth prediction tends to fail on edges and intricate details, artifacts can often be observed on structures such as people's hair and animals' fur, as shown in \cref{fig:teaser}.
As state-of-the-art diffusion models (\eg, SDXL~\cite{podell2023sdxl}) are already capable of generating photorealistic lens blur effects from text instructions~\cite{yuan2025generative}, \textit{can they be applied to render lens blur effects from a given image?}

The answer is frustrating, primarily due to diffusion models' inherent tendency to alter the content of the input image. The problem traces down to the iterative denoising process of diffusion methods, where the input image is injected into the model to guide the denoising process. 
The original noise introduces much uncertainty and tends to break the original structure of the input image. The denoising process is also too time-consuming to serve as a lens blur rendering tool, making the rich generative priors difficult to exploit.
BokehDiff proposes to denoise the input image with \underline{only one denoising step}, without adding any noise. It simply treats the all-in-focus image as the combination of the image with lens blur and unknown noise that needs to be removed. The noise prediction network is finetuned to learn the noise for transformation, and acquires the image lens blur with only one forward pass. BokehDiff effectively preserves the structures since no noise is added.

Another problem of diffusion models lies in the design of self-attention module. To emphasize more important features, self-attention may discard less important ones, even contradicting the underlying physics.
It performs well in tasks like inpainting~\cite{li2022mat,avrahami2022blended,avrahami2023blended} and image super resolution~\cite{osediff,lu2022transformer,zamir2022restormer,chen2023activating}, where adjacent pixels are not influenced by each other. But for the task of lens blur rendering where the blur is aggregated from neighboring pixels, it is difficult for self-attention to control the results, due to the global receptive field and the neglect of unimportant pixels.
The proposed BokehDiff, features a \emph{physics-inspired self-attention} (PISA) module that is designed to \underline{immitate the physics in the image formation process}. For the light sources in an image, the PISA module normalizes their contribution in an energy-conserved way, limits their impact by a physics-based \emph{circle-of-confusion} (CoC) term, and mask the self-occlusion in light propagation.

For learning-based methods, the scarcity of high-quality paired data also poses a problem. Real-world paired data~\cite{ignatov2020aim, Mandl2024NeuralBokeh} tend to suffer from misalignment caused by motion, lens breathing, or different exposure, with examples shown in \cref{fig:sampleEBB}.
As for synthetic data, applying 3D engines to render bokeh from user-defined assets is constrained by the numbers of available scenes~\cite{peng2022bokehme,Mandl2024NeuralBokeh}, and the CG rendering differs from the reality.
Another trend is to perform ray-tracing from several image layers~\cite{zhang2019synthetic,peng2022mpib,peng2022bokehme,selective2023peng,yang2023bokehornot,sheng2024dr}, but the imperfect matting contents make the final rendered results look fake, especially for intricate structures such as hair and hands.
BokehDiff proposes a data synthesis paradigm to synthesize \underline{paired and aligned high-quality data} for training and evaluation, by exploiting an off-the-shelf text-to-image model to synthesize photorealistic foreground with transparency~\cite{zhang2024transparent} instead of segmenting the foreground from photos and build a synthetic dataset for training and testing.

We propose the first neural lens blur rendering pipeline based on pretrained diffusion priors, outperforming previous works in error-prone depth discontinuous areas. The contributions are summarized as follows:
\begin{itemize}
    \item a physics-inspired self-attention module that follows the image formation model, considering the energy conservation laws, circle of confusion, and self-occlusion;
    \item an efficient one-step inference scheme with diffusion models, exploiting the generative priors;
    \item a new scalable data synthesis paradigm as well as a curated dataset for bokeh rendering, which solves the dilemma of ground-truth accuracy and scene diversity.
\end{itemize}

\section{Related Works}
\label{sec:related}
\subsection{Bokeh Rendering}
As a common technique in photography and 3D rendering, lens blur is caused by the wide aperture of the camera. Mathematically, it equals the weighted sum of views from the neighborhood of the principle point~\cite{potmesil1981lens,lee2008real,yan2015fast}. 

However, real-world cases lack the complete 3D model or multiple view input. With the image as the only input, researchers face two main challenges, namely the missing information about the hidden surface and the inaccurate depth estimation. For the first problem, classical rendering extrapolate visible pixels to occluded ones~\cite{kraus2007depth,lee2010real,lee2008real} or performs inpainting~\cite{vaidyanathan2015layered,busam2019sterefo,peng2022mpib,sheng2024dr} to hallucinate the missing information. Either way, however, requires segmenting the scene into multiple planes, which is error-prone on depth discontinuous regions. Though efforts have been made to make the operation smooth~\cite{peng2022mpib,busam2019sterefo} or differentiable~\cite{sheng2024dr}, they are outperformed by neural rendering methods when handling scenes with complex geometry.

Neural rendering uses a neural network to mimic the image formation model, and is often trained end-to-end on synthetic data with ground truth depth map~\cite{peng2022bokehme,xiao2018deepfocus,deeplens2018,ignatov2020rendering,qian2020bggan,dutta2021stacked,yang2023bokehornot,Mandl2024NeuralBokeh}. As the network learns to add specific amount of blur to the input image, the problem of inaccurate depth estimation constitutes the major bottleneck of performance, as seen in \cref{fig:teaser}.

In this paper, we endow the diffusion priors to bokeh rendering, and significantly improves photorealism in regions where depth prediction methods fail.

\subsection{Image Editing with Diffusion Models}
As a powerful tool for image generation, diffusion models~\cite{song2020ddim,ho2020ddpm} have caught much attention in the community, especially about the possibility exploiting the diffusion priors for controllable generation~\cite{zhang2023adding,yang2024depth,cao2023masactrl,podell2023sdxl} and editing~\cite{brooks2023instructpix2pix,avrahami2022blended,avrahami2023blended,xu2024inversion,jiang2024scedit,meng2021sdedit,cao2023masactrl,epstein2023diffusion}. However, the stochastic nature of adding and removing noise makes it difficult for previous diffusion models to retain the original structure. Some guide the generation with the original latent map~\cite{epstein2023diffusion,xu2024inversion,meng2021sdedit,cao2023masactrl,lin2024ctrl} or information injection~\cite{yang2023pixel,zhang2023adding}, but cannot preserve the pixel-wise structure; Others propose to blend the edited part with the original image~\cite{avrahami2022blended,avrahami2023blended}, but is limited to image inpainting task where only a small part should change.

Recently, researchers have found it more structure-preserving if some initial denoising steps, when the coarse structure is hallucinated from noise, are truncated~\cite{sun2401improving}. 
The diffusion process can even be removed completely~\cite{osediff}, with the low-quality image being the input to be denoised. The multi-step is also found redundent as it introduces accumulating error~\cite{he2024lotus}.

Taking the idea one step further, we use the all-in-focus input image as input, without adding noise to it, and adopt the efficient inference scheme of one-step denoise for the task of neural lens blur rendering.

\begin{figure*}[t]
    \centering
    \includegraphics[width=\linewidth]{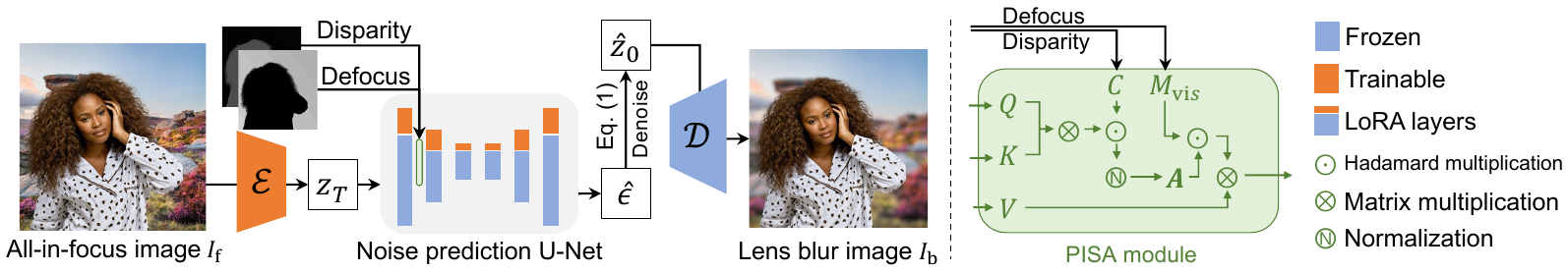}
    \vspace{-1.5em}
    \caption{The framework of the proposed method. Given a paired synthetic data with disparity map, we optimize a LoRA of the U-Net and the encoder $\mathcal{E}$, while the decoder $\mathcal{D}$ remains frozen. A tailored PISA module (colored in green) is applied during downsampling, and is detailed in the right column, which is introduced in \cref{sec:selfattention}. }
    \label{fig:pipeline}
    \vspace{-.5em}
\end{figure*}

\section{Method}
The task of lens blur rendering takes an all-in-focus image $I_\text{f}$ as input, and blurs it with respect to the disparity map $d$ and focus disparity $d_\text{f}$. The goal is to acquire the image with the correct lens blur $I_\text{b}$.
Classical rendering methods apply a physics-based image formation model, such as the one illustrated in \cref{fig:pisa_illu}, while neural rendering methods learn the mapping from $I_\text{f}$ to $I_\text{b}$ directly. We aim to imitate the image formation model in the diffusion model, and prove that diffusion models can be adapted for the task.
We first introduce the one-step diffusion framework in \cref{sec:osdiff}, and then detail the PISA module in \cref{sec:selfattention}, designed to make the diffusion model aware of the physics-related constraints. The framework of the proposed method is shown in \cref{fig:pipeline}. We then describe the data synthesis paradigm in \cref{sec:our_dataset} and the supervision in \cref{sec:supervision}.

\subsection{One-Step Diffusion for Bokeh Rendering}\label{sec:osdiff}
To save memory, diffusion models perform on latent space nowadays~\cite{ho2020ddpm}, with a pretrained encoder $\mathcal{E}$ to compress the images into latents and another decoder $\mathcal{D}$ to revert latents back to image space. 
Given a noisy latent $z_t$ at timestep $t$, the denoised latent $\hat{z}_0$ is estimated by
\begin{equation}
    \hat{z}_0 = \frac{z_t - \beta_t \cdot  \epsilon_\theta(z_t; c_\text{txt})}{\alpha_t},
\label{eq:diffusion}
\end{equation}
where $c_\text{txt}$ is the encoded text embedding as condition, and $\epsilon_\theta$ stands for the U-Net predicting the noise. It's worth noting that the nature of \cref{eq:diffusion} is only a transformation from $z_t$ to $\hat{z}_0$, and the generative priors lie in the seeming omnipotence of $\epsilon_\theta$, as it is pre-trained on massive amount of data for noise prediction. In this sense, the pretraining of latent diffusion models is to map the Gaussian distribution into a desired output distribution. 
To exploit the rich generative priors of diffusion models, we base our generation on finetuning an off-the-shelf SDXL~\cite{podell2023sdxl} text-to-image model. 

While diffusion models are originally trained to remove noise, it is recently found that diffusion models can also be trained to invert other imposed degradation such as blurring, masking, or downsampling~\cite{bansal2024cold}. This motivates us to take one step further and ponder the possibility of implicitly defining the image quality as the amount of lens blur effect, and learn the transformation from $I_\text{f}$ to $I_\text{b}$ with physically correct lens blur effects.

As found by previous works, diffusion models tend to perform better in the timesteps they are trained on~\cite{hang2023minsnr}, implying the possibility of a one-step inference diffusion model~\cite{parmar2024img2imgturbo,osediff}, which is trained on that particular timestep. In this paper, as the all-in-focus image is already close to the target domain, we fix the timestep as $T=499$, and fintunes the LoRA~\cite{hu2021lora} of the U-Net and the encoder $\mathcal{E}$, to fit the altered latent distribution.

\subsection{Physics-Inspired Self-Attention Module} \label{sec:selfattention}

\begin{figure}[t]
    \centering
    \vspace{-1.5em}
    \includegraphics[width=\linewidth]{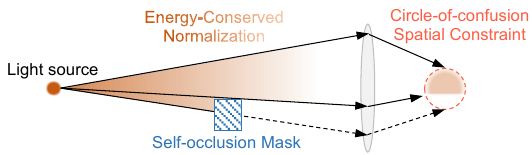}
    \vspace{-2em}
    \caption{An illustration of the image formation model, and the three physics-related aspects considered in the PISA module.}
    \label{fig:pisa_illu}
    \vspace{-1em}
\end{figure}

The Achilles's heel of applying the noise prediction network for neural lens rendering lies in the self-attention module, because it is ignorant of the 3D formulation of lens blur. 
We design the PISA module that follows the three physics-related aspects as illustrated in \cref{fig:pisa_illu}.

\noindent\textbf{Energy-Conserved Normalization.}
In the vanilla self-attention formulation, the output equals the product of the value vectors and the normalized similarity between query vectors and key vectors, namely
\begin{equation}
\!\!\text{Attn}(Q,K,V) = \textbf{A}^\text{(K)} V, \text{where }\textbf{A}^\text{(K)}_{qk} = \frac{\exp(A_{qk})}{\sum_s \exp(A_{qs})}.
\label{eq:prev_attn_formulation}
\end{equation}
Here $Q$, $K$, and $V$ represent the query, key and value matrix, with each row representing a point in the latent map. $A=d_\text{key}^{-\frac{1}{2}}QK^\top$ is the similarity matrix, and $d_\text{key}$ is the size of the key matrix. For convenience of notation, we use the subscript $q$ and $k$ to refer to the row of query point and the column of key point. $d$ is the number of pixels, placed in the denominator for numerical stability. As previous works suggest, the normalized similarity $\textbf{A}^\text{(K)}$ between $Q$ and $K$ contains the structural information~\cite{hussain2022global}, while $V$ possess the appearance information in the context of vision tasks. Thus, the result is a structurally weighted sum of appearance. 
Note that the normalization operation, $\text{Softmax}(\cdot)$ is applied on the channel of key, which guarantees that each row in the output is a normalized linear combination of the rows in $V$, with the weights summed up as $1$. Since the latent pixels corresponds to the image pixels spatially, and $V_{k}$ stands for the appearance feature at pixel $k$, the contribution of pixel $k$ towards pixel $q$ in the attention output can be measured by $\textbf{A}_{qk}$. In most cases, the formulation enables neural networks to focus on important appearance features, without the concern that the rows in $V$ can contribute very differently to the output. 

We first redesign the normalization scheme, so that the energy of light does not increase or diminish as it spreads to neighboring pixels. As self-attention is originally designed to emphasize important features while discarding trivial ones, the total contribution of any given row $V_k$ towards the output matrix varies drastically. Based on the physical inspiration, we propose to modify the softmax operation to normalize on query dimension, simply by
\begin{equation}
    \textbf{A}^\text{(Q)}_{qk}=\frac{\exp(A_{qk}) }{\sum_s \exp(A_{sk})},
\label{eq:normalize_Q}
\end{equation}
in which the energy conservative law holds, and the total contribution from any row in $V$ to the output matrix is $1$, with $\sum_i \textbf{A}_{ik}^\text{(Q)}=1$.

\noindent\textbf{Circle-of-Confusion Spatial Constraint.} For a light source $k$ that is off the focal plane, the CoC is formed on the camera sensor. Its radius $r_c(k)$ describes the extent of blurriness, and is proportionate to the disparity difference between the point and the focal plane~\cite{lee2008real,lee2010real}, given by
\begin{equation}
    r_c(k) = |d_f - \text{dis}(P_k)|\cdot A,
\label{eq:coc_radius}
\end{equation}
 where $A$ is the camera parameters of the aperture diameter, and $d_f$ is the disparity of the focal plane.
$k$ is any point light source in the context of self-attention, while $P_k$ is the pixel location of point light source $k$. Let $\text{dis}(P_k)$ denote the disparity (the reciprocal of depth) of $k$, shortened as $d_k$ for convenience. For a practical application as lens blur rendering, the lens can be assumed as thin lens model~\cite{lee2008real,kraus2007depth,sheng2024dr}, and thus \cref{eq:coc_radius} holds. In practice, $r_c(k)$ marks the theoretical limit of how far $k$ can influence, by casting a cone of light through space. Without the spatial constraint, every feature can have an unlimited global effect, making it difficult for the network to neglect irrelevant pixels.

To consider the spatial constraint into the self-attention design, we propose to mask it at the softmax module. In this way, the conservation of energy still holds inside the circle-of-confusion, while the impact from outside is discarded by design, formulated as
\begin{equation}
\textbf{A}^\text{(QC)}_{qk}=\frac{\exp(A_{qk})\odot C_{qk}}{\sum_s \exp(A_{sk}) \odot C_{qk}},
\label{eq:coc_attn}
\end{equation}
and the mask $C_{qk}$ is computed via
\begin{equation}
    C_{qk} = \text{Soft}[r_c(k) - c_i \cdot \|P_q - P_k\|_2].
\end{equation}
For easier optimization, we apply a differentiable soft edge function $\text{Soft}(\cdot)$, which becomes sharper as the training goes, following previous works~\cite{sheng2024dr}. The detailed implementation is given in \cref{sec:soft_edge}.

\noindent\textbf{Self-Occlusion Mask.}
So far, the attention module has been modified to focus on the neighborhood with a given radius calculated from per-pixel disparity. We then consider the self-occlusion, caused by other pixels blocking the light propagation in 3D space. Different from previous methods that builds upon multi-plane images~\cite{peng2022mpib,sheng2024dr}, we calculate the pixel-wise occlusion map with sampling. 

In practice, if a light source $s$ is visible to $P_q$ on the camera sensor, for any sampling point with disparity $\tilde{d}$ that lies between the light source $s$ and $P_q$, it should not be blocked by the scene. 
Through the collinear relationship, the pixel location of the sampling point $\tilde{P}$ can be computed as 
\begin{equation}
    \tilde{P} = \frac{(1-\tilde{d})d_s}{(1-d_s)\tilde{d}}(P_s-P_q)+P_q.
\end{equation}
Assuming a simple geometry of the scene, any sampling point should be closer to the camera so as to be not occluded. Therefore the visibility mask $M_\text{vis}$ is given by %
\vspace{-.5em}
\begin{equation}
\!\!\!\!M_\text{vis} =\!\!\!\! \bigwedge_{\tilde{d}\in (d_s, 1) }\!\!\left[\text{dis}\!\left(\!\frac{(1-\tilde{d})d_s}{(1-d_s)\tilde{d}}(P_s-P_q) + P_q\!\right) \!\!<\! \tilde{d} \right].
\label{eq:exp_occ}
\end{equation}
For a more accurate rendering of the light source's impact, we super-sample $k$ in the $\epsilon_s$ neighborhood of point light source $s$, and the PISA module is formulated as
\begin{equation}
    \text{Attn}(Q,K,V)_{qk} = (\textbf{A}^\text{(QC)}_{qk} \odot \mathbb{E}_{s\sim\mathcal{N}(P_k,\epsilon_s)}[M_\text{vis}]) V
\label{eq:occ_attn}
\end{equation}

\subsection{Data Synthesis Pipeline}\label{sec:our_dataset}
\begin{figure}[b]
    \centering
    \vspace{-.5em}
    \includegraphics[width=\linewidth]{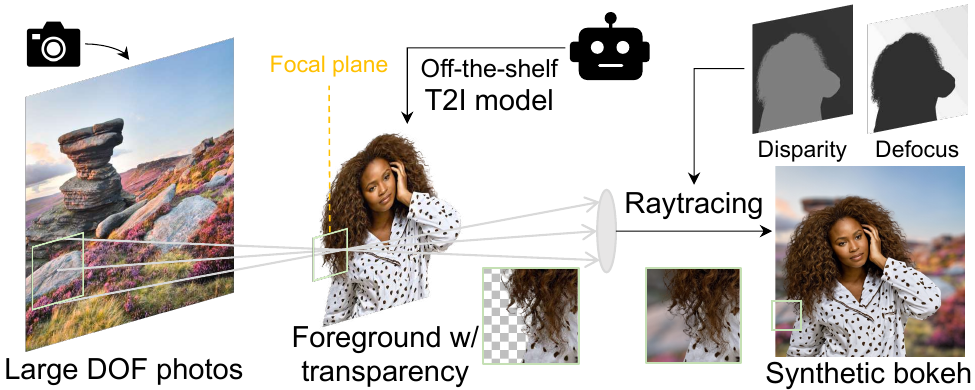}
    \vspace{-1.5em}
    \caption{The data synthesis pipeline. A pretrained text-to-image model is applied to generate foreground with transparency~\cite{zhang2024transparent}, and the large depth-of-field background is selected from real-world images. With the layers randomly placed with various facing angles and various depths, a classical ray-tracing method is applied to render the image with lens blur.}
    \label{fig:data_syn}
\end{figure}

To learn the mapping of $I_\text{f}\rightarrow I_\text{b}$, high-quality paired data is needed for fine-tuning the noise prediction network. As depicted in \cref{fig:data_syn}, the data synthesis pipeline follows previous works~\cite{peng2022bokehme,peng2022mpib,NTIRE_2023_CVPR} by using a ray tracing pipeline to synthesize images with various defocus amount and focus distances, given multiple layers of all-in-focus images. 
The bottleneck of previous works lies in the fact that high-quality foreground images are hard to acquire. 
Bounded by the accuracy of object segmentation, separating foreground objects from photos creates fake-looking photos~\cite{yang2023bokehornot}, especially on regions like hair and fur. On the other hand, photos with green screen background and 3D models are not suitable for synthesizing data at a large scale. 

We propose to synthesize photorealistic foreground with a state-of-the-art diffusion model~\cite{zhang2024transparent}, alleviating the dilemma of scalability and data quality.
Samples of the synthetic dataset are demonstrated in \cref{fig:example_dataset,fig:example_dataset2}. As shown in \cref{fig:data_syn}, we use real-world photos captured with a small aperture for background, and overlap it to synthetic foreground with transparency. By randomly placing the locations and facing angles of the layers, while controlling the focus on the average disparity of background or foreground, we are able to generate photorealistic synthetic data with a simplified ray tracer, with known disparity and focus distance, following the practice of previous methods~\cite{peng2022bokehme,peng2024bokehmepp}. Note that the skewed facing angles makes it possible to learn the progressive blurring caused by a continuously changing disparity map. In this way, the model can learn to render the scene faithfully to the disparity map, instead of to semantic information only.

\subsection{Supervision}\label{sec:supervision}
Previous latent diffusion methods usually calculate the loss function in latent space, but as this paper aims for detail reconstruction, the loss are calculated in pixel space. For a robust reconstruction of the shape, we first calculate the Mean Square Error (MSE) $\mathcal{L}_\text{MSE}$ between the predicted image $\hat{I}$ and the ground truth $I_\text{b}$. But as MSE is insensitive to blurriness, relying on MSE can lead to the trivial solution of returning the all-in-focus input, or the other extreme of over-blurring. Therefore we consider the following loss functions which should be more sensitive to the lens blur:

\noindent\textit{(i)} Perceptual loss $\mathcal{L}_\text{VGG}$. We apply the LPIPS loss which computes the distance between the image features extracted by a pretrained VGG network~\cite{zhang2018unreasonable}. 

\noindent\textit{(ii)} Multi-scale edge loss. As a strong visual clue, an obvious edge often indicates the image being in focus or not. To overcome the shortcomings of MSE, which can lead to blurry results, we follow previous works~\cite{seif2018edge,chen2023multi,nazeri2019edgeconnect} and design the loss to focus more on the edges before and after the lens blur effect is applied, given by
\vspace{-0.5em}
\begin{equation}
    \!\! \!\mathcal{L}_{\text{edge}} = \sum_{l=1}^{3} \frac{1}{l^2} \!\left\|(\nabla_l \hat{I} - \nabla_l I_\text{b}) \odot \!\max_{I\in\{I_\text{b},I_\text{f}\}}|\nabla_l I| \right\|_1,
\end{equation}
where $\nabla_l$ is the extended Sobel operator pair with the kernel size of $l$, in both horizontal and vertical directions. The term $\max_{I\in\{I_\text{b},I_\text{f}\}}|\nabla I|$ basically neglects smooth regions, which is already a easy target to be optimized with $\mathcal{L}_\text{MSE}$.

\noindent\textit{(iii)} Adversarial loss $\mathcal{L}_\text{adv}$. It employs a discriminator network $D$ with a pretrained ConvNext~\cite{convnext} backbone to distinguish real images with lens blur $I_b$ and generated images $\hat{I}$. The loss for the discriminator is given by 
\begin{equation}
    \mathcal{L}_{\text{D}} = \mathbb{E}_{I}[\log D(I_b)] + \mathbb{E}_{\hat{I}}[\log(1 - D(\hat{I}))],
\end{equation}
while $\mathcal{L}_{\text{adv}} = -\mathbb{E}_{\hat{I}}[\log D(\hat{I})]$ is used for finetuning diffusion model. In all, the finetuning loss is given by
\begin{equation}
    \mathcal{L} = \lambda_\text{MSE}\mathcal{L}_\text{MSE} + \lambda_\text{VGG}\mathcal{L}_\text{VGG} + \lambda_\text{edge}\mathcal{L}_\text{edge} + \lambda_\text{adv}\mathcal{L}_\text{adv}.
\end{equation}

\section{Experiments}\label{sec:exp}

\subsection{Experimental Settings}
\noindent\textbf{Baselines.} We select the following open-source state-of-the-art methods for baselines:
DeepLens~\cite{deeplens2018}, an early end-to-end neural rendering method trained on synthetic data;
MPIB~\cite{peng2022mpib}, a physics-based method that considers the scene in layers, which inpaints on each layer and then blends the multi-layer by classical rendering;
BokehMe~\cite{peng2022bokehme}, a hybrid rendering method that applies neural rendering in error-prone depth discontinuous regions, complementing the rest with a more controllable classical renderer;
Dr.Bokeh~\cite{sheng2024dr}, a hybrid rendering method that uses neural network for salient object segmentation and inpainting, and blends the layers differentiably.

We finetune the off-the-shelf BokehMe~\cite{peng2022bokehme} model with the same synthetic data, input (disparity map and all-in-focus image), and loss terms as BokehDiff, to further validate the effectiveness of the model design, in addition to a Restormer model~\cite{zamir2022restormer} trained from scratch.

\noindent\textbf{Datasets.} Quantitative experiments are conducted on the real-world EBB Val294~\cite{yang2023bokehornot,Mandl2024NeuralBokeh} dataset, and the synthetic datasets of BLB (Level 5)~\cite{peng2022bokehme} and \textsc{SynBokeh300} (synthesized as described in \cref{sec:our_dataset}). As EBB Val294 dataset contains slight misalignments, we align the global mean value of the input image to the ground truth bokeh image. Please refer to \cref{sec:dataset_description} for examples and more descriptions about the quantitative datasets.

For qualitative comparison and user study, the input images are gathered from the Unsplash dataset~\cite{unsplash}, the Easy Portrait dataset~\cite{EasyPortrait}, and some photos taken by the authors in the wild with an aperture of f/22.
The disparity maps are estimated by Depth Anything V2~\cite{depth_anything_v2,depth_anything_v1}, and are shared across all the methods for a fair comparison.

\noindent\textbf{Metrics.} Following previous works, we report Peak Signal-to-Noise Ratio (PSNR) that focuses on pixel-wise accurate estimation, and Structural Similarity (SSIM) that measures structural similarity to the ground truth. As pointed by previous works~\cite{sheng2024dr,zhang2018unreasonable}, PSNR is not sensitive to blurring. To complement the insufficient metrics, we additionally include LPIPS~\cite{zhang2018unreasonable} and DISTS~\cite{ding2020image} for perceptual similarity, which mimics the response of human vision. 

\begin{figure*}[b]
    \centering
    \vspace{-1em}
    \hspace*{\fill}
    \rotatebox{90}{\quad\quad\quad  \sffamily Defocus}\hfill
    \includegraphics[width=0.31\linewidth]{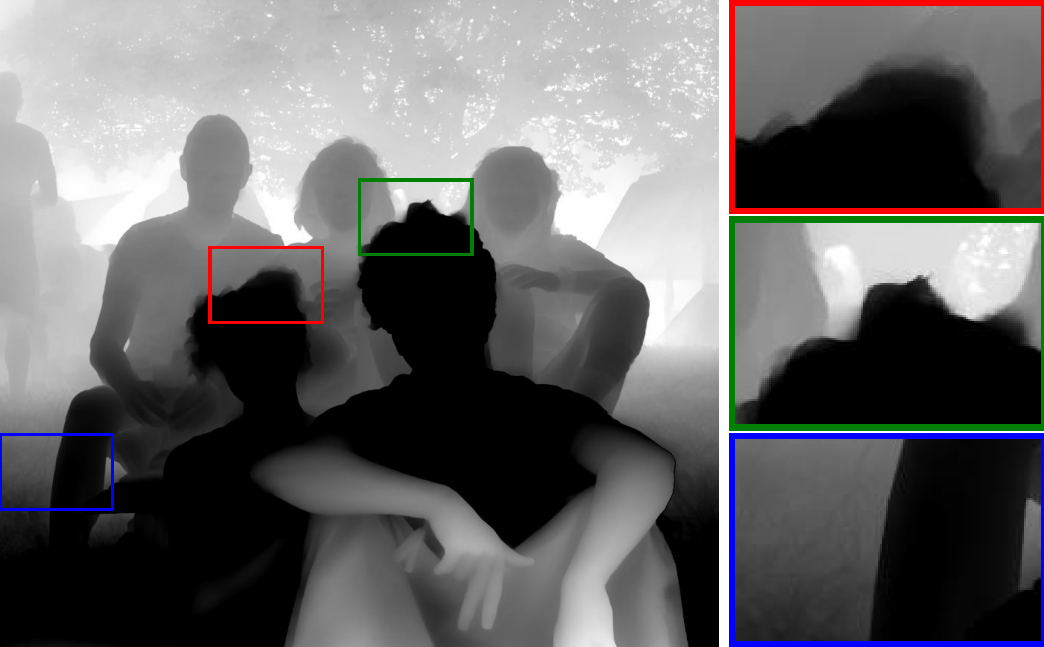}\hfill
    \includegraphics[width=0.31\linewidth]{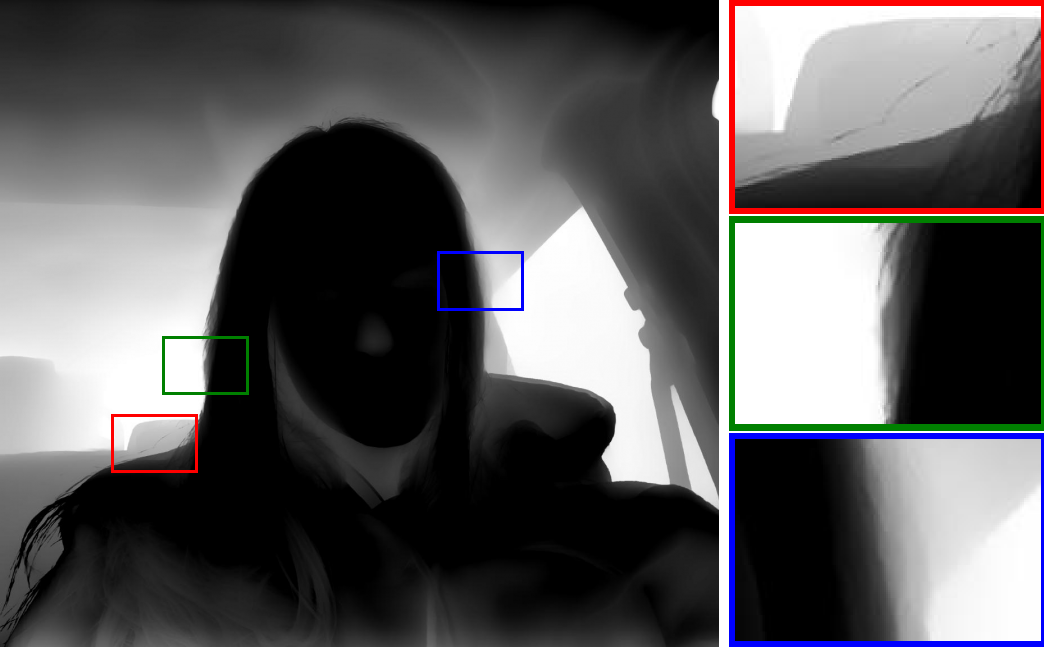}\hfill
    \includegraphics[width=0.31\linewidth]{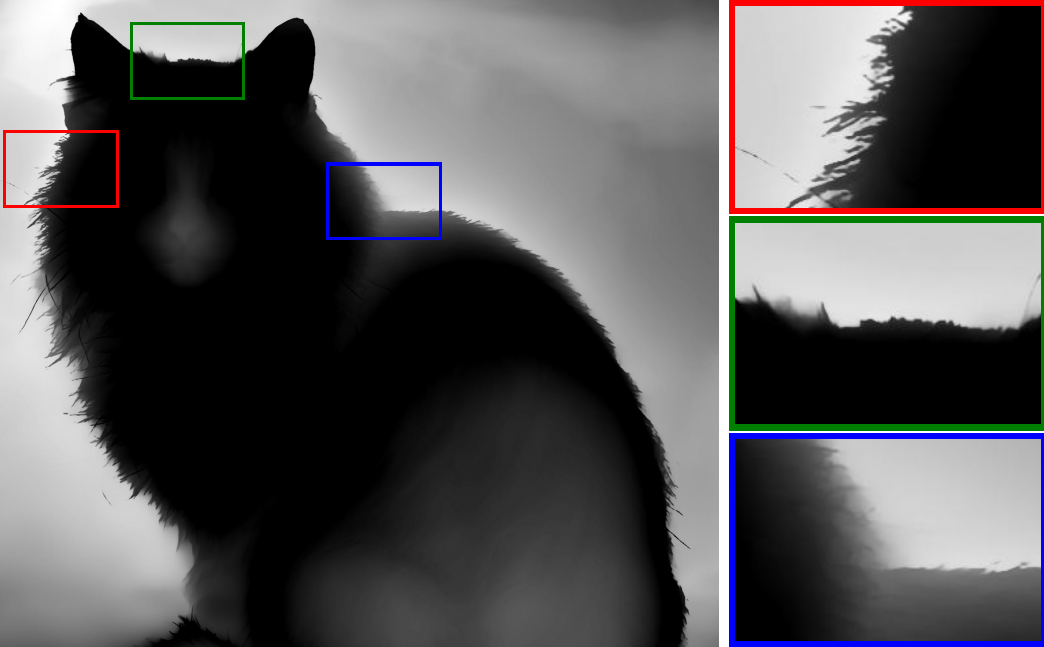}\hfill
    \hspace*{\fill}
    
\hspace*{\fill}
    \rotatebox{90}{\quad\quad \  \sffamily All-in-focus}\hfill
    \includegraphics[width=0.31\linewidth]{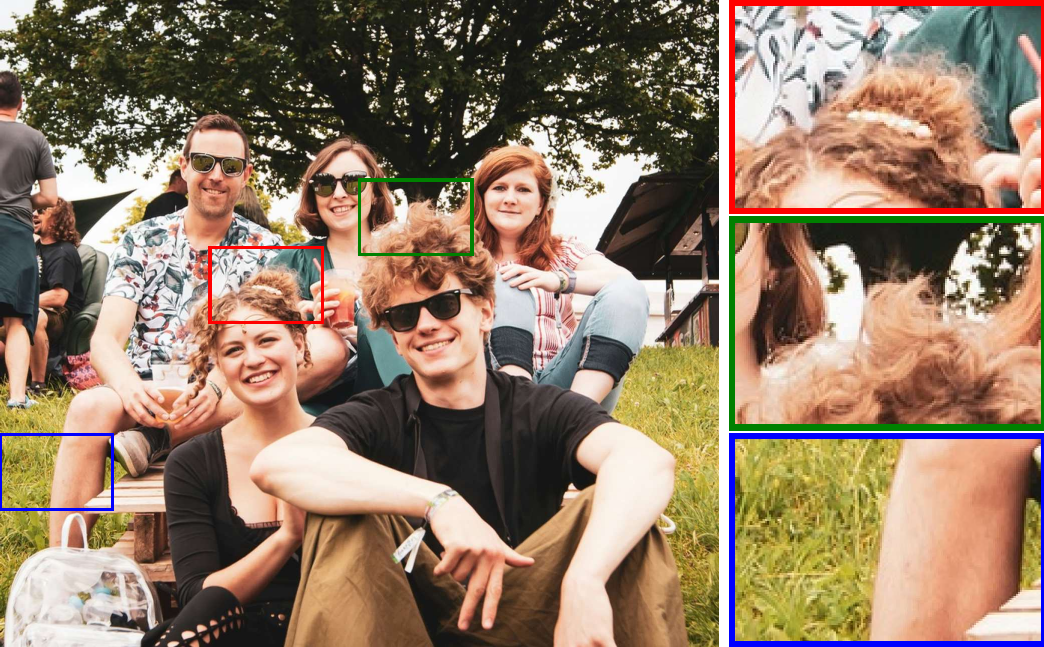}\hfill
    \includegraphics[width=0.31\linewidth]{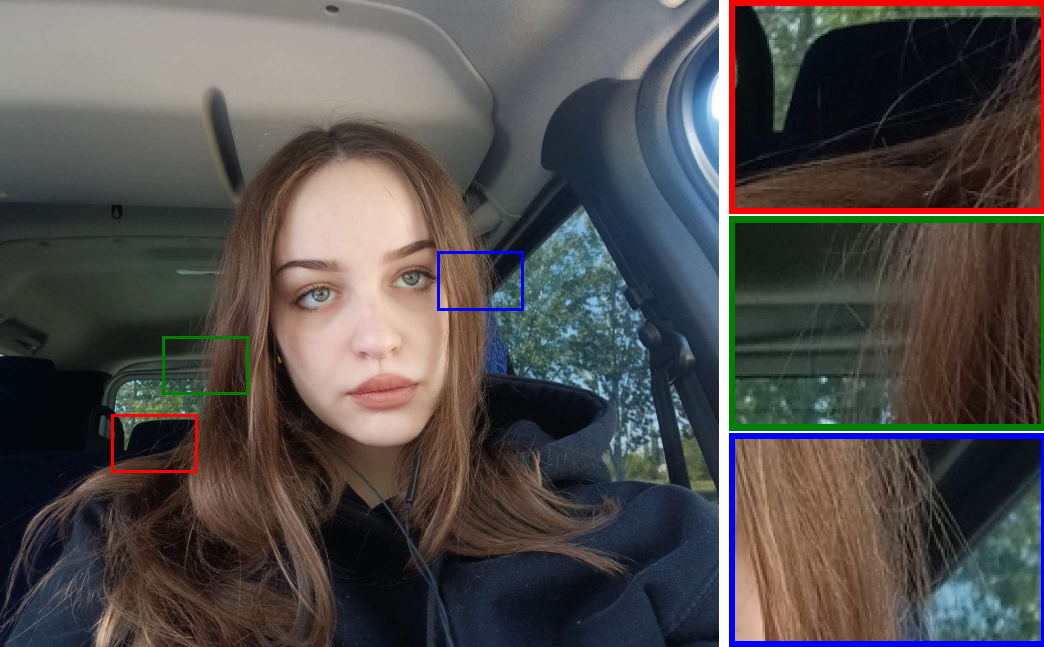}\hfill
    \includegraphics[width=0.31\linewidth]{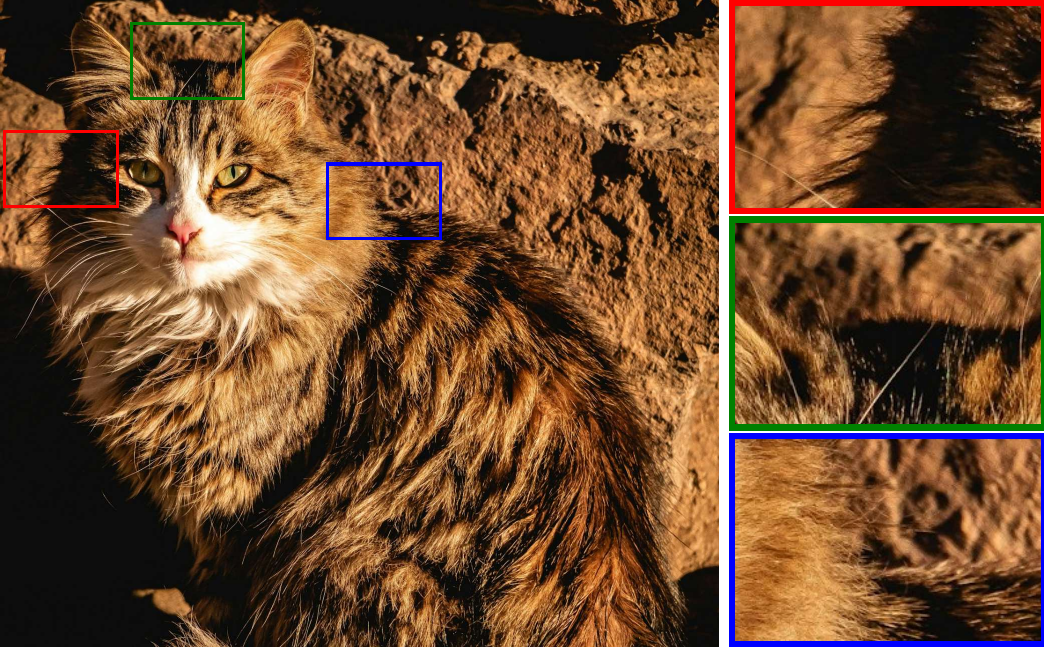}\hfill
    \hspace*{\fill}
    
    \hspace*{\fill}
    \rotatebox{90}{\quad\quad\ \ \sffamily BokehDiff}\hfill
    \includegraphics[width=0.31\linewidth]{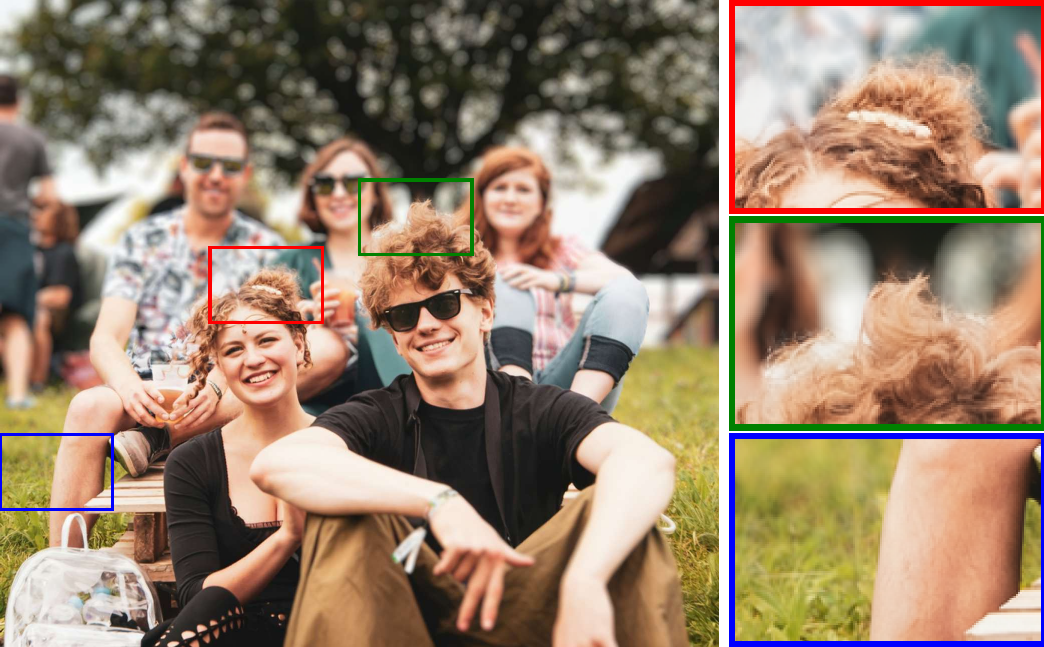}\hfill
    \includegraphics[width=0.31\linewidth]{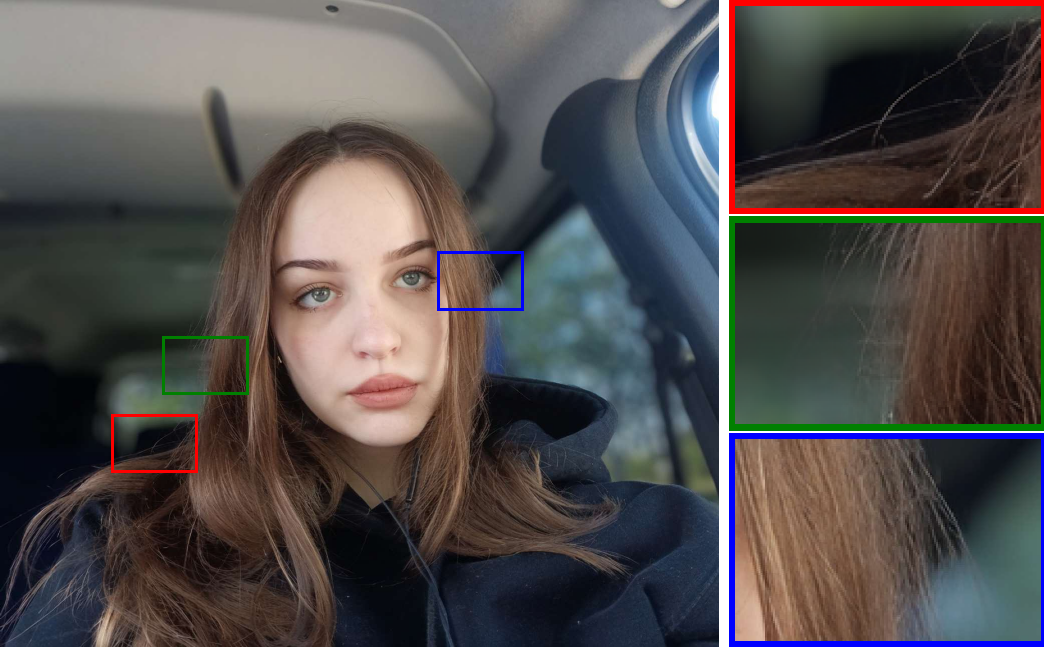}\hfill
    \includegraphics[width=0.31\linewidth]{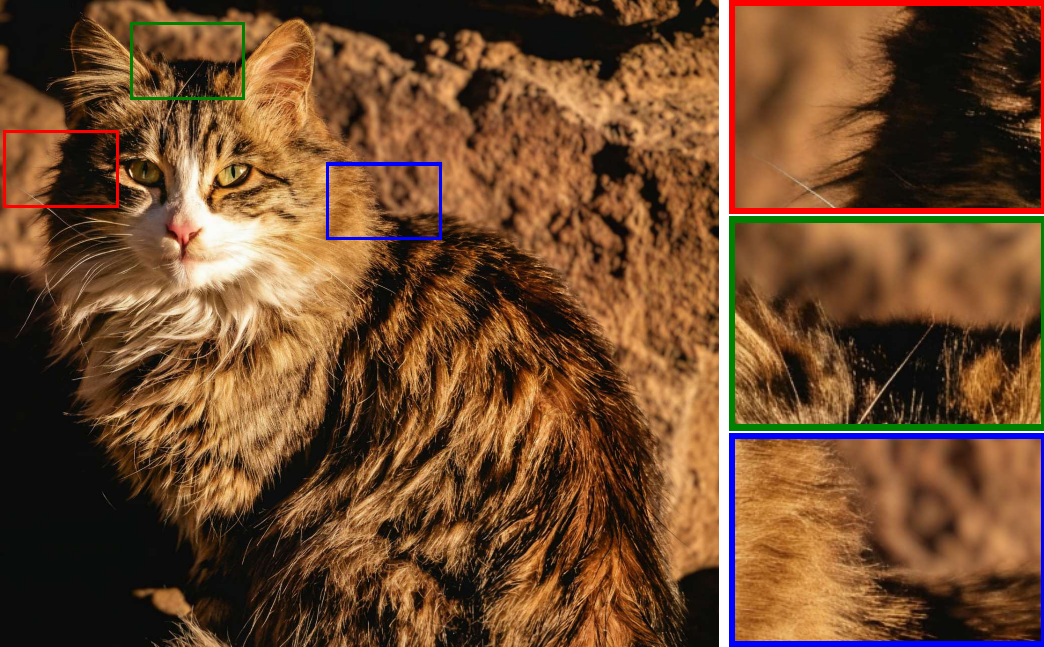}\hfill
    \hspace*{\fill}

    \hspace*{\fill}
    \rotatebox{90}{\quad\quad\  \sffamily BokehMe}\hfill
    \includegraphics[width=0.31\linewidth]{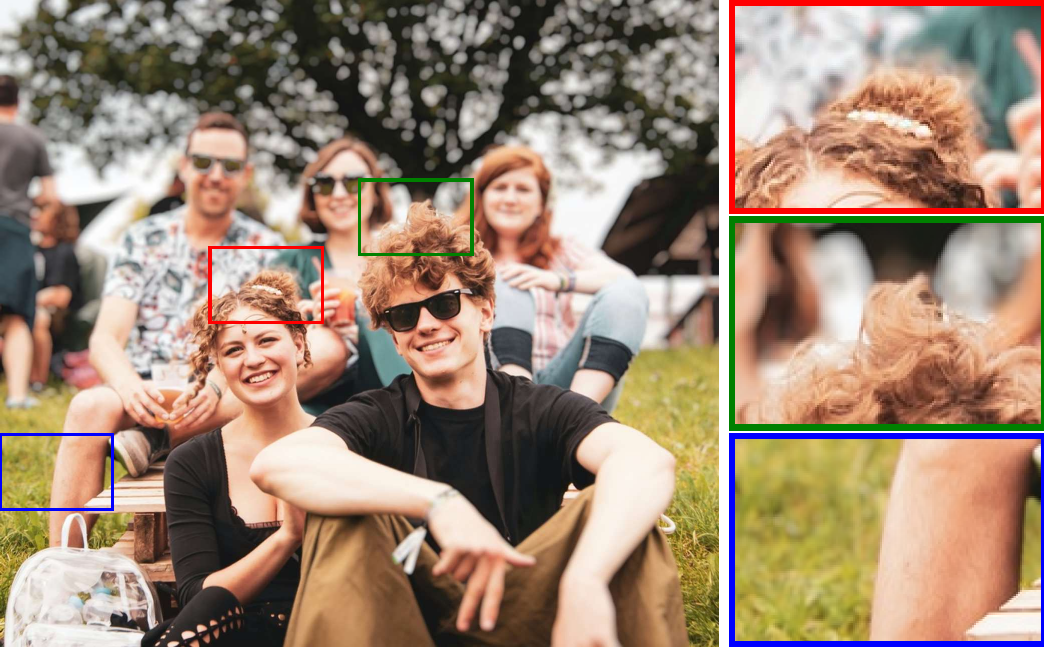}\hfill
    \includegraphics[width=0.31\linewidth]{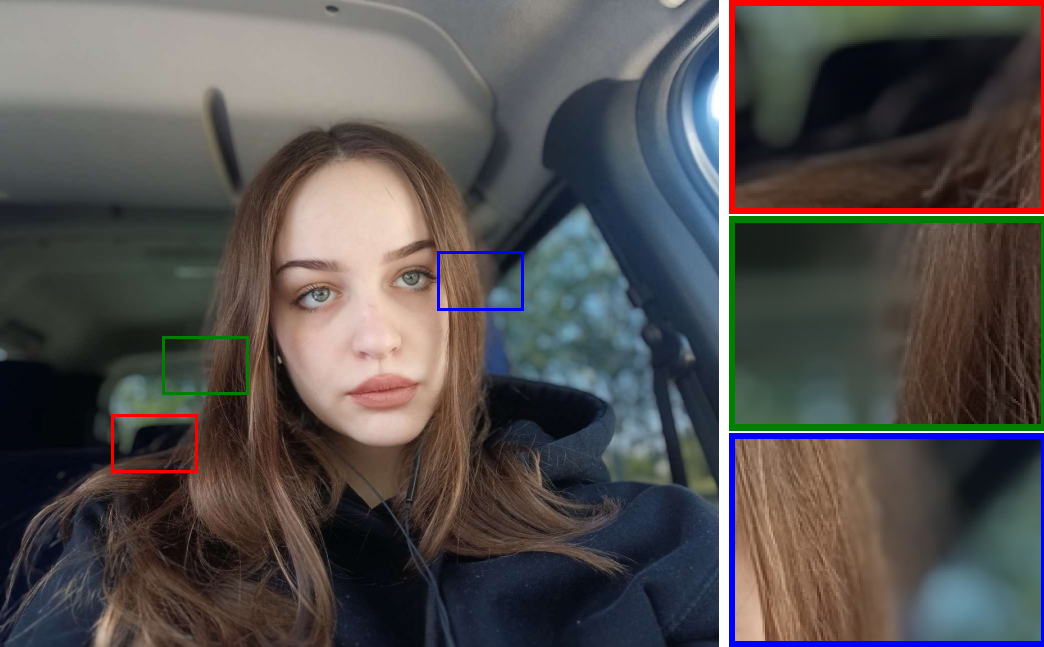}\hfill
    \includegraphics[width=0.31\linewidth]{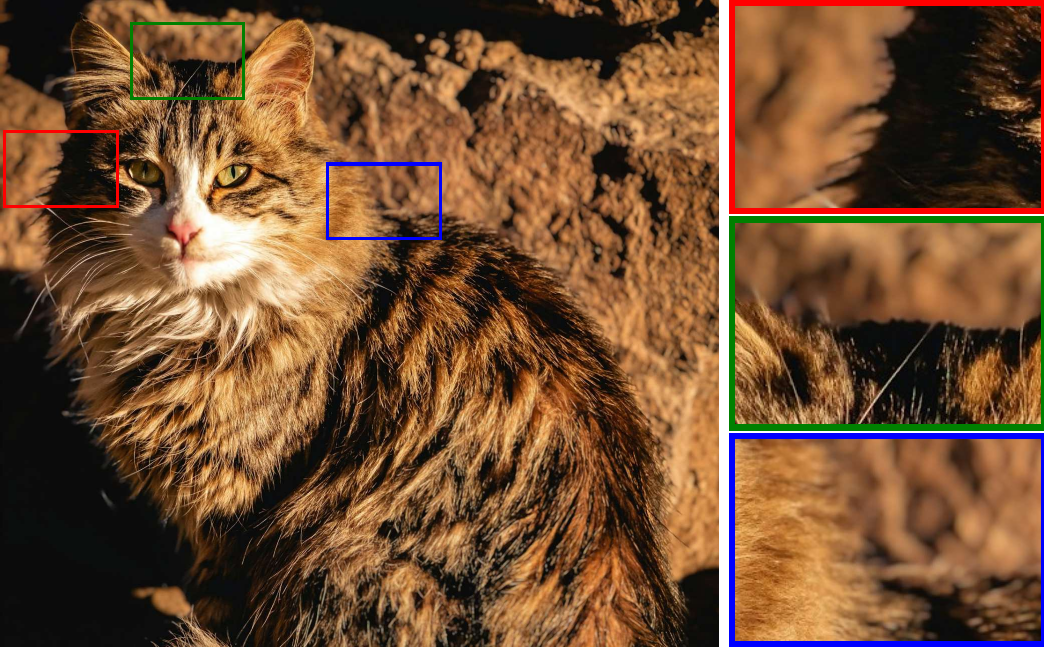}\hfill
    \hspace*{\fill}

     \hspace*{\fill}
    \rotatebox{90}{\quad\quad\quad\ \sffamily MPIB}\hfill
    \includegraphics[width=0.31\linewidth]{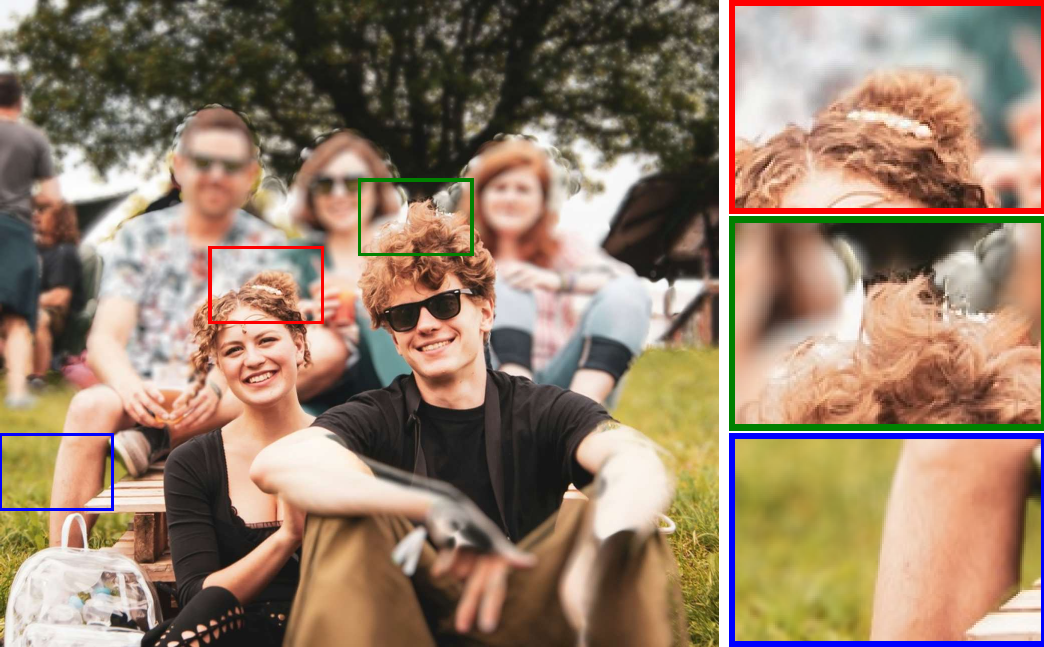}\hfill
    \includegraphics[width=0.31\linewidth]{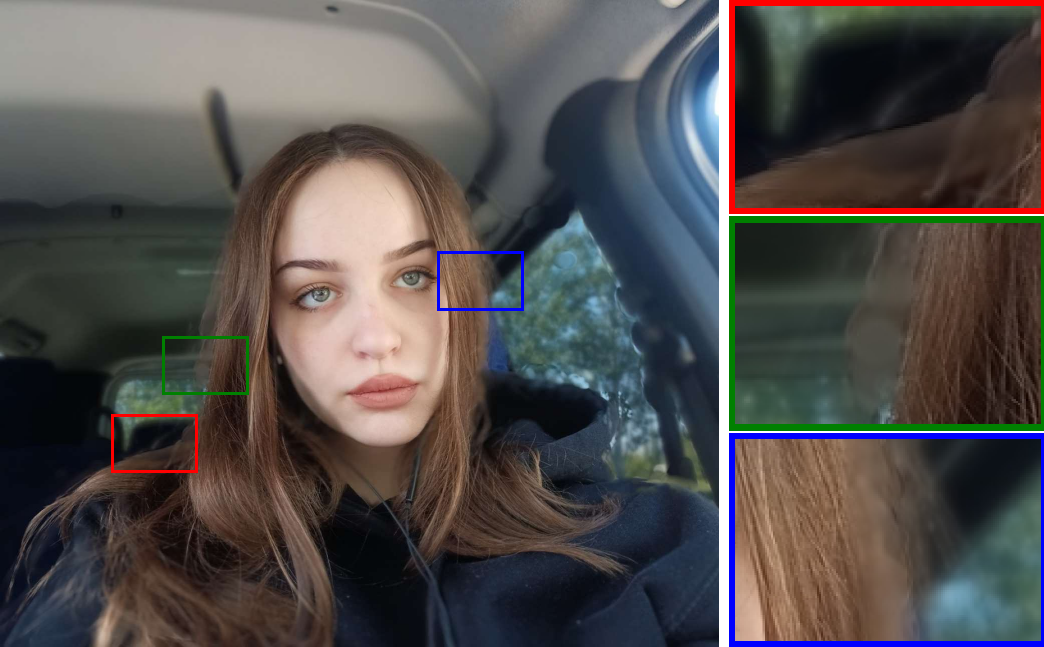}\hfill
    \includegraphics[width=0.31\linewidth]{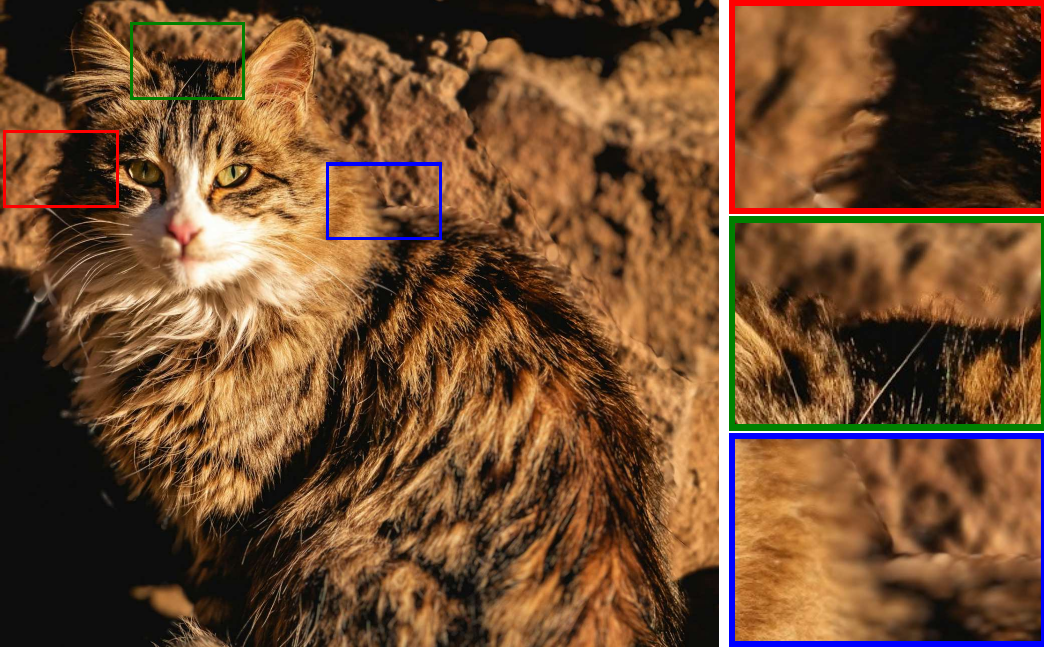}\hfill
    \hspace*{\fill}
    
    \hspace*{\fill}
    \rotatebox{90}{\quad\quad\ \ \sffamily  Dr.~Bokeh}\hfill
    \includegraphics[width=0.31\linewidth]{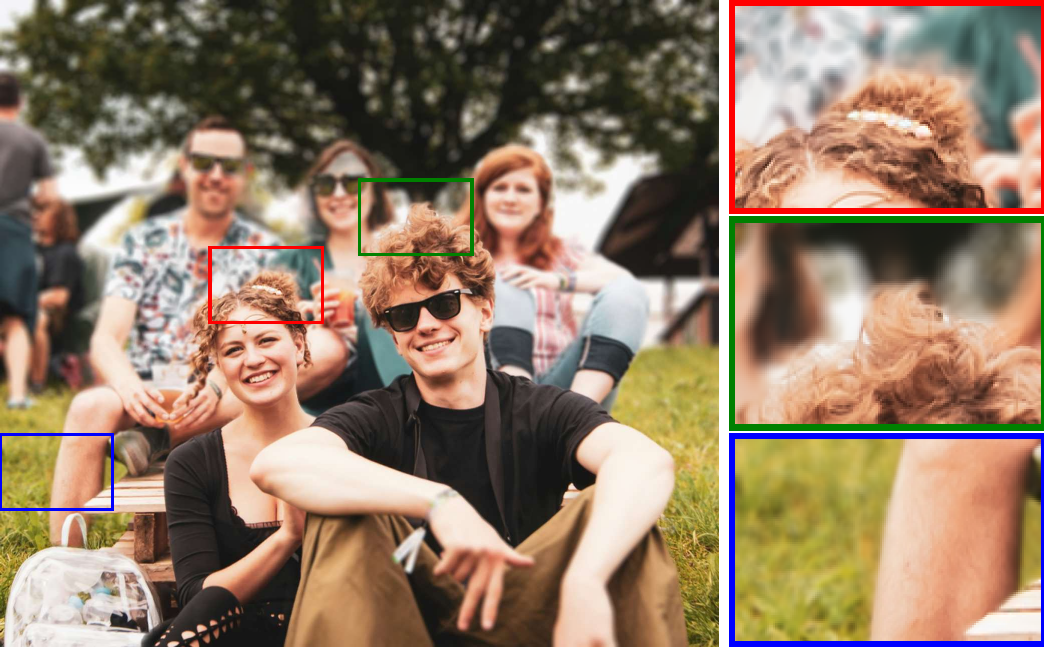}\hfill
    \includegraphics[width=0.31\linewidth]{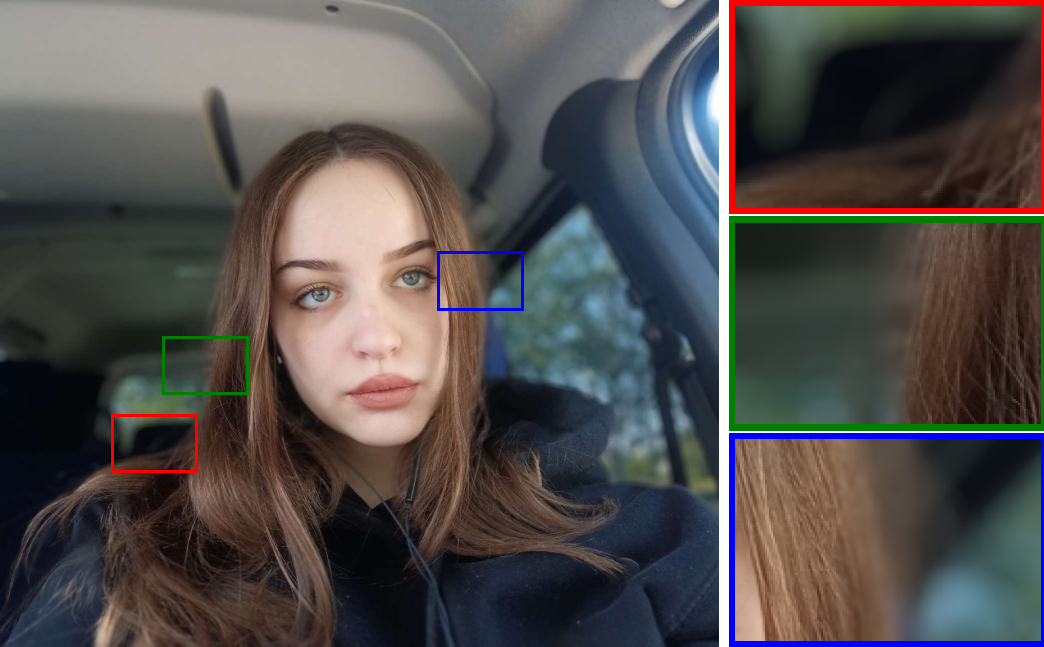}\hfill
    \includegraphics[width=0.31\linewidth]{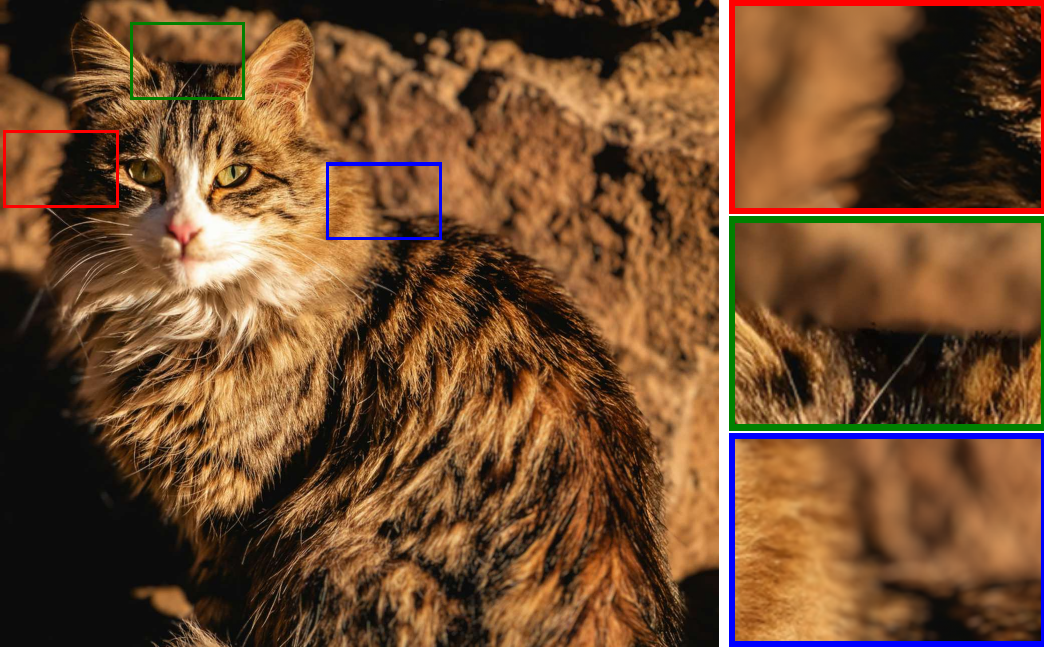}\hfill
    \hspace*{\fill}
    
    \vspace{-0.5em}
    \caption{The qualitative comparisons of BokehDiff with BokehMe~\cite{peng2022bokehme}, MPIB~\cite{peng2022mpib}, and Dr.~Bokeh~\cite{sheng2024dr}. Calculated from disparity, the defocus map is shared across the methods to be compared, and three patches are zoomed in for closer observation in each scene. Whiter region in the defocus map indicates more lens blur should be added, but is prone to error caused by depth estimation.}
    \label{fig:quali4}
    \vspace{-1em}
\end{figure*}

\noindent\textbf{Implementation Details.} The backbone model is a pretrained SDXL model~\cite{sg161222}, and only the LoRA~\cite{hu2021lora} of the downsampling layers in $\epsilon_\theta$ and the middle block and output layers of $\mathcal{E}$ are trained, and the rest of the diffusion network is fixed. The AdamW~\cite{adamw} optimizer is used, with a cosine annealing learning rate scheduler, starting from $10^{-4}$. The finetuning takes about 12 hours on a single NVIDIA L40s GPU, with a batch size of $2$. The rank of LoRA module is set at $8$ empirically.
For hyper-parameter settings, we have $\lambda_\text{MSE}=1$, $\lambda_\text{VGG}=5$, $\lambda_\text{adv}=0.5$, and $\lambda_\text{edge}=1$. 

\begin{table*}[t]
    \caption{Quantitative comparison on the exposure-aligned EBB Val294~\cite{ignatov2020aim} dataset (left), and the user study results (right). The ratings are for \textit{(i)} accuracy, \textit{(ii)} authenticity, and \textit{(iii)} preference. $\uparrow$ ($\downarrow$) indicates larger (smaller) values are better, and \textbf{bold} font indicates the best results. $^\bigstar$ denotes that the method is trained or finetuned on the same dataset as BokehDiff.}\label{tab:quanti_EBB}
\centering
\vspace{-1em}
\begin{tabular}{lcccccccccc}
\toprule
Dataset & \multicolumn{4}{c}{EBB Val294~\cite{ignatov2020aim} (real)} & \multicolumn{3}{c}{BLB Level 5~\cite{peng2022bokehme} (synthetic)} & \multicolumn{3}{c}{Real (user study)} \\
\cmidrule(r){1-1}\cmidrule(lr){2-5}\cmidrule(lr){6-8}\cmidrule(l){9-11}
Method  & PSNR$\uparrow$ & SSIM$\uparrow$& DISTS$\downarrow$ & LPIPS$\downarrow$ & PSNR$\uparrow$ & SSIM$\uparrow$& LPIPS$\downarrow$ & \textit{(i)}$\uparrow\!\!$ & \textit{(ii)}$\uparrow\!\!$ & \textit{(iii)}$\uparrow\!\!$ \\ 
\midrule
DeepLens~\cite{deeplens2018} & 22.703 & 0.7623 & 0.1483 & 0.4191 & 20.301 & 0.6901 & 0.2976 & 1.55 & 1.68 & 1.96 \\
MPIB~\cite{peng2022mpib} & 23.334 & 0.7920  & 0.1581 & 0.4031 & 28.162 & 0.8997 & 0.2561 & 1.83 &1.89 & 2.04 \\ %
BokehMe~\cite{peng2022bokehme} & 24.014 & 0.8134 & 0.1460 & 0.3921 & \textbf{38.802} & \textbf{0.9870} & 0.1404 &3.81 & 3.93 & 4.03
 \\ %
Dr.Bokeh~\cite{sheng2024dr}  & 23.479 & 0.8221  & 0.1225 & 0.3771 & 22.650 & 0.7452 & 0.4539 & 3.41 & 3.38 & 3.64 \\
\hline
Restormer$^\bigstar$~\cite{zamir2022restormer} & 23.960 & 0.7961  & 0.1297& 0.3778 & 16.781 & 0.6866 & 0.7802  & 2.85 & 3.02 & 2.92 \\
BokehMe$^\bigstar$~\cite{peng2022bokehme} & 23.753 & 0.7919 & 0.1437 & 0.3967 & 30.044 & 0.9409 & 0.1660 & 3.67 & 3.80 & 3.48 \\
BokehDiff  & \textbf{24.652} &\textbf{0.8357} & \textbf{0.1155} & \textbf{0.3737} & 36.798 & 0.9814 &   \textbf{0.0888} & \textbf{4.42} & \textbf{4.37} & \textbf{4.56} \\ %
\bottomrule
\end{tabular}
\vspace{-1em}
\end{table*}

\subsection{Results and Comparisons}
\noindent\textbf{Quantitative Comparisons.} Though the EBB Val294 dataset~\cite{ignatov2020aim} involves aberration, camera motion, and other uncontrollable factors, BokehDiff still surpasses all previous baselines, as shown in the left columns in \cref{tab:quanti_EBB}.
For a more informed comparison, the comparison on the \textit{original} (not exposure-aligned) EBB Val294 dataset~\cite{ignatov2020aim} is listed in \cref{tab:quanti_EBB_original}. 

For the BLB dataset, the multi-layer based methods (MPIB~\cite{peng2022mpib} and Dr.~Bokeh~\cite{peng2022bokehme}) fail due to the complex scene layout, while the learning based DeepLens~\cite{deeplens2018} and Restormer~\cite{zamir2022restormer} also fails due to the insufficient knowledge of the underlying physics, as shown in the middle columns of \cref{tab:quanti_EBB}. Both BokehMe~\cite{peng2022bokehme} and BokehDiff have a decent performance, while the blur-sensitive LPIPS~\cite{zhang2018unreasonable} indicates that BokehDiff renders more realistic bokeh pattern.

\begin{figure}[tb]
    \centering
    \vspace{-1em}
    \includegraphics[width=\linewidth]{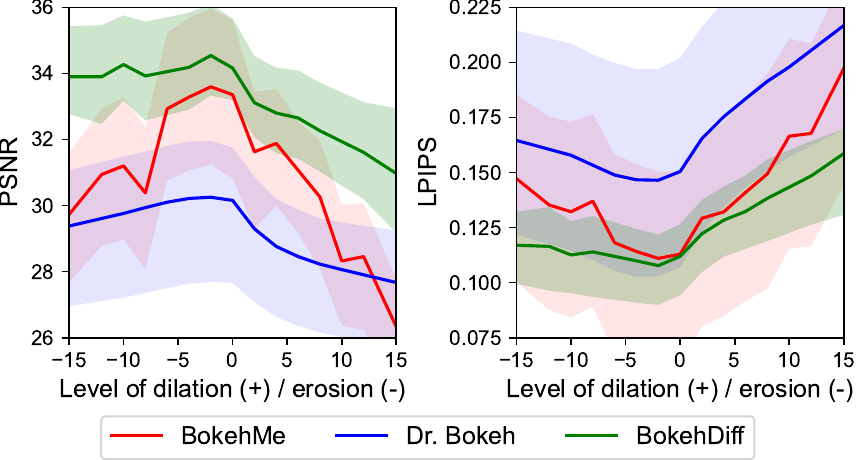}
    \vspace{-2em}
    \caption{PSNR and LPIPS performance drop with respect to the erosion or dilation to the disparity map, on \textsc{SynBokeh300}. The semi-transparent area around each line is bounded by the quartiles.}
    \label{fig:erosion}
\end{figure}

To measure the robustness to depth prediction error, we follow BokehMe~\cite{peng2022bokehme} and conduct a test on \textsc{SynBokeh300} dataset by eroding and dilating the disparity map.
Shown in \cref{fig:erosion}, BokehDiff constantly outperforms BokehMe~\cite{peng2022bokehme} and Dr.~Bokeh~\cite{sheng2024dr}, with a less performance drop as the degeneration level raises, and the narrower quartiles further shows the stability of BokehDiff.

\noindent\textbf{User Study.} 
We conduct a user study, in which $50$ volunteers with at least 1 year of photography experience are involved. Participants are shown with the all-in-focus image and the rendered results, and are asked to rate the results from 1 to 5. For each case, participants are randomly asked to focus on one of the following aspects: \textit{(i)} accuracy, \eg, the edge should be the same blurry as the surface on which it is located; \textit{(ii)} authenticity, \eg, the blurriness should change gradually with respect to the distance from focal plane; or simply \textit{(iii)} preference as users. The results are listed on the rightmost columns in \cref{tab:quanti_EBB}.

\begin{figure*}
    \centering
    \includegraphics[width=\linewidth]{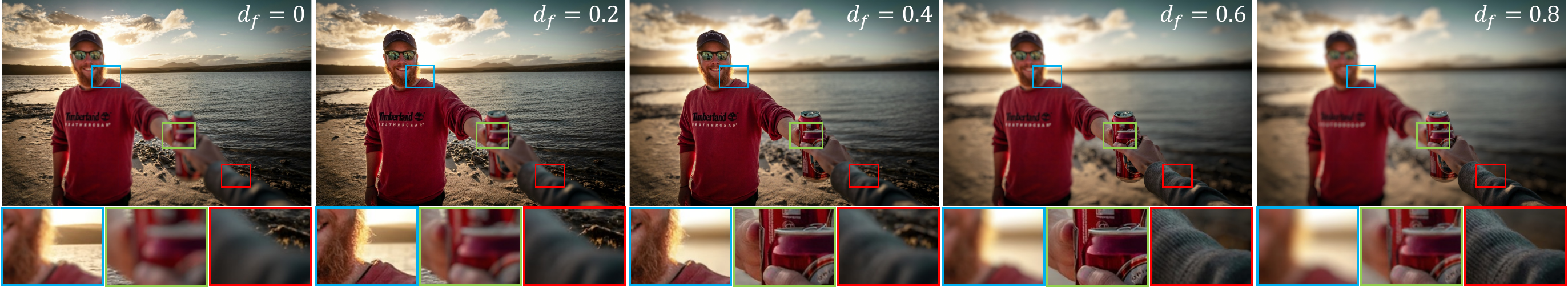}
    \vspace{-1.5em}
    \caption{A synthetic focal stack of BokehDiff, given an all-in-focus image selected from the Unsplash~\cite{unsplash} dataset.}
    \label{fig:focal_stack}
    \vspace{-0.5em}
\end{figure*}

\noindent\textbf{Qualitative Comparisons.} According to \cref{tab:quanti_EBB}, we only show the methods with superior quantitative performance here.
In \cref{fig:quali4}, three exemplar cases are shown, with more shown in the supplementary material. BokehDiff manages to maintain the intricate hair and fur details of the focused foreground in every example, even when the erroneous depth estimation erodes or dilates the defocus map. The transition from the focal plane to blurriness is smooth, as shown from the grass in the first column and the car roof in the second column. It can also blur the foreground off focus, such as the hands of the teenager in the first example.

As for the baselines, BokehMe~\cite{peng2022bokehme} has the second best quality, by being loyal to the defocus map. Thus it also fails when depth estimation is inaccurate, especially in intricate depth discontinuities. In the zoomed patches of \cref{fig:quali4}, it over-blurs the hair and the cat's whiskers that should be focused.
MPIB~\cite{peng2022mpib} fails to piece together layers where the complex scene cannot be easily separated into layers, such as the obvious artifacts around the teenager's arm and the man in the back rank in the first example. Though it sometimes renders the hair streaks right, it cannot generate progressive blur as the more focused background in the second and third examples show.
Dr.~Bokeh~\cite{sheng2024dr} does well in cases with a clean separation between foreground and background, but is also limited to the accuracy of the depth estimation, and the number of layers in question. It shows dark tints on the woman in the back rank, and also fails due to the inaccurate depth estimation.

In addition, we adjust the disparity of focal plane $d_f$, and show a focal stack in \cref{fig:focal_stack}, verifying the ability to focus on any designated depth of BokehDiff.

\subsection{Ablation Study}

The results for the ablation study are listed in \cref{tab:ablation2}
We first ablate the supervisions for the one-step diffusion scheme by removing $\mathcal{L}_\text{adv}$, $\mathcal{L}_\text{VGG}$, and $\mathcal{L}_\text{edge}$. With the same training iterations, these settings achieve an inferior performance, especially when removing the multi-scale edge loss and the perceptual loss.
We then consider the PISA module, namely the energy-conserved normalization (\cref{eq:normalize_Q}), the circle-of-confusion constraint (\cref{eq:coc_attn}), and the self occlusion (\cref{eq:occ_attn}). The complete model excels in LPIPS and visual effects (shown in the supplementary material), validating the design of the PISA module. A fixed encoder slightly decreases the performance, as the backbone needs to modify the latent more in this setting.
Different timestep configurations are also tested, and the results indicate similar performance with $T=249$ or $T=749$. But in practice, extremely large or low timestep can easily lead to gradient explosion, and $499$ is the choice of balance.

\section{Conclusions}\label{sec:conclusions}
The paper proposes BokehDiff, a diffusion framework with only one inference step that achieves outstanding quality compared with previous methods, especially in regions where depth prediction fails. The diffusion priors, combined with the PISA module which is specifically designed for physics constraint, shed light on a new possibility for neural lens blur rendering and physic-based deep learning. Quantitative comparisons, visual results, and a user study all validate that BokehDiff is able to synthesize photorealistic lens blur, and robust against error in depth estimation.
\begin{table}[t]
\vspace{-1em}
\caption{The ablation study conducted on the exposure-aligned EBB Val294~\cite{ignatov2020aim} dataset. The setting of ``SoftmaxQ'', ``CoC'', and ``occlusion'' are short for the energy-conserved normalization, circle of confusion constraint, and self-occlusion respectively. }\label{tab:ablation2}
\vspace{-0.5em}
\centering
\begin{tabular}{lccc}
\toprule
Setting  & PSNR$\uparrow$ & SSIM$\uparrow$& LPIPS$\downarrow$ \\ \midrule
w/o $\mathcal{L}_\text{adv}$ &24.623 &0.8322 &0.3768 \\
w/o $\mathcal{L}_\text{VGG}$ &24.285&0.8196&0.4218 \\
w/o $\mathcal{L}_\text{edge}$ &24.628 &0.8346&0.3785\\
fixed $\mathcal{E}$ & 24.266 & 0.8286 & 0.3811    \\\midrule
w/o CoC & 22.217 & 0.6881 & 0.4280 \\ %
w/o SoftmaxQ & 24.468 & 0.8325 & 0.3800 \\
w/o occlusion & 24.399 & 0.8291 & 0.3808 \\ %
\midrule
$T=249$ & 24.646 & 0.8335 & 0.3781 \\ 
$T=749$ & 24.481 & 0.8319 & 0.3838 \\
Complete model &\textbf{24.652} &\textbf{0.8357} & \textbf{0.3737}\\
\bottomrule
\end{tabular}
\vspace{-0.5em}
\end{table}

\noindent\textbf{Limitations.} 
Though the finetuned diffusion network keeps the majority of the structures from the all-in-focus image, the decoder of the VAE still cause inevitable changes to less noticeable structures. The issues can be addressed by changing the diffusion backbone~\cite{flux,sd3} with less information compression and better detail preservation. 

\section*{Acknowledgement}
This work is supported by National Natural Science Foundation of China under Grant No.~62136001, 62088102, and 62276007. 
PKU-affiliated authors would like to thank \href{openbayes.com}{openbayes.com} for
providing computing resource.
We thank Zhifeng Wang, Zhihao Yang, and Yichen Sheng for providing access and advice for the EBB!~\cite{ignatov2020aim} dataset. 
As an important source for lens blur effects rendering demonstration, we appreciate the photographers with Unsplash, the Unsplash developers, and Victor Ballesteros for granting the access to the Unsplash dataset~\cite{unsplash}.
Also thanks to Jiangang Wang and other colleagues for the discussions during Chengxuan Zhu's internship at Vivo.

\appendix
\section{Experimental Details}
\subsection{Network Design}\label{sec:soft_edge}
As mentioned in Eq.~(6) of the paper, we show the implementation of function $\text{Soft}(\cdot)$ at the iteration of $k$ as
\begin{equation}
    \text{Soft}(x, k) = (1+\min(k,k_\text{max}) \exp(x))^{-1},
\end{equation}
where the threshold $k_\text{max}$ is set as $10^6$ empirically. The design rationale is to gradually approach a step function, and to keep it unchanged after the training step reaches the threshold. During inference, we fix $k=k_\text{max}$.

\subsection{Efficiency Comparison}
As an important factor in real-world deployment, the durations of the proposed method and the baselines are compared. Profiled on EBB Val294~\cite{ignatov2020aim}, the average interval of bokeh rendering is listed in \cref{tab:quanti_EBB_original}. All the tests are conducted on a single NVIDIA RTX A6000 GPU, and only the duration of the model forward time is calculated. Note that MPIB~\cite{peng2022mpib} and Dr.~Bokeh requires significantly longer time to run, because they requires per-layer inpainting in their multi-layer representations.

\begin{table*}[b]
    \caption{Quantitative comparison on the exposure-aligned and the original EBB Val294~\cite{ignatov2020aim} dataset with the same optimized aperture as the paper (left), and the aperture that is optimized in the original EBB Val294~\cite{ignatov2020aim} datset (right). $\uparrow$ ($\downarrow$) indicates larger (smaller) values are better, and \textbf{bold} font indicates the best results. $^\bigstar$ denotes that the method is trained with the same dataset as BokehDiff.}\label{tab:quanti_EBB_original}
\centering
\vspace{-.5em}
\begin{tabular}{lccccccccc}
\toprule
\multicolumn{2}{c}{Dataset}  & \multicolumn{4}{c}{Exposure-aligned EBB Val294~\cite{ignatov2020aim}} & \multicolumn{4}{c}{Original EBB Val294~\cite{ignatov2020aim}} \\
\cmidrule(r){1-2}\cmidrule(lr){3-6}\cmidrule(l){7-10}
Method & Duration (s) & PSNR$\uparrow\!\!$ & SSIM$\uparrow\!\!$& DISTS$\downarrow\!\!$ & LPIPS$\downarrow\!\!$ & PSNR$\uparrow\!\!$ & SSIM$\uparrow\!\!$& DISTS$\downarrow\!\!$ & LPIPS$\downarrow\!\!$ \\ 
\midrule
DeepLens~\cite{deeplens2018} & 0.402 & 22.703 & 0.7623 & 0.1483 & 0.4191 & 22.065  & 0.7604 & 0.1509 & 0.4224 \\
MPIB~\cite{peng2022mpib}& 31.87 & 23.334 & 0.7920  & 0.1581 & 0.4031 & 22.450 & 0.7892 & 0.1616 & 0.4056  \\ %
BokehMe~\cite{peng2022bokehme}& 1.531 & 24.014 & 0.8134 & 0.1460 & 0.3921 & 23.247 & 0.8117 & 0.1463 & 0.3918
 \\ %
Dr.Bokeh~\cite{sheng2024dr} & 99.67 & 23.479 & 0.8221  & 0.1225 & 0.3771  & 21.298 & 0.8061 & 0.1338 & 0.3878  \\
Restormer$^\bigstar$~\cite{zamir2022restormer}& 0.962 & 23.960 & 0.7961  & 0.1297& 0.3778 & 23.188 & 0.7964 & 0.1314 & 0.3801 \\
BokehMe$^\bigstar$~\cite{peng2022bokehme}& 1.531 & 23.753 & 0.7919 & 0.1437 & 0.3967 &  22.857 & 0.7886 & 0.1458 & 0.3998 \\
BokehDiff& 3.974  & \textbf{24.652} &\textbf{0.8357} & \textbf{0.1155} & \textbf{0.3737} & \textbf{23.728} &  \textbf{0.8390} &  \textbf{0.1148} &  \textbf{0.3711} \\ \bottomrule
\end{tabular}
\end{table*}

\subsection{Datasets and Quantitative Comparisons}\label{sec:dataset_description}
\begin{figure}
    \centering
    \begin{minipage}[c]{\linewidth}
    \subfloat[All-in-focus image]{
    \centering
    \includegraphics[width=0.31\linewidth,height=0.2\linewidth]{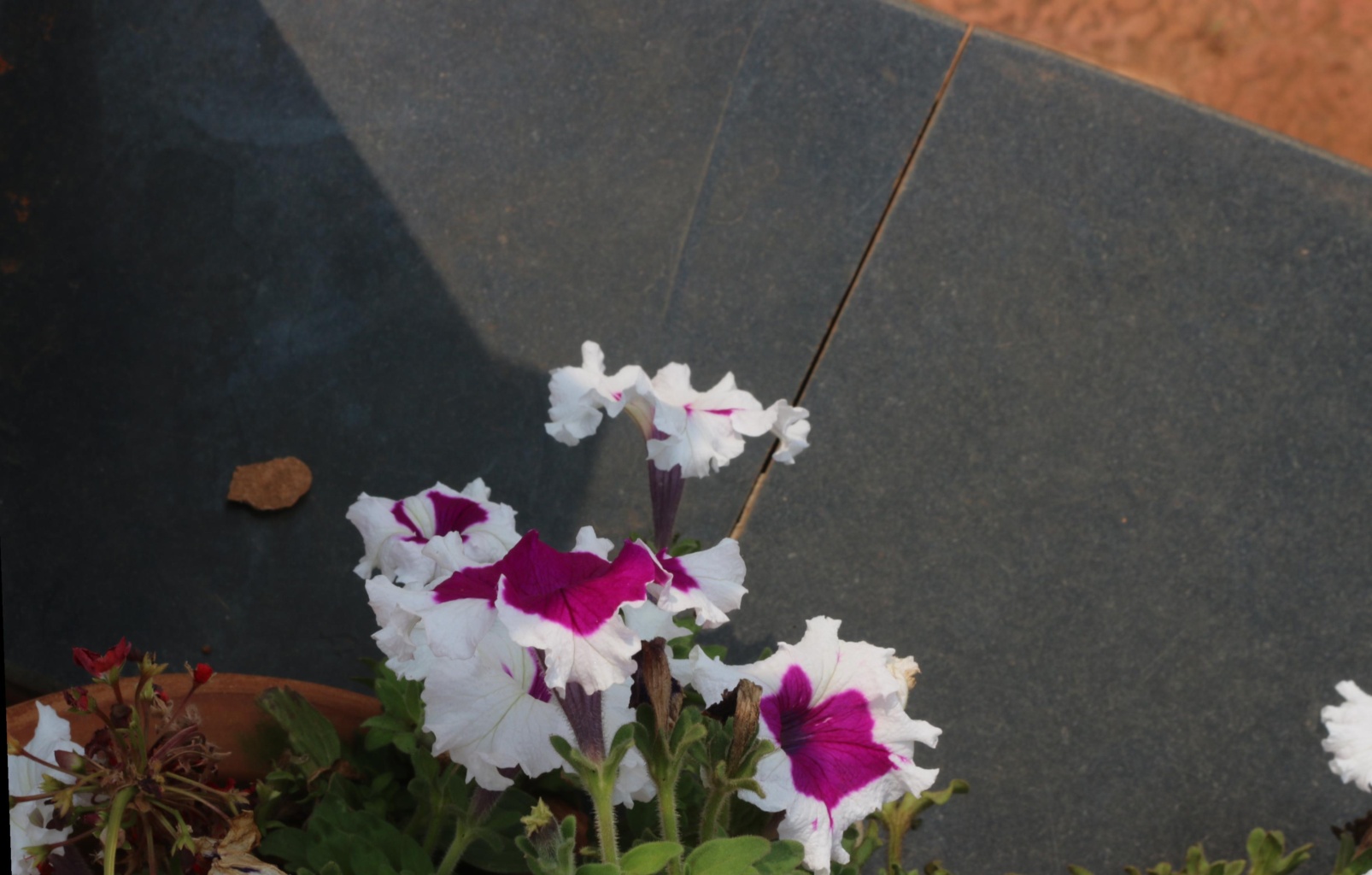} 
    \includegraphics[width=0.31\linewidth,height=0.2\linewidth]{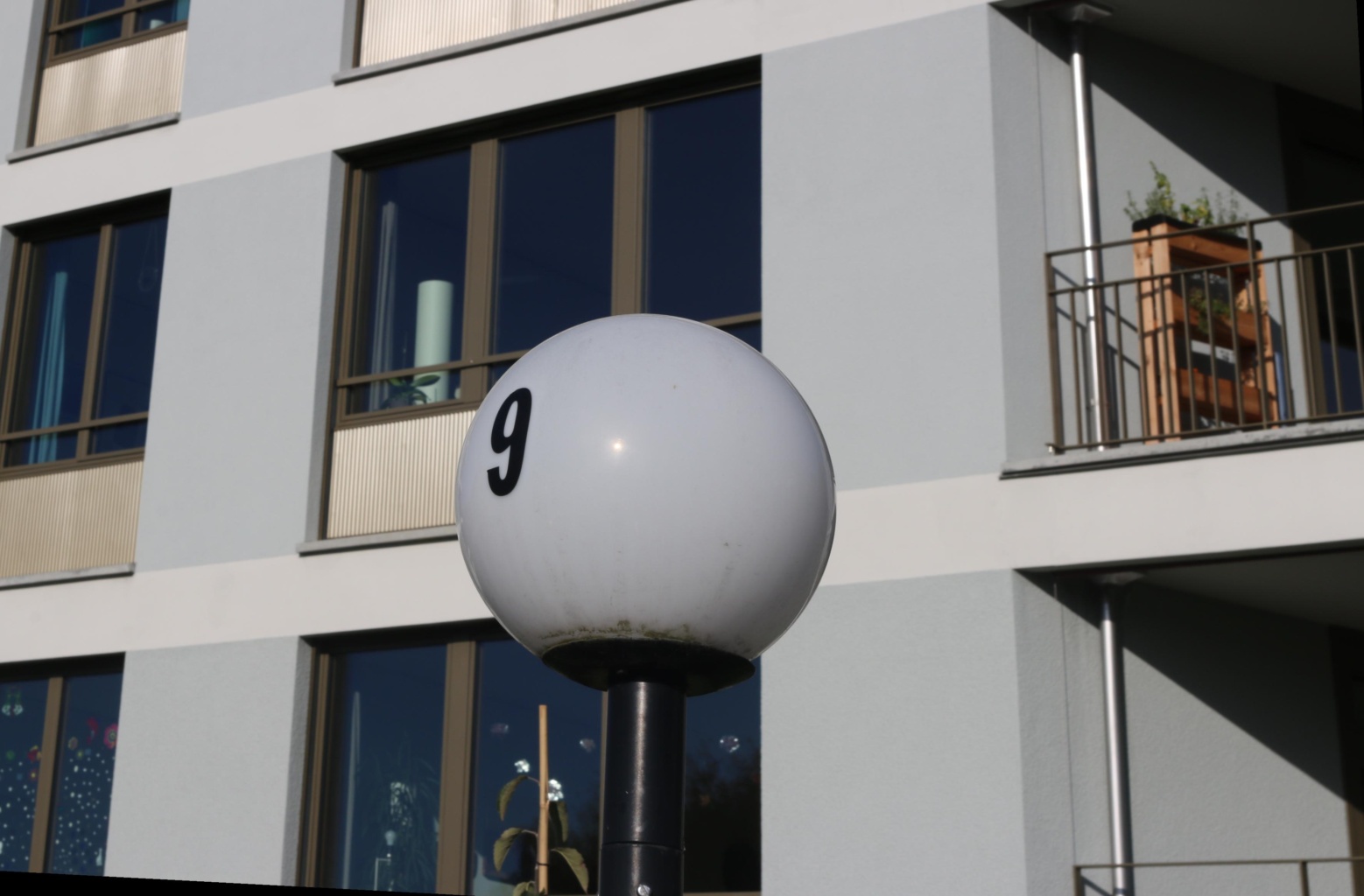}
    \includegraphics[width=0.31\linewidth,height=0.2\linewidth]{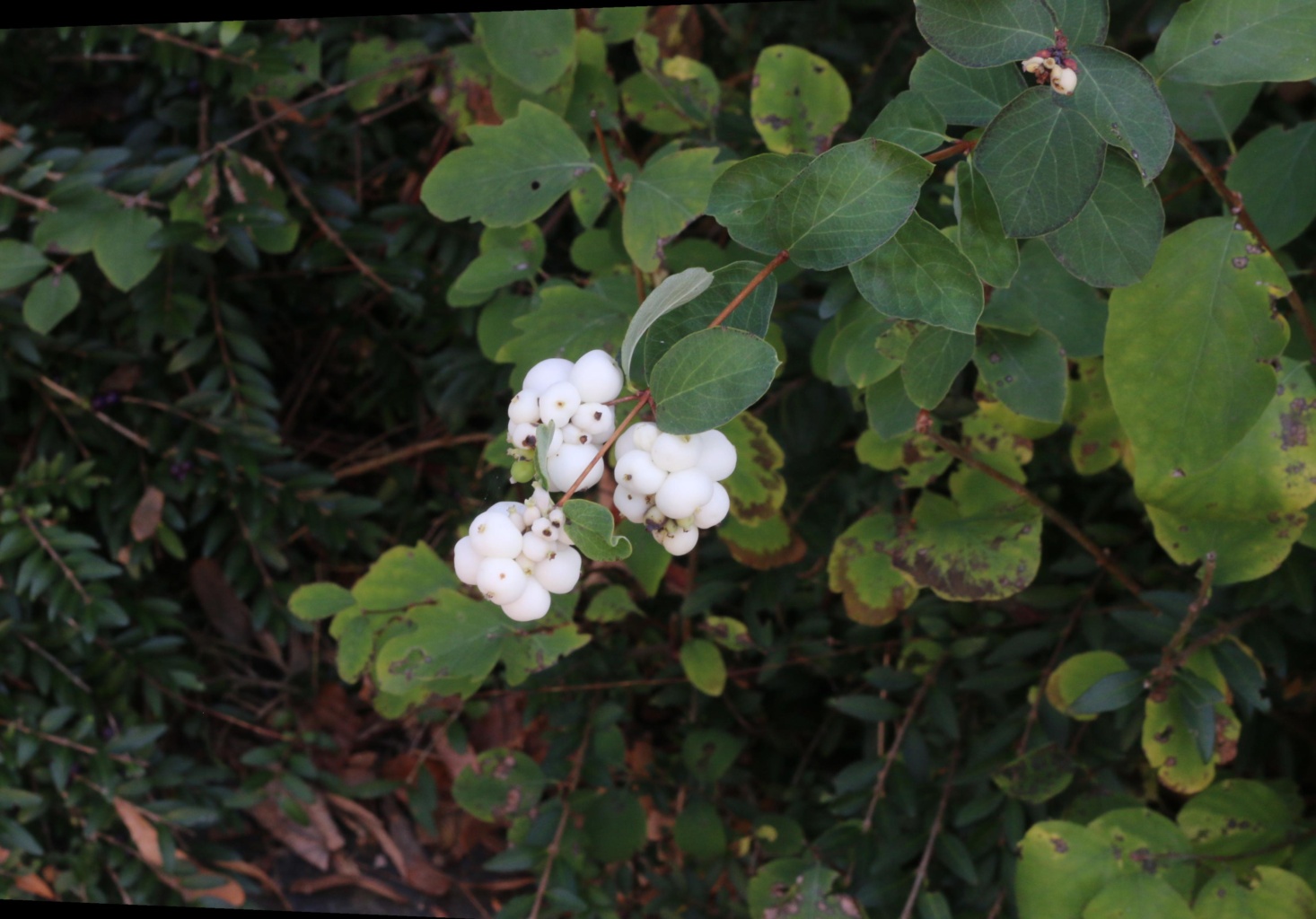} 
    }\hfill
    \\
    \subfloat[Lens blur image (real)]{
    \centering
    \includegraphics[width=0.31\linewidth,height=0.2\linewidth]{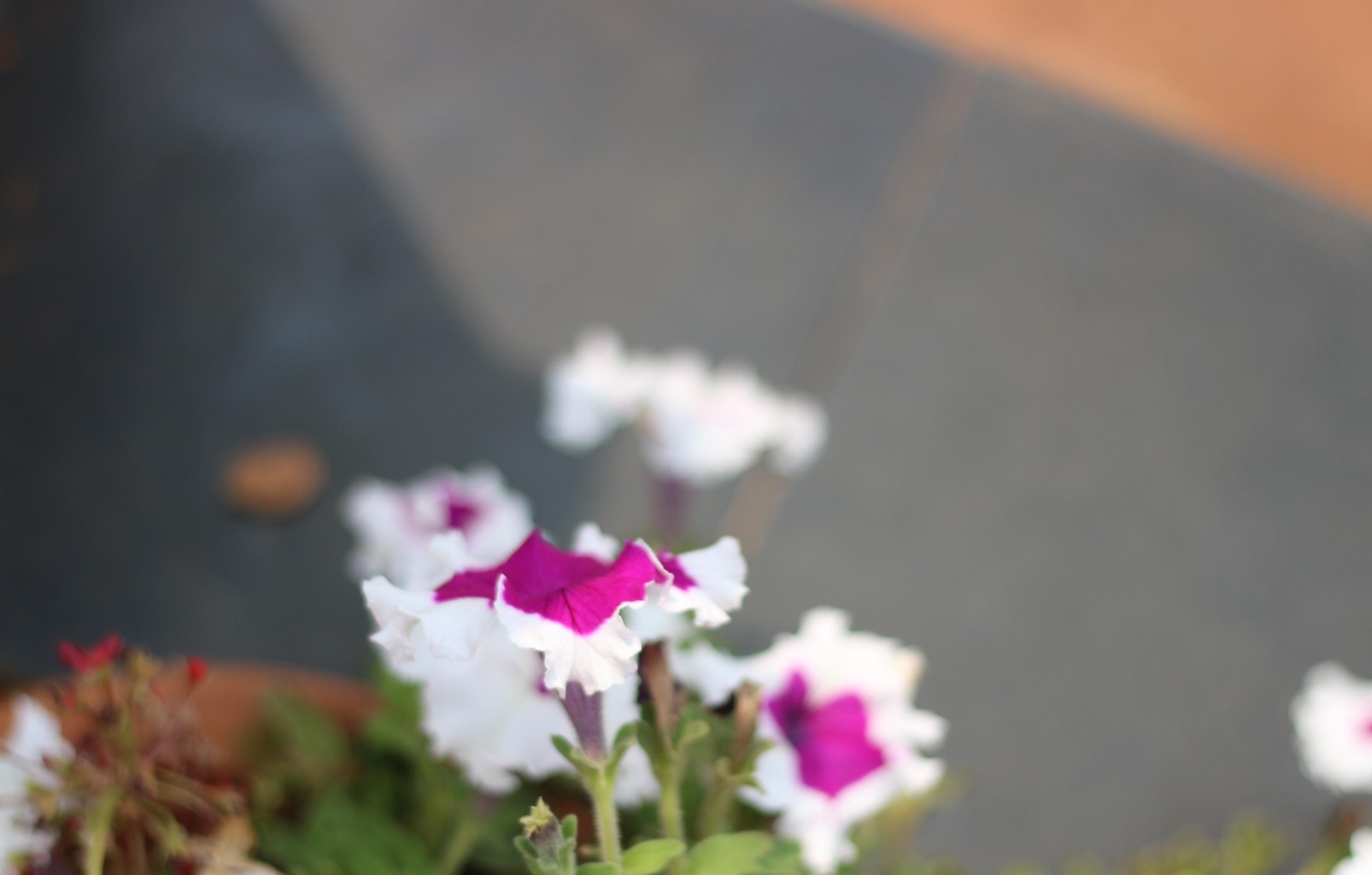} 
    \includegraphics[width=0.31\linewidth,height=0.2\linewidth]{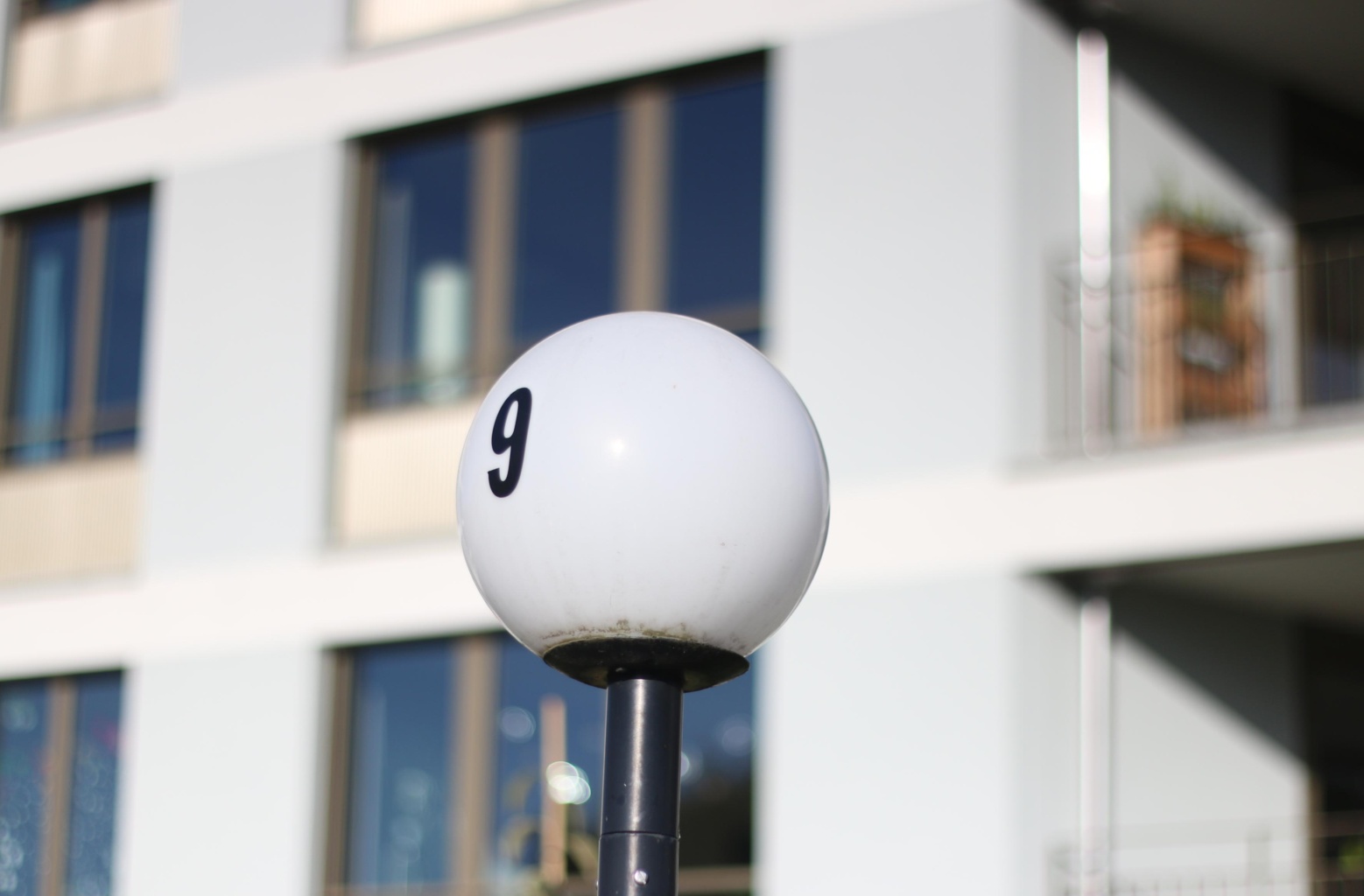} 
    \includegraphics[width=0.31\linewidth,height=0.2\linewidth]{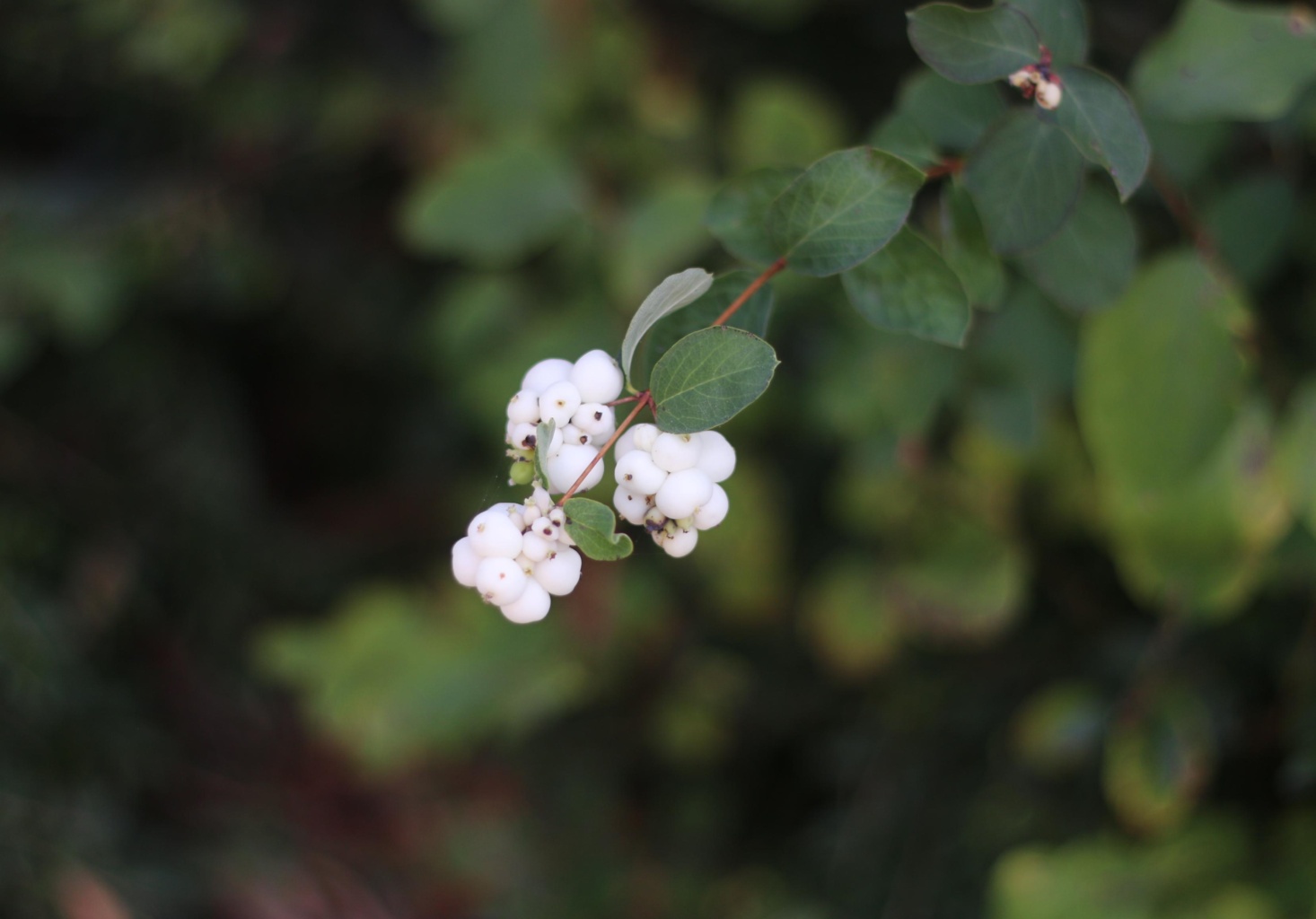}
    }\hfill
    \end{minipage}
    
    \vspace{-.5em}
    \caption{Examples from the original EBB~\cite{ignatov2020aim} dataset that shows misalignment. In the first two example, the lens blur image has clearly more exposure than the all-in-focus image; In the second and the third example, the images show some black edges near the image border, which is caused by image registration.}
    \label{fig:sampleEBB}
    \vspace{-.5em}
\end{figure}

EBB Val294 is an established subset~\cite{yang2023bokehornot,Mandl2024NeuralBokeh} of the EBB! dataset~\cite{ignatov2020aim}. It is composed of image pairs of wide and shallow depth of field captured by a DSLR camera. 
Though image registration is already performed~\cite{ignatov2020aim}, there are many cases where the ground truth deviates from the input in terms of global exposure level.
As shown in \cref{fig:sampleEBB}, the all-in-focus image is obviously darker in the first two examples, and shows black edges near the image edge, which is caused by image registration. These artifacts combined makes the metrics of pixel-wise correspondence less persuasive in the original dataset.

For a more informed comparison, we test the performance of the images by comparing them to the original EBB Val294~\cite{ignatov2020aim} dataset, and the results are shown in \cref{tab:quanti_EBB_original}. Note that as the apertures used in the EBB~\cite{ignatov2020aim} dataset are unknown, we find the optimal aperture by binary searching, similar to the approach taken by previous methods~\cite{peng2022bokehme,peng2022mpib}. As the quantitative metrics in the paper are calculated on the exposure-aligned EBB Val294~\cite{ignatov2020aim} dataset, we first report the quantitative performance with the same aperture as the paper in the left columns of \cref{tab:quanti_EBB_original}. Then we search for the optimal aperture on the original EBB Val294~\cite{ignatov2020aim} dataset, and list the metrics in the right columns of \cref{tab:quanti_EBB_original}.

BLB~\cite{peng2022bokehme}, a synthetic dataset proposed by BokehMe~\cite{peng2022bokehme}, consisting of 10 scenes, and 10 focal settings for each, rendered by Blender. The rendered bokeh can be significantly larger than real-world bokeh, so it can measure the accuracy of the underlying physics model. The most challenging level 5 is used for evaluation.

\textsc{SynBokeh300}, a new synthetic benchmark generated as described in the paper. It is composed of $300$ images, at $4$ levels of different lens blur strengths, the ground truth disparity map, focus distance, and the all-in-focus input images. The dataset excels others in terms of photorealism and diversity, and can be used to evaluate the performance in real-world scenarios. The results are listed in \cref{tab:quanti_synbokeh300}.

\section{Explanation on the EBB Dataset}

Note that given a method, we select the ``best'' result by selecting the aperture parameters that has the best SSIM performance. As the original EBB Val294~\cite{ignatov2020aim} dataset is not aligned in exposure level, it may hinder the optimal exposure level selection process. As observed in \cref{tab:quanti_EBB_original}, both of the two groups are tested on the original EBB Val294~\cite{ignatov2020aim} dataset, but the LPIPS~\cite{zhang2018unreasonable} performance of the group optimized on exposure-aligned EBB Val294~\cite{ignatov2020aim} dataset is obviously superior. As discussed in the paper, LPIPS is more sensitive to blurriness by design, and less sensitive to pixel-level difference. The efficacy of BokehDiff is further proved by the performance listed in \cref{tab:quanti_EBB_original}.

\begin{table}[t]
    \caption{Quantitative comparison on the \textsc{SynBokeh300} dataset. $\uparrow$ ($\downarrow$) indicates larger (smaller) values are better, and \textbf{bold} font indicates the best results. $^\bigstar$ denotes that the method is trained with the same dataset as BokehDiff.}\label{tab:quanti_synbokeh300}
\centering
\vspace{-.5em}
\begin{tabular}{lcccc}
\toprule
Method  & PSNR$\uparrow\!\!$ & SSIM$\uparrow\!\!$& DISTS$\downarrow\!\!$ & LPIPS$\downarrow\!\!$ \\
\midrule
DeepLens~\cite{deeplens2018} & 24.824 & 0.8121 & 0.1403 & 0.3218 \\
MPIB~\cite{peng2022mpib} &31.588 & 0.9465& 0.0499 & 0.1129\\ %
BokehMe~\cite{peng2022bokehme} & 33.357 & 0.9532 & 0.0459 & 0.1129
 \\ %
Dr.Bokeh~\cite{sheng2024dr} & 30.157 & 0.9532 & 0.0682 & 0.1504 \\
Restormer$^\bigstar$~\cite{zamir2022restormer} & 32.016 & 0.9220 & 0.0695 & 0.1695\\
BokehMe$^\bigstar$~\cite{peng2022bokehme} & 31.329 & 0.9403 & 0.0641 & 0.1231 \\
BokehDiff & \textbf{34.165} &  \textbf{0.9784} &  \textbf{0.0433} &  \textbf{0.1119} \\ \bottomrule
\end{tabular}
\vspace{-1em}
\end{table}

\section{Samples of the Data Synthesis Pipeline}
\begin{figure}[h]
    \centering
    \includegraphics[width=\linewidth]{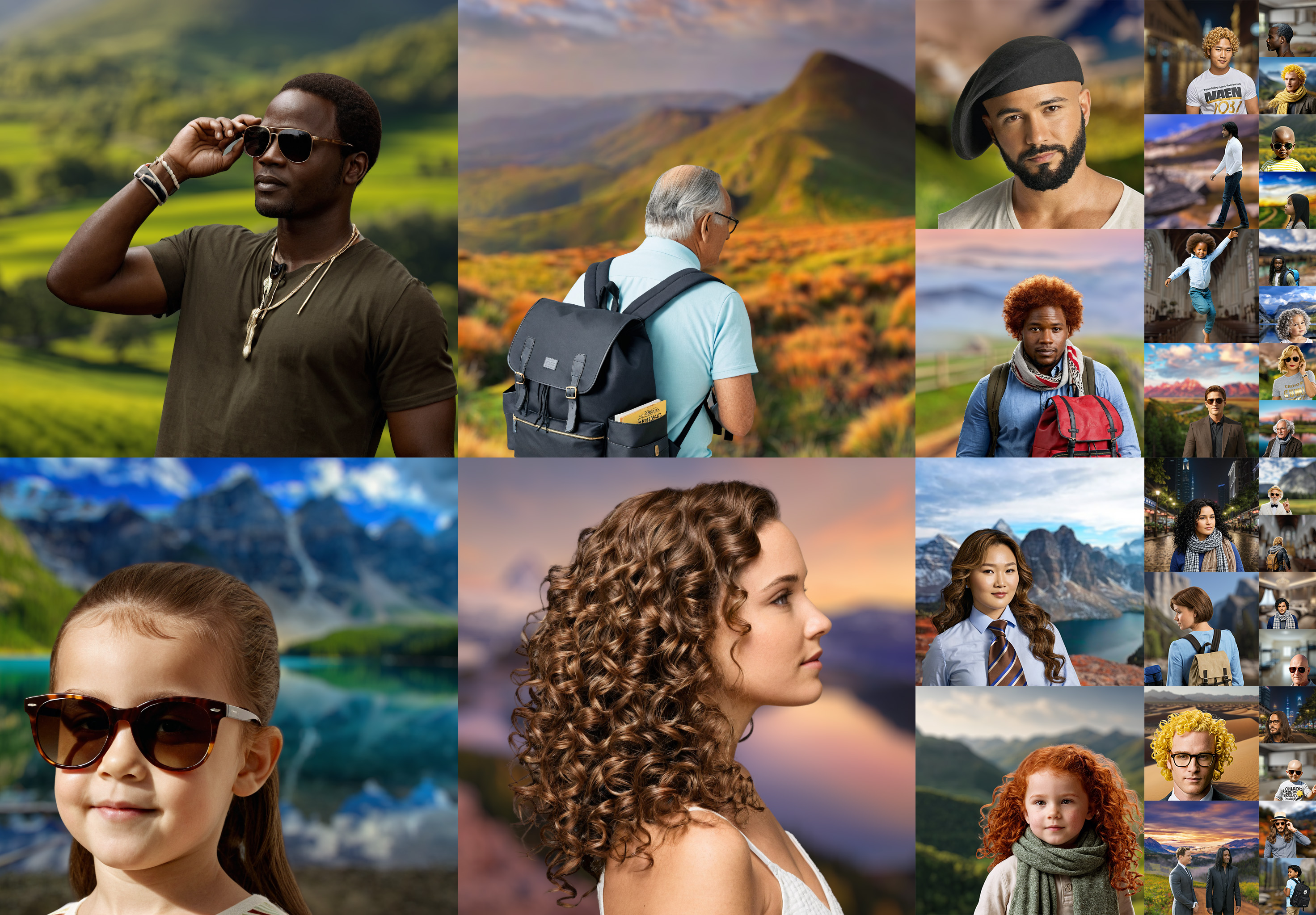}
    \caption{Diverse scenes sampled from the \textsc{SynBokeh300} dataset, to verify its photorealism and diversity.}
    \label{fig:example_dataset}
\end{figure}

\begin{figure*}[t]
\begin{minipage}[c]{0.185\linewidth}
    \subfloat[Disparity]{
    \includegraphics[width=\linewidth]{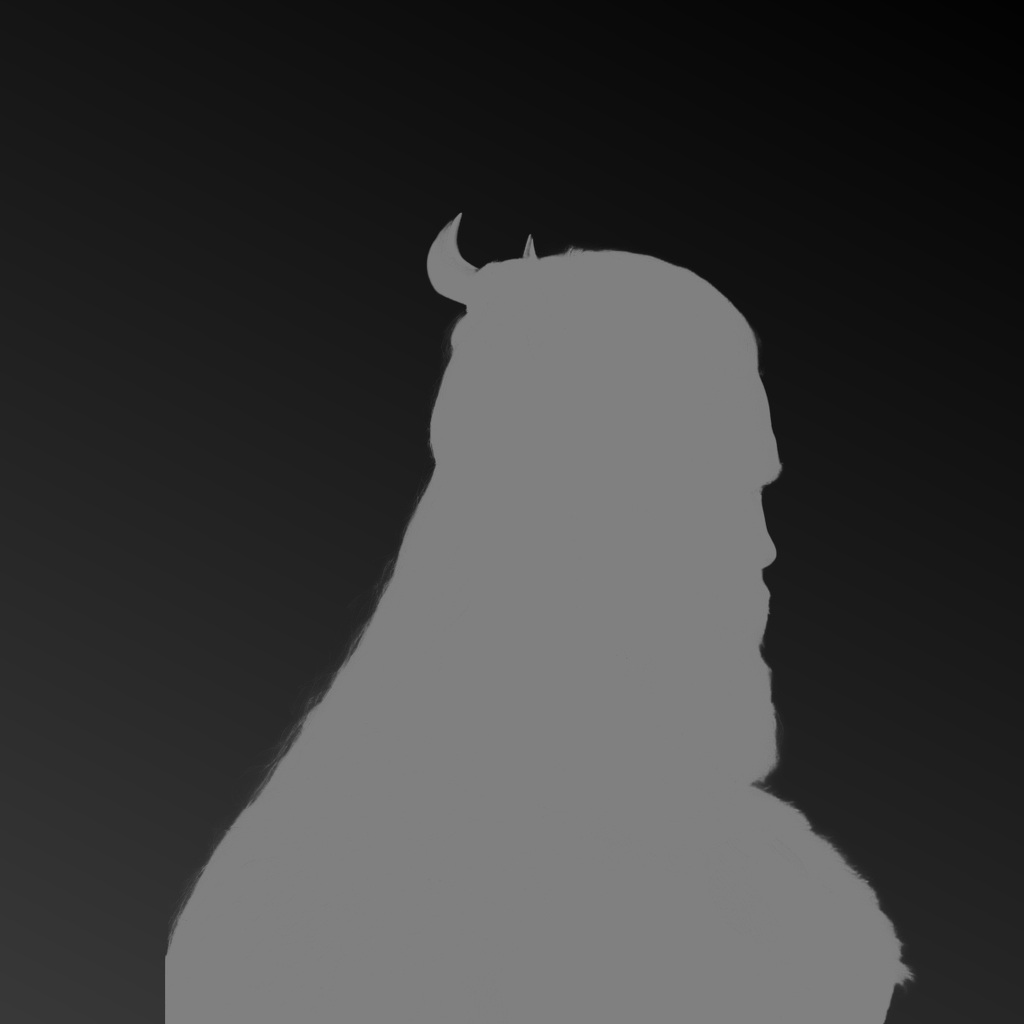}
    }\\
    \subfloat[All-in-focus]{
    \includegraphics[width=\linewidth]{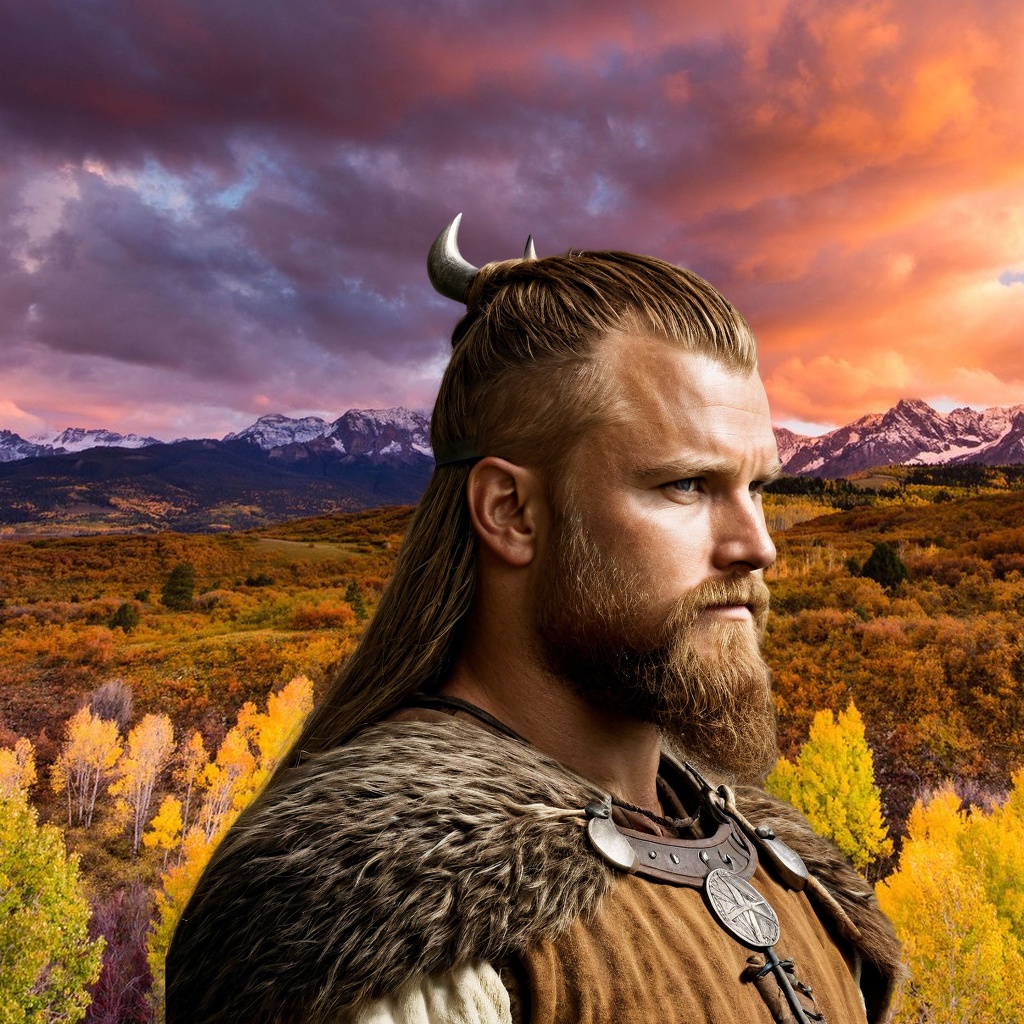}
    }
\end{minipage}\hfill
\begin{minipage}[c]{0.795\linewidth}
    \subfloat[Synthetic ground truth with different aperture parameters and different focus distance.]{
    \includegraphics[width=\linewidth]{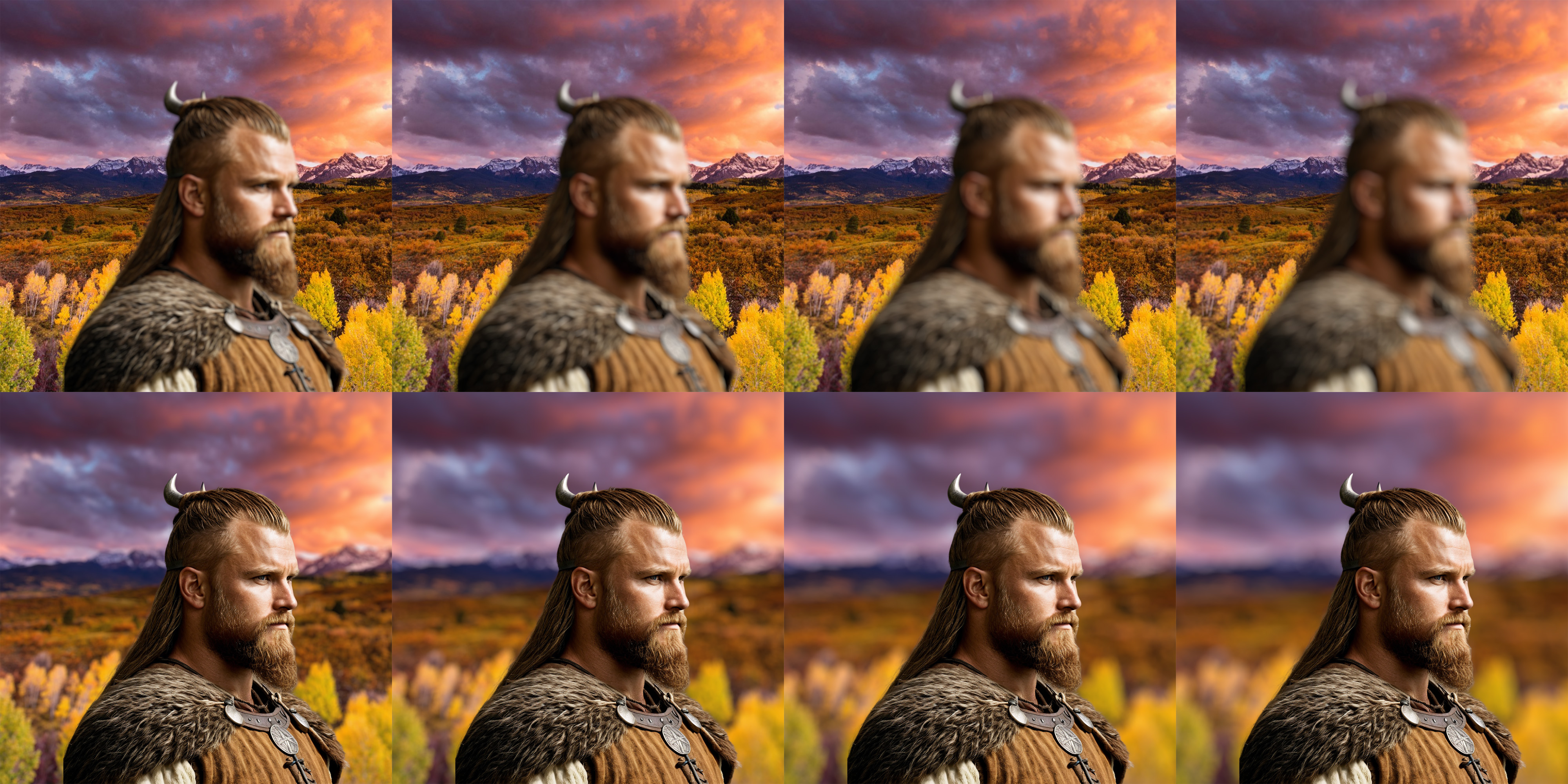}
    }
\end{minipage}\hfill
\vspace{-.5em}
\caption{An example of the synthetic data. With the mechanism described in the paper, we can get (a) disparity map, (b) all-in-focus image as input, and (c) synthetic ground truth images under different apertures and focus distance settings. Note that the first row in (c) is focused on the background, and the second is on the foreground, with aperture growing larger from left to right.}
\label{fig:example_dataset2}
\vspace{-.5em}
\end{figure*}

To demonstrate the results of the proposed data synthesis, we provide some samples of the \textsc{SynBokeh300} dataset. In \cref{fig:example_dataset}, the scene diversity of the \textsc{SynBokeh300} dataset is demonstrated sufficiently. In \cref{fig:example_dataset2}, we further show that the synthesis pipeline can generate both background-focused and foreground-focused images photo-realistically.

Note that the ability of BokehDiff to focus on any specific depth (as shown in the paper and later in \cref{sec:sup_more_results}) originates from the training data. By randomly placing the location and facing angles, the rendered data contains progressively blur with respect to the changing depth, as well as the different amount of blur caused by the disparity offset from the focal plane. 

\section{More Results}\label{sec:sup_more_results}

\begin{figure*}
    \centering
        \subfloat[Inputs]{
        \begin{minipage}[c]{0.1385\linewidth}
        \includegraphics[width=\linewidth]{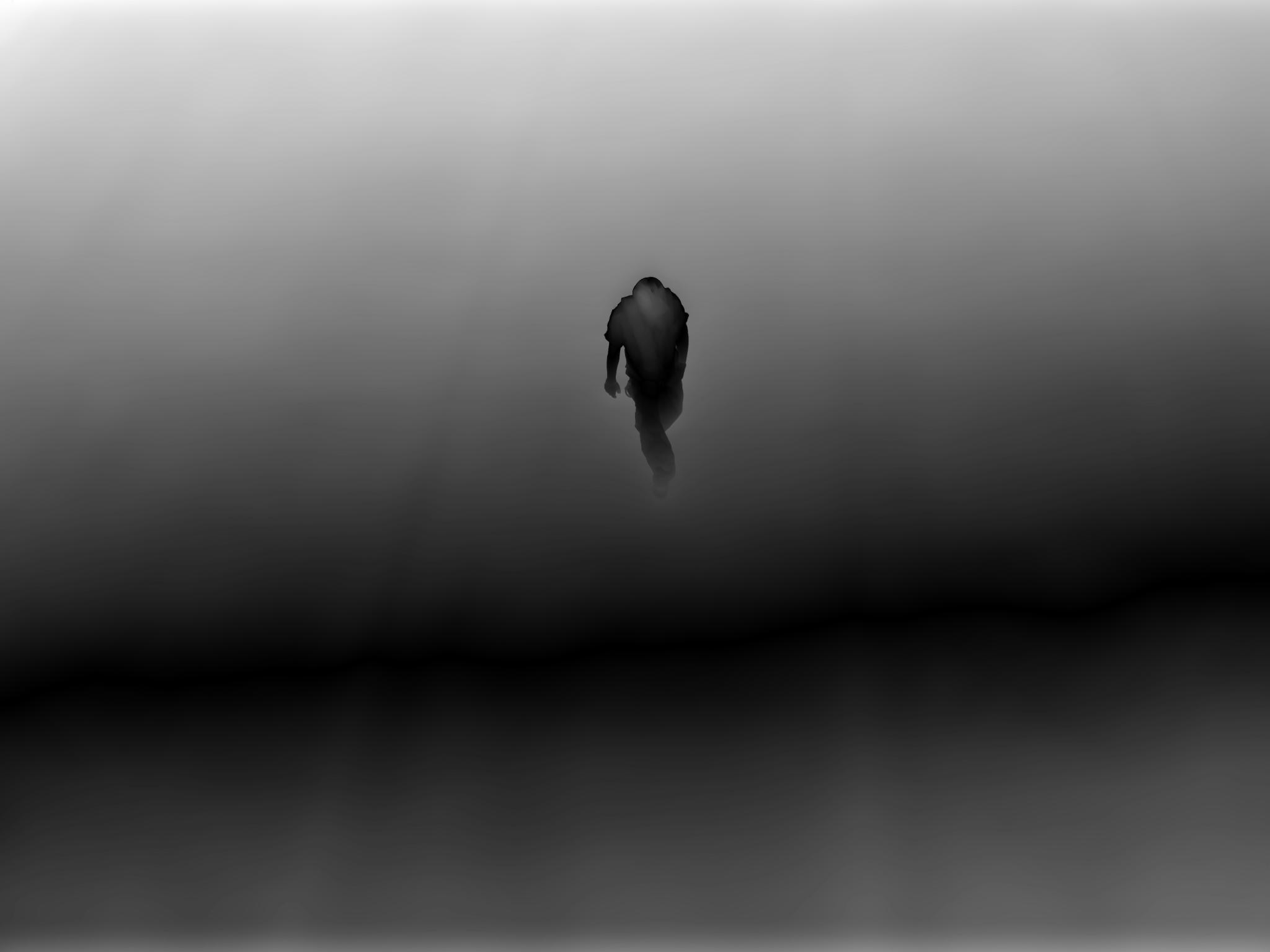}
        \includegraphics[width=\linewidth]{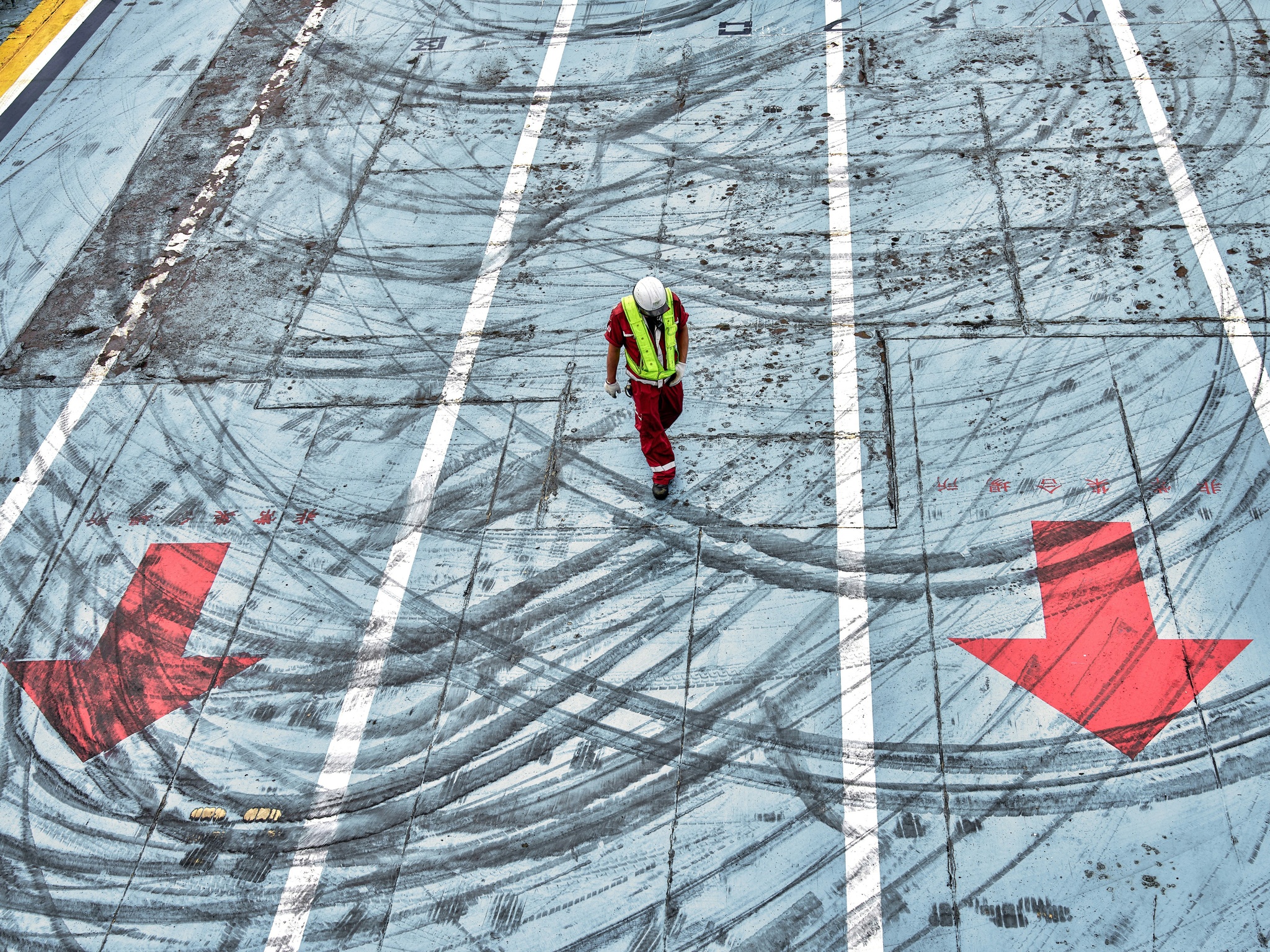}
        \includegraphics[width=\linewidth]{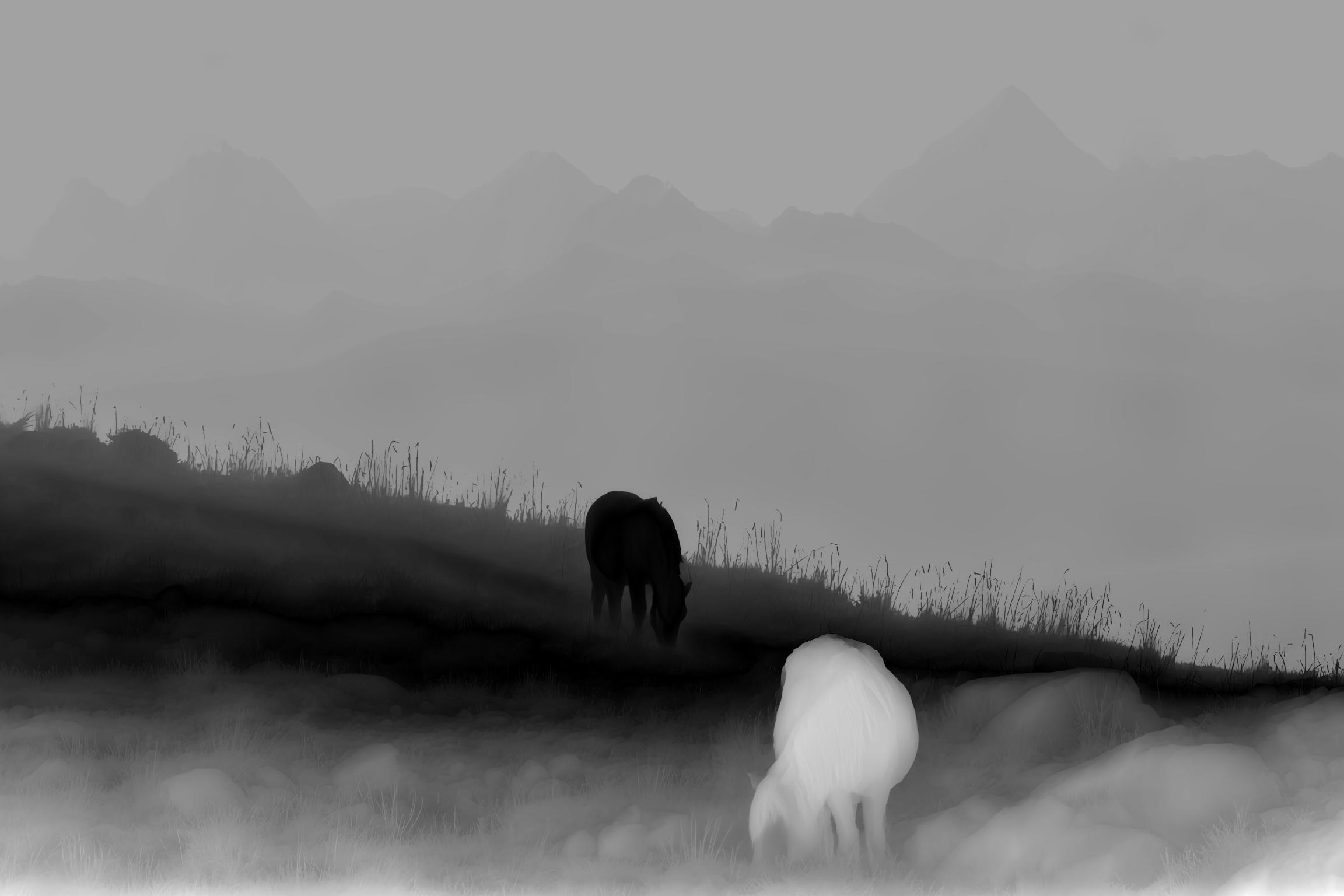}
        \includegraphics[width=\linewidth]{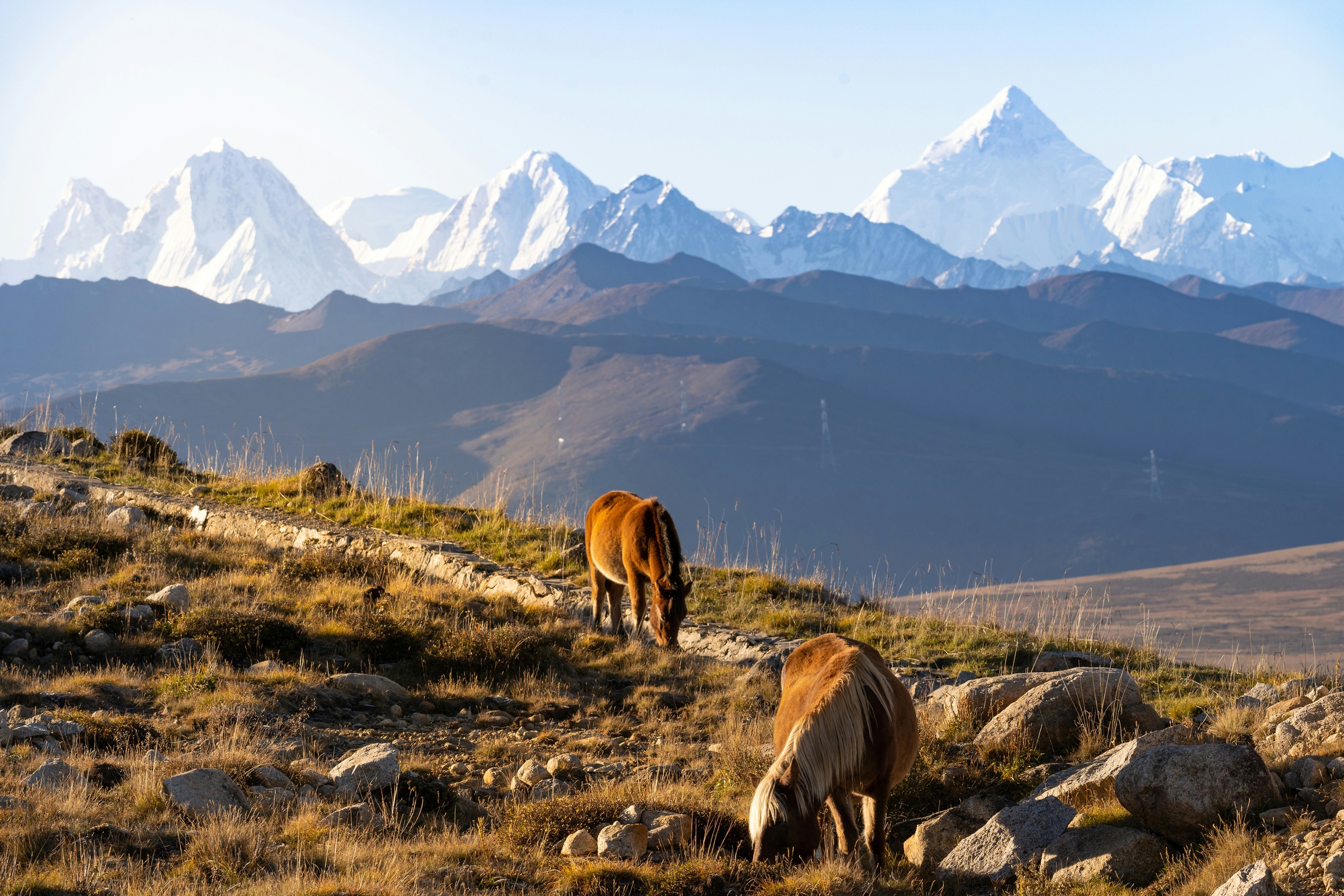} 
        \end{minipage}
        }
        \hspace{-0.6em}
        \subfloat[1x blur]{
        \begin{minipage}[c]{0.28\linewidth}
        \includegraphics[width=\linewidth]{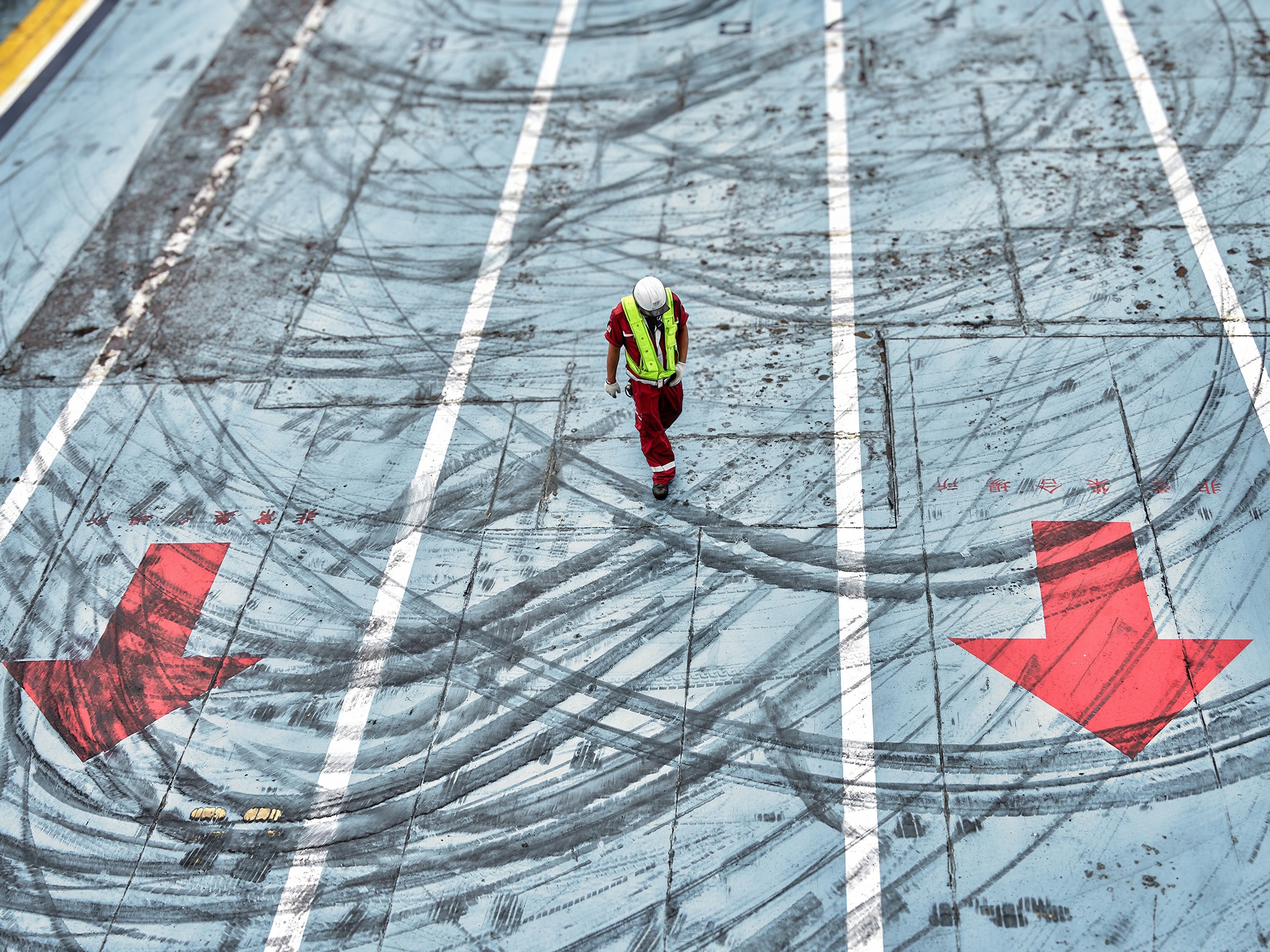}
        \includegraphics[width=\linewidth]{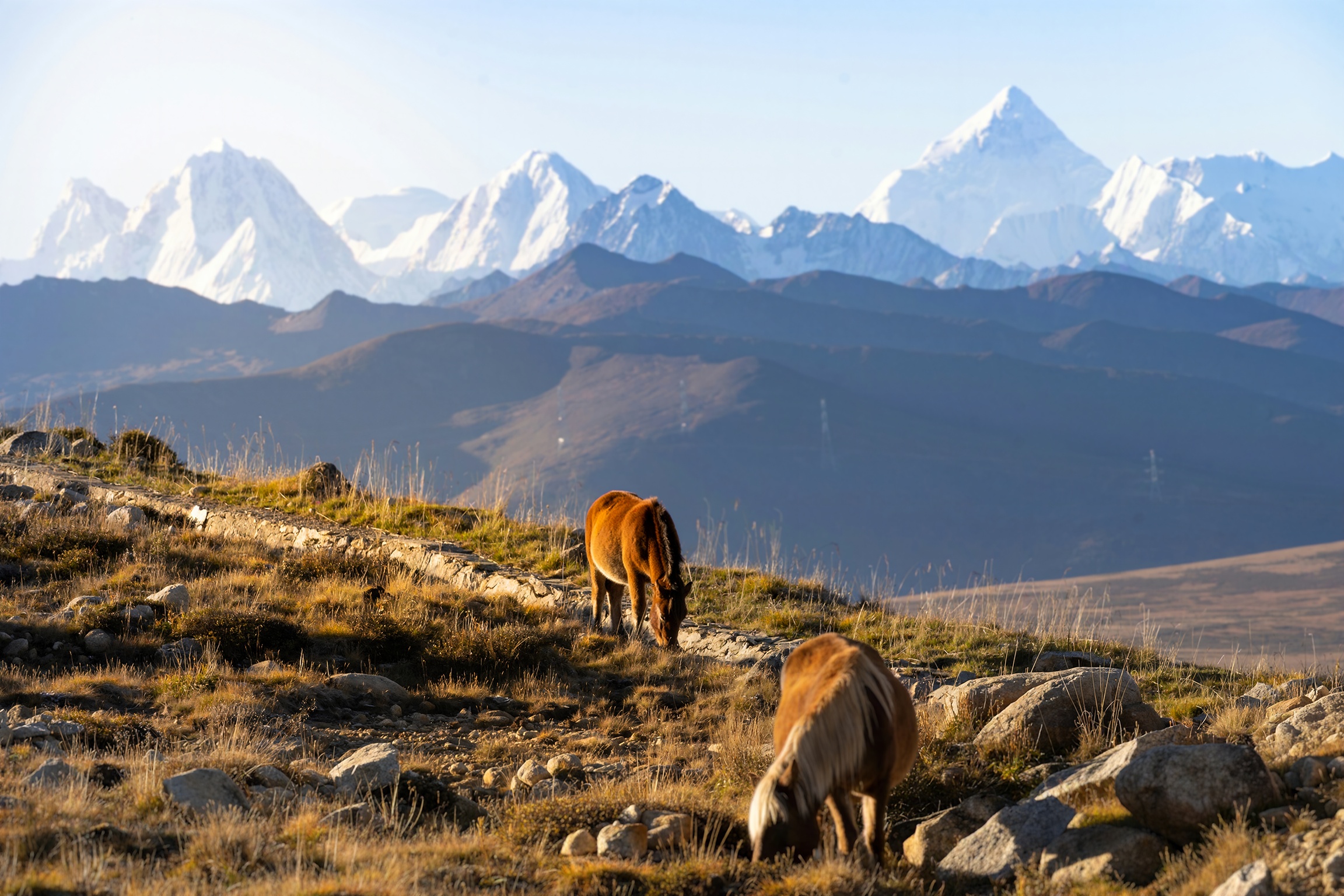} 
        \end{minipage}
        }\hspace{-0.6em}
        \subfloat[2x blur]{
        \begin{minipage}[c]{0.28\linewidth}
        \includegraphics[width=\linewidth]{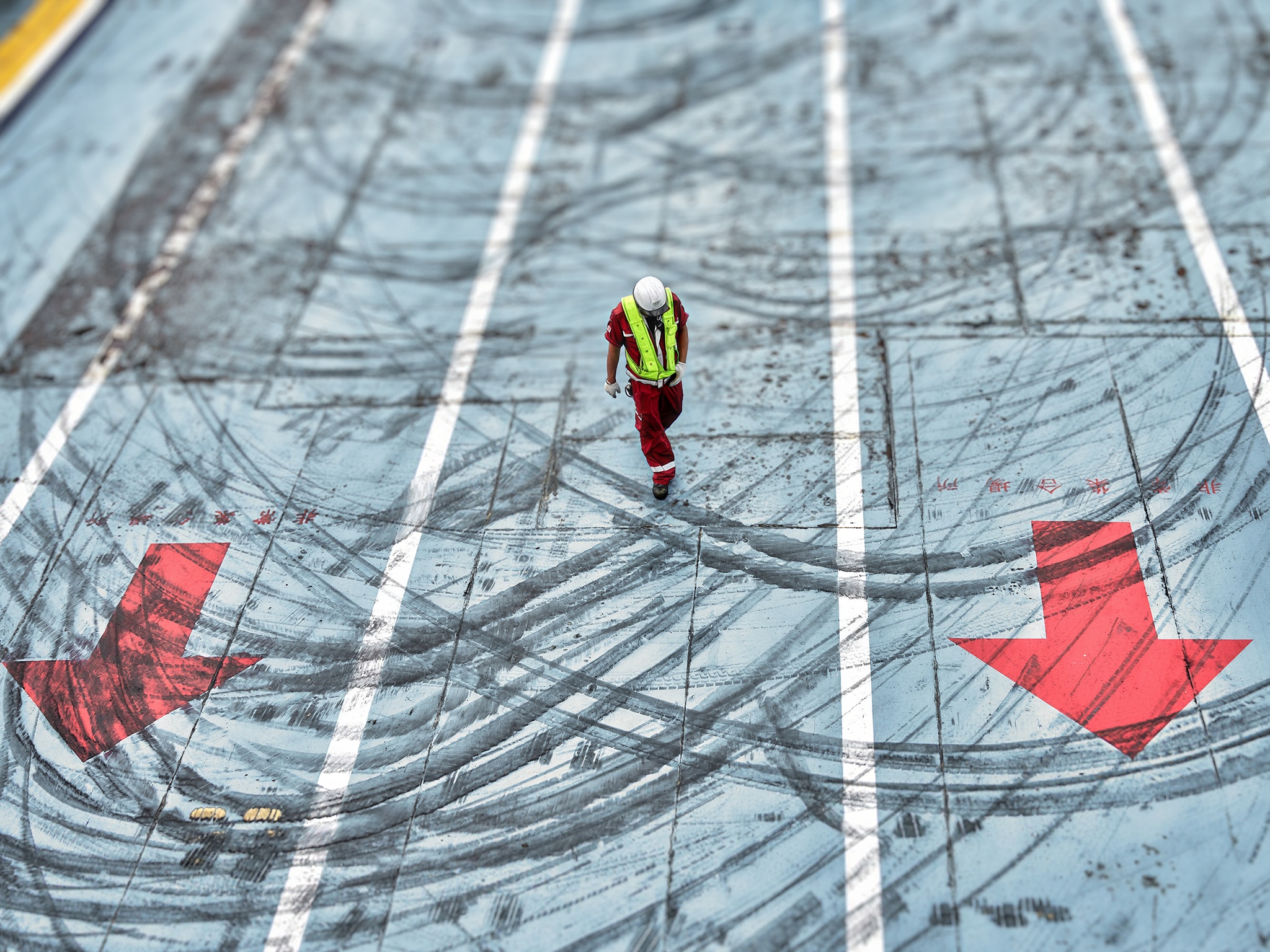}
        \includegraphics[width=\linewidth]{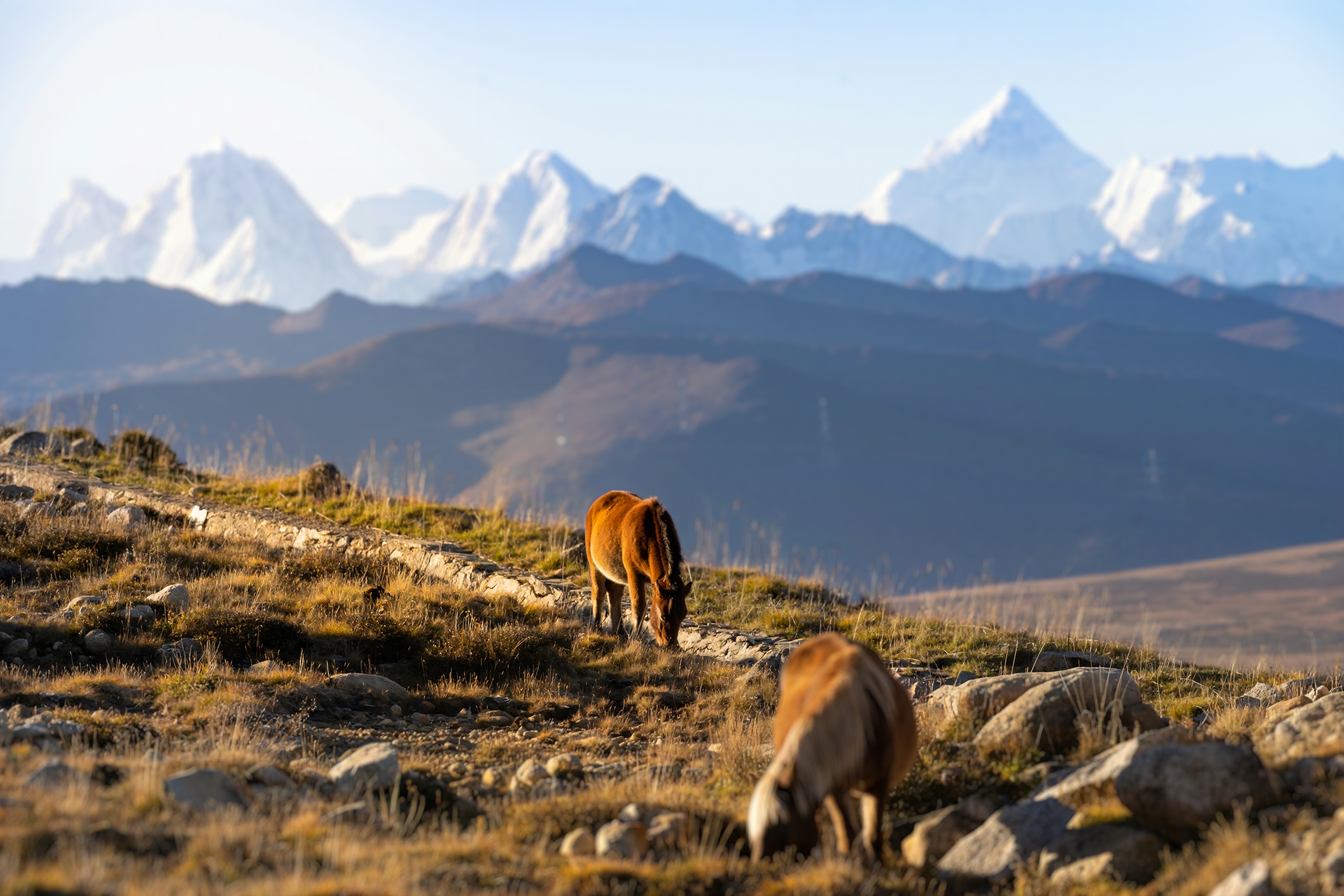}
        \end{minipage}
        }\hspace{-0.6em}
        \subfloat[3x blur]{
        \begin{minipage}[c]{0.28\linewidth}
        \includegraphics[width=\linewidth]{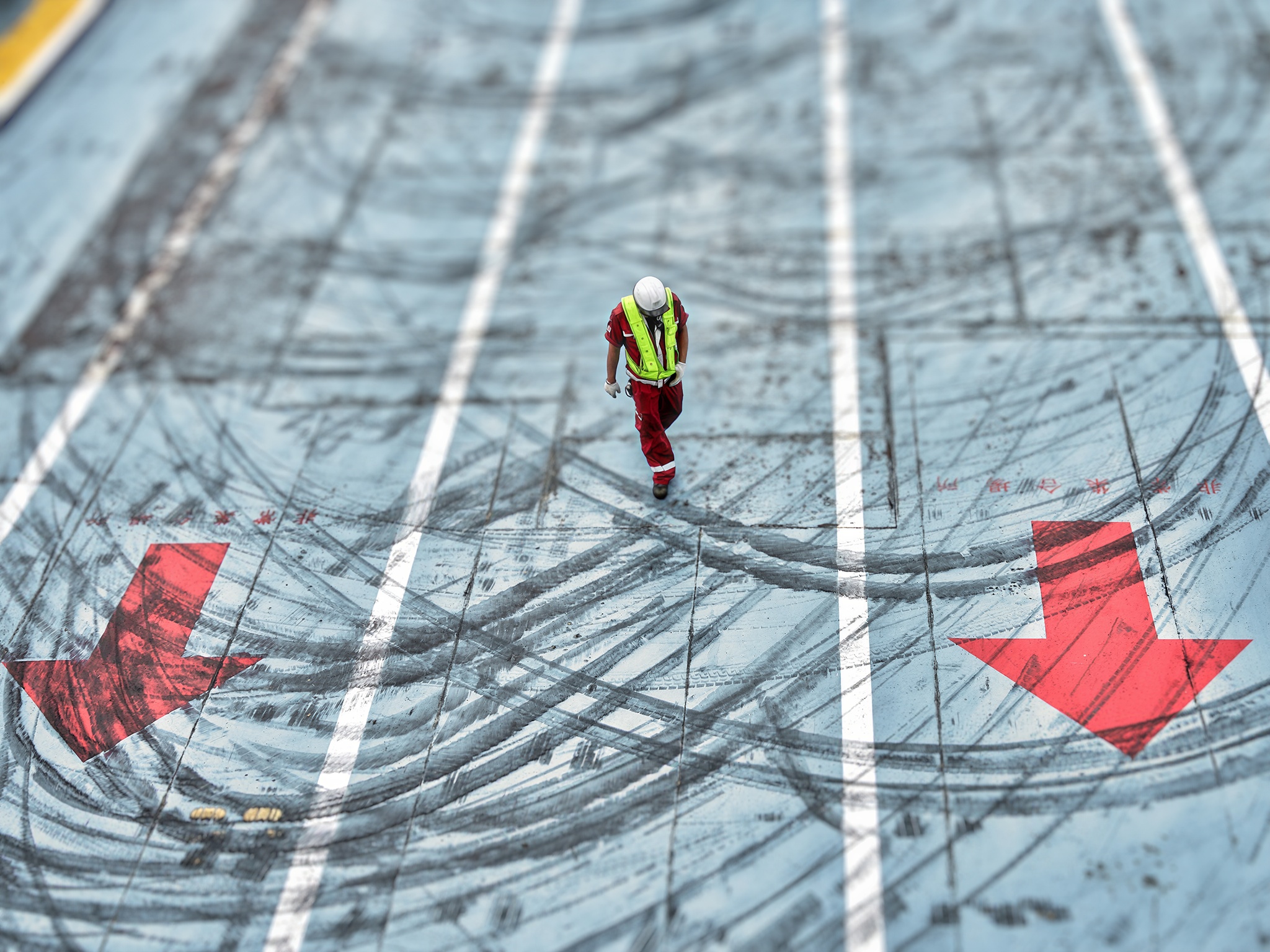}
        \includegraphics[width=\linewidth]{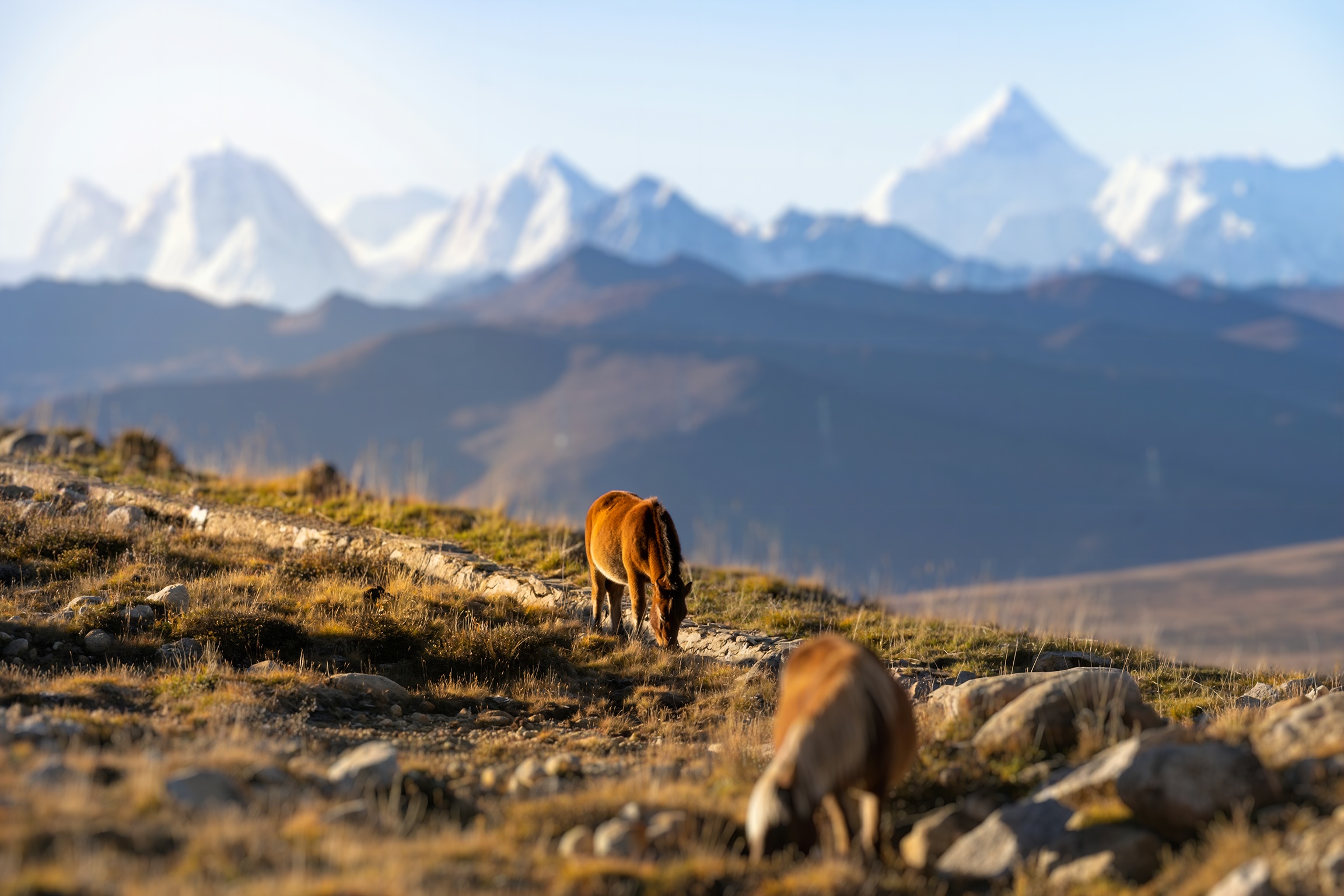}
        \end{minipage}
        }
    \begin{minipage}[c]{0.78\linewidth}
        
    \end{minipage}
    \caption{Given the defocus map and all-in-focus image shown in (a), we demonstrate the results of gradually increasing the aperture parameters, from (b) 1x blurriness to (c) 3x blurriness. Please zoom in for details.}
    \label{fig:increasing}
\end{figure*}
\subsection{Adjusting Aperture}
We first demonstrate the results of increasing the blurriness in \cref{fig:increasing}. In the first case, BokehDiff successfully creates the desired progressive blurriness, and in the second case, manages to blur both the foreground and the background that are off the focal plane. In both examples, BokehDiff is able to follow the underlying physics rules, and creates the right results at depth discontinuities. 

\subsection{Adjusting Focus Distance}
\begin{figure*}
    \centering
    \includegraphics[width=\linewidth]{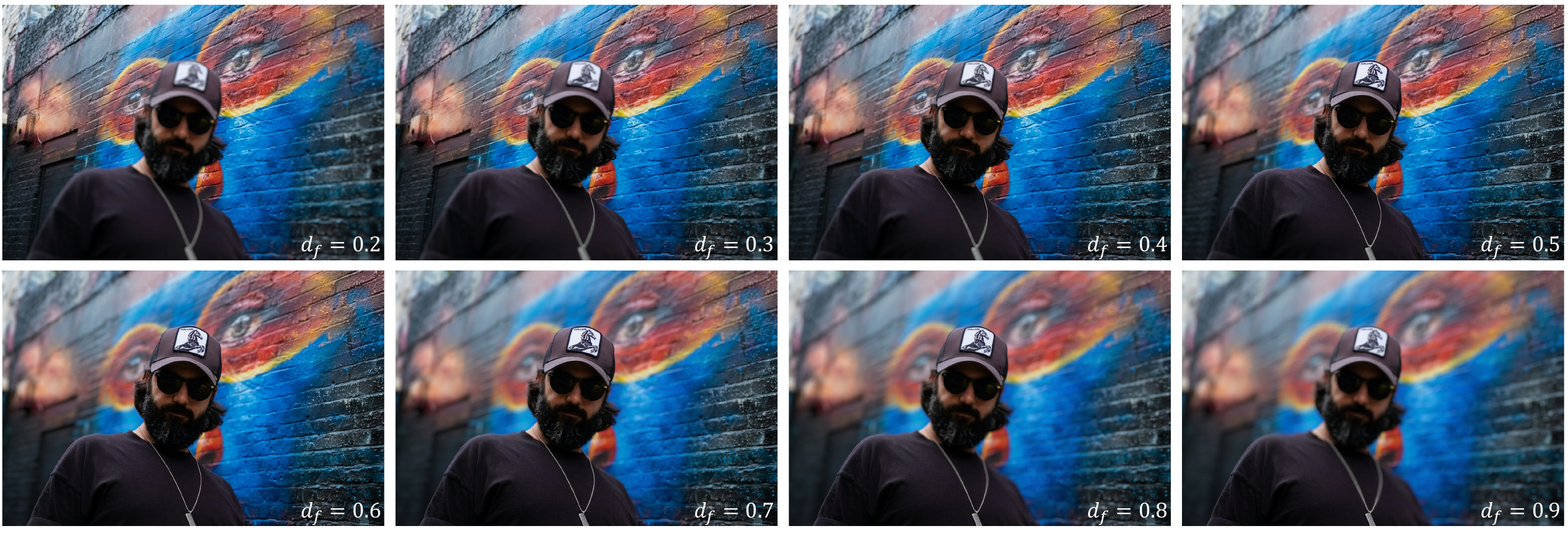}
    \caption{A synthetic focal stack of BokehDiff, given an all-in-focus image selected from the Unsplash~\cite{unsplash} dataset.}
    \label{fig:focal_supp}
\end{figure*}

We provide another example of changing focus distance in \cref{fig:focal_supp}. As the error is more subtle when the background and foreground are both out of focus, we mainly present the images that focuses on the foreground.

\subsection{Comparisons}
Here we provide some more comparisons of BokehDiff and the baselines, to further validate the efficacy of BokehDiff.

\begin{figure*}[h]
    \centering
    \vspace{-1em}
    \hspace*{\fill}
    \rotatebox{90}{\quad\quad\quad  \sffamily Defocus}\hfill
    \includegraphics[height=8.8em]{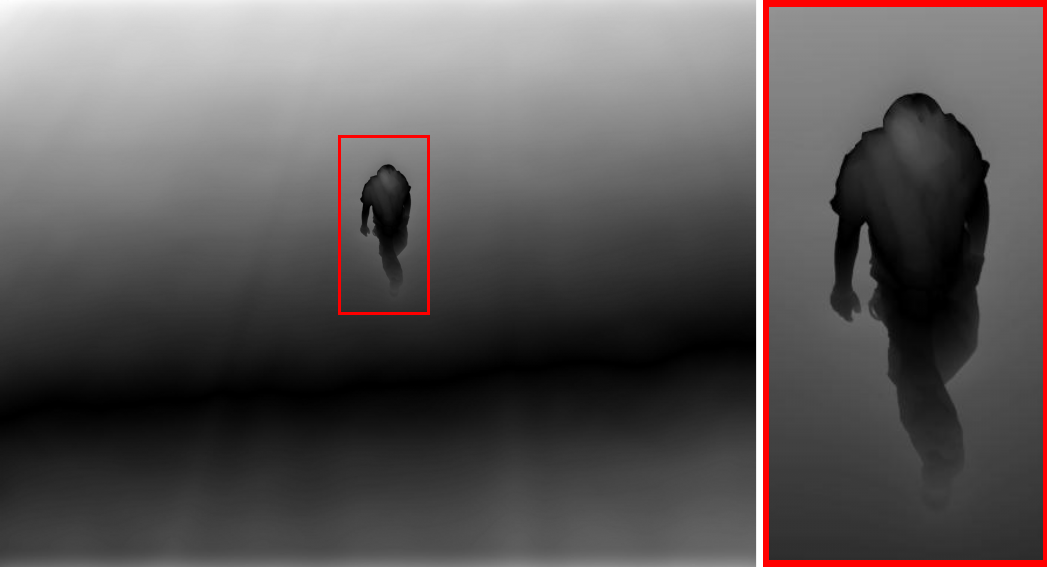}\hfill
    \includegraphics[height=8.8em]{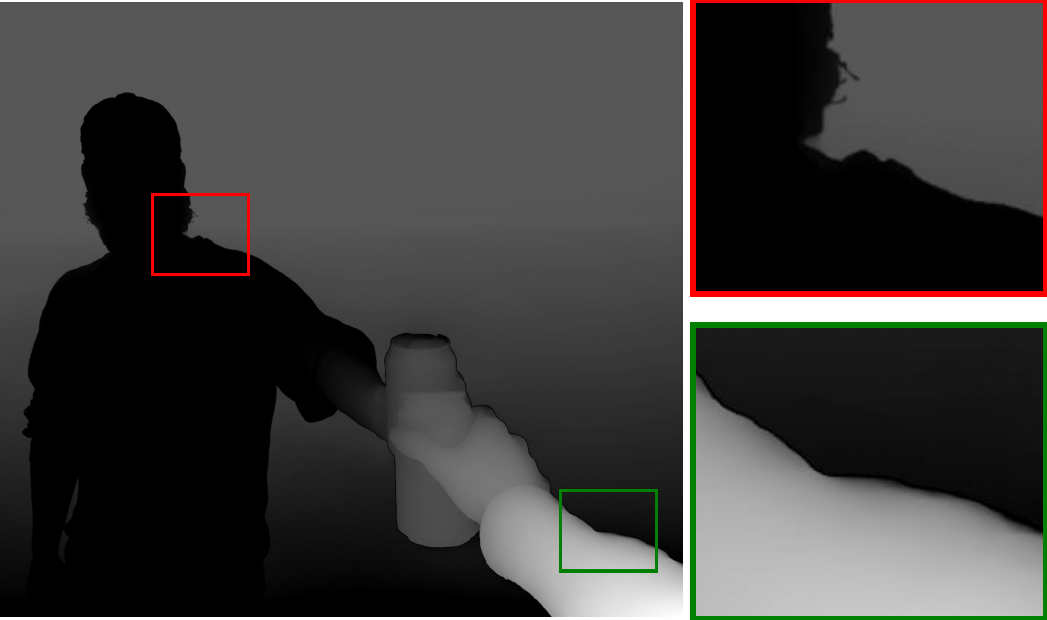}\hfill
    \includegraphics[height=8.8em]{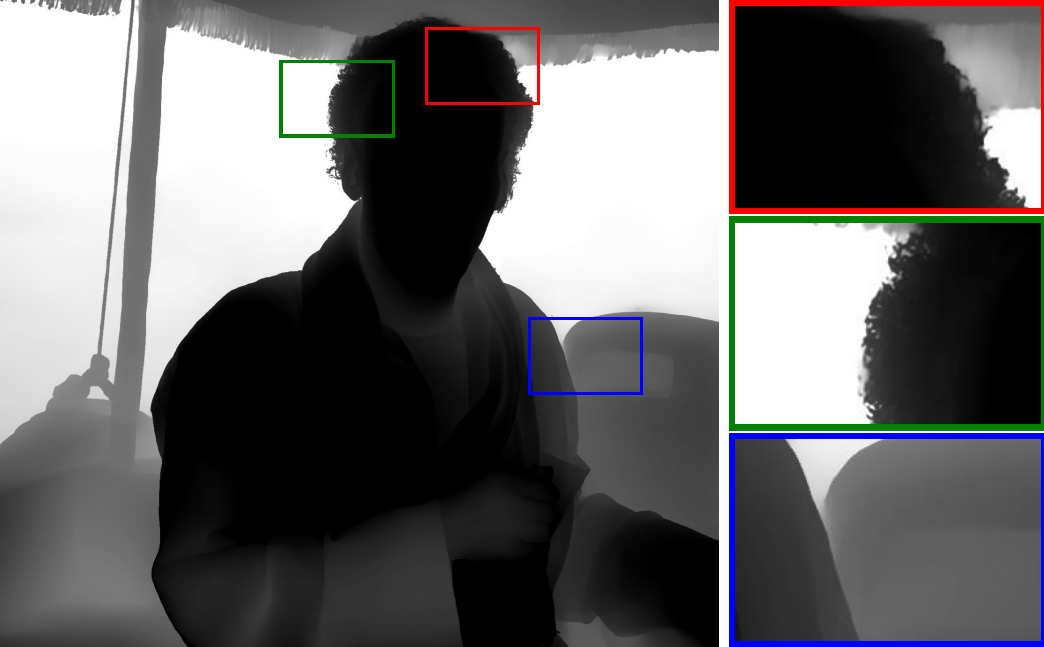}\hfill
    \hspace*{\fill}
    
\hspace*{\fill}
    \rotatebox{90}{\quad\quad \  \sffamily All-in-focus}\hfill
    \includegraphics[height=8.8em]{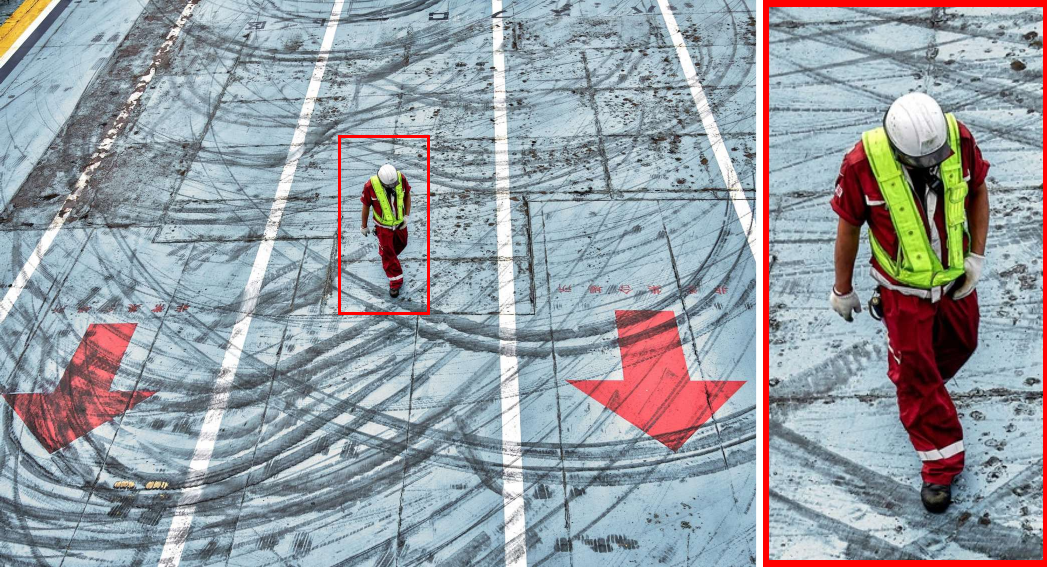}\hfill
    \includegraphics[height=8.8em]{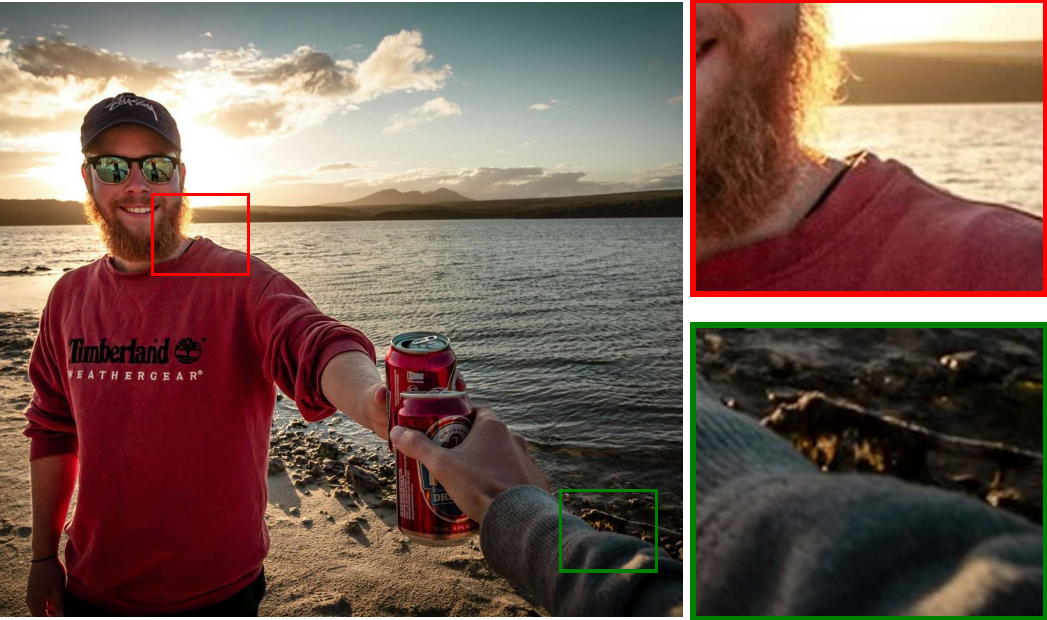}\hfill
    \includegraphics[height=8.8em]{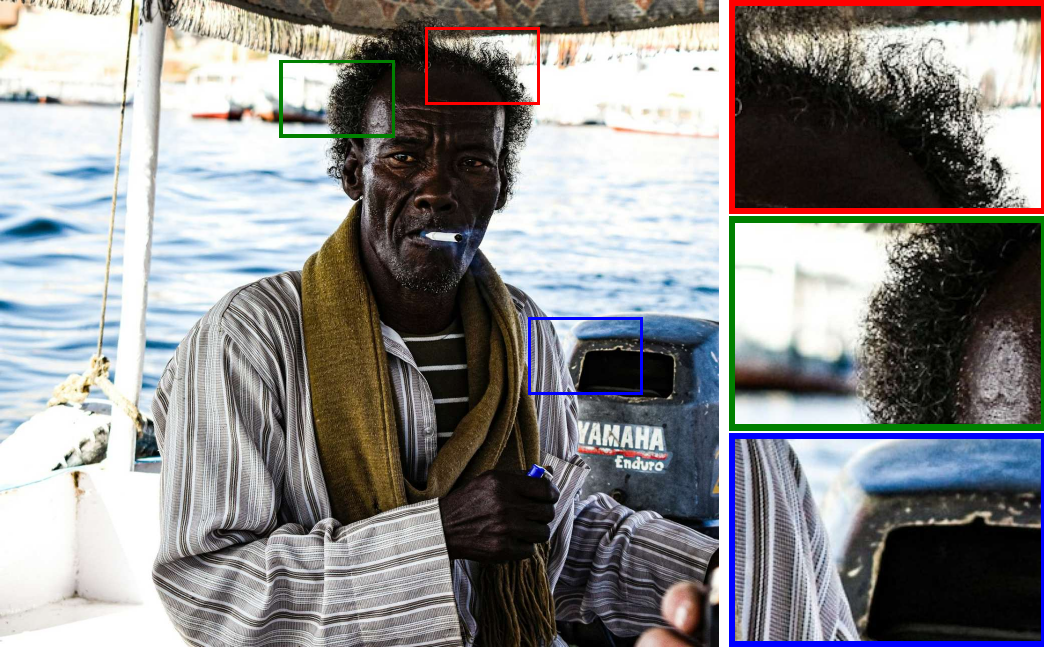}\hfill
    \hspace*{\fill}
    
    \hspace*{\fill}
    \rotatebox{90}{\quad\quad\ \ \sffamily BokehDiff}\hfill
    \includegraphics[height=8.8em]{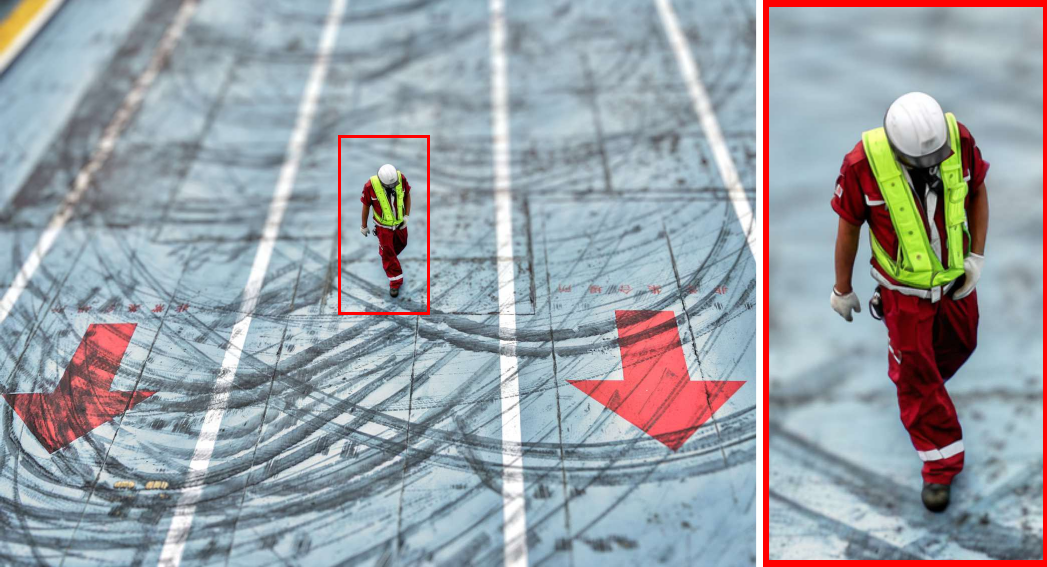}\hfill
    \includegraphics[height=8.8em]{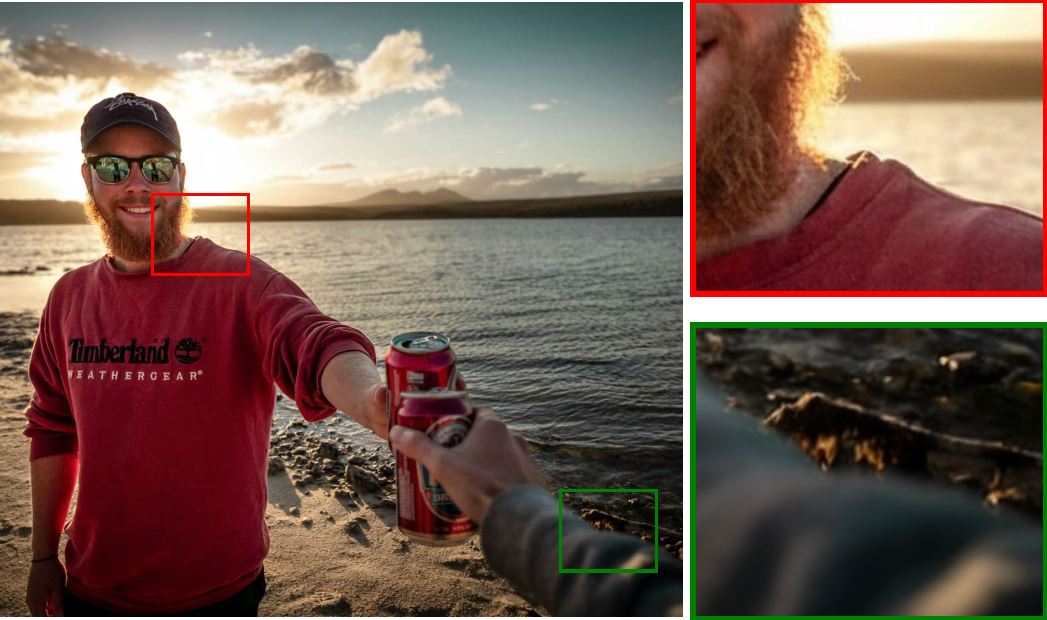}\hfill
    \includegraphics[height=8.8em]{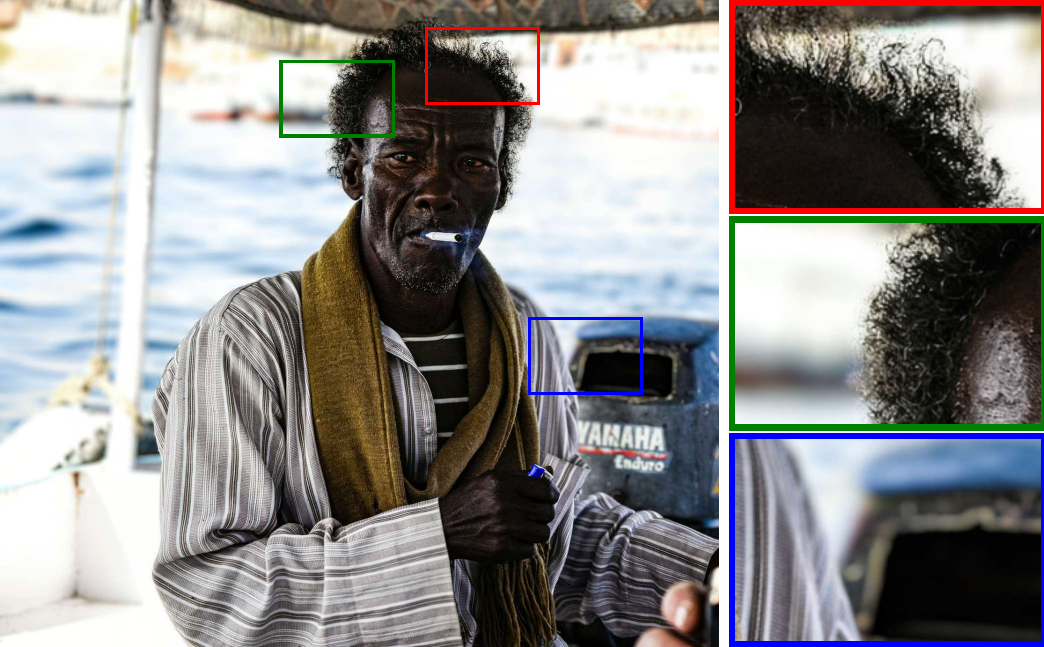}\hfill
    \hspace*{\fill}

    \hspace*{\fill}
    \rotatebox{90}{\quad\quad\  \sffamily BokehMe}\hfill
    \includegraphics[height=8.8em]{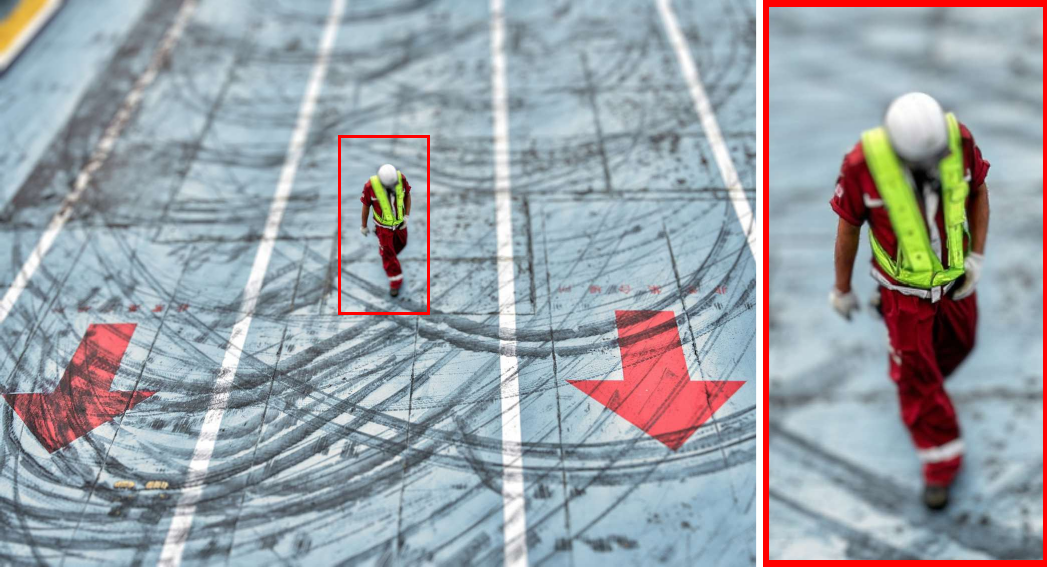}\hfill
    \includegraphics[height=8.8em]{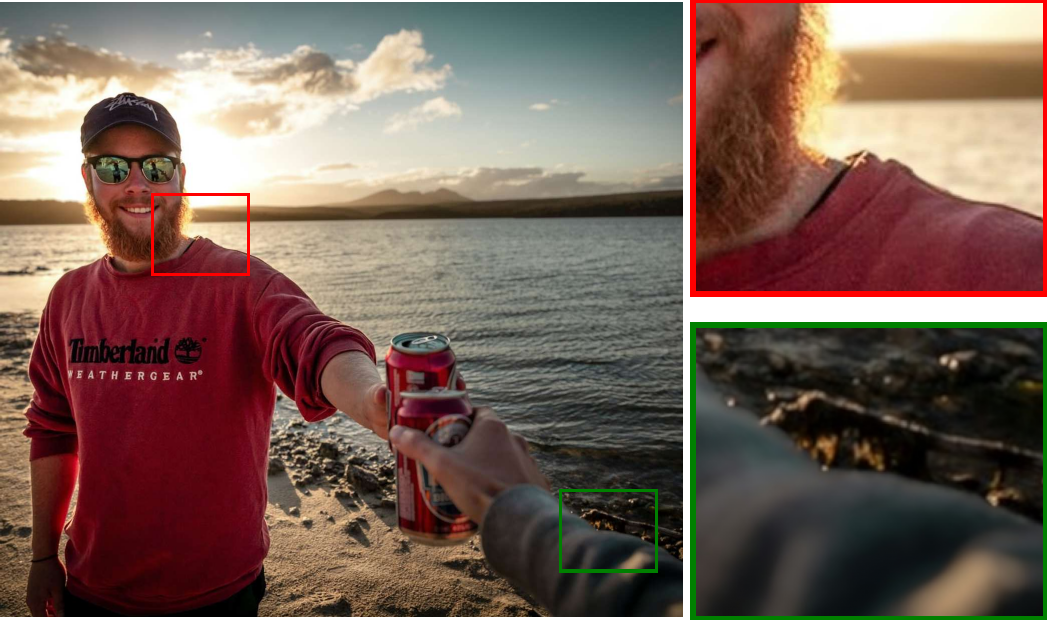}\hfill
    \includegraphics[height=8.8em]{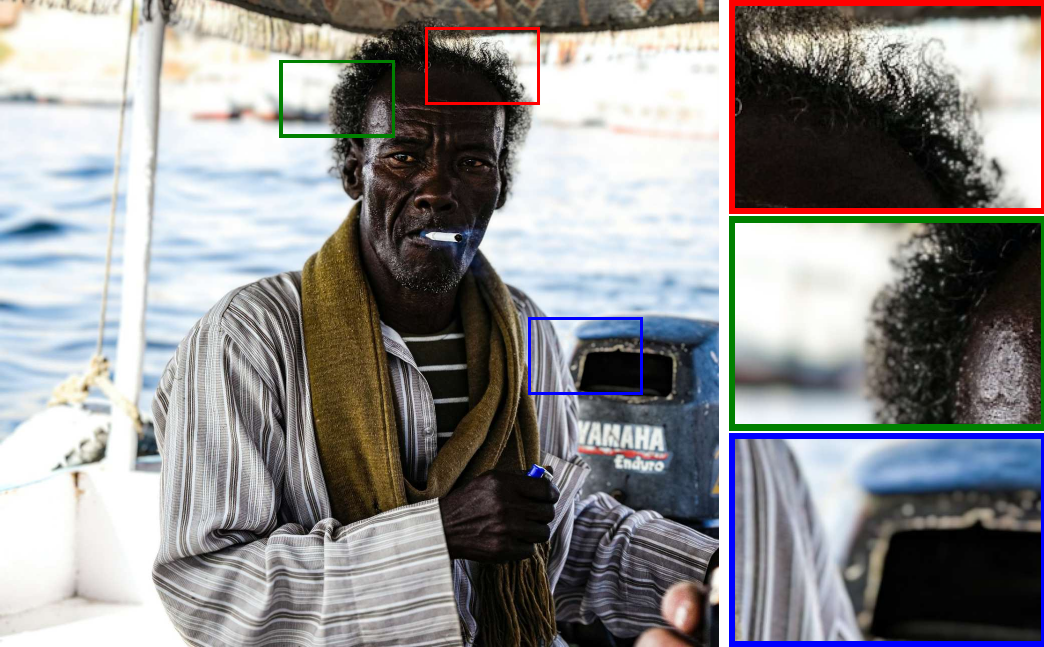}\hfill
    \hspace*{\fill}

     \hspace*{\fill}
    \rotatebox{90}{\quad\quad\quad\ \sffamily MPIB}\hfill
    \includegraphics[height=8.8em]{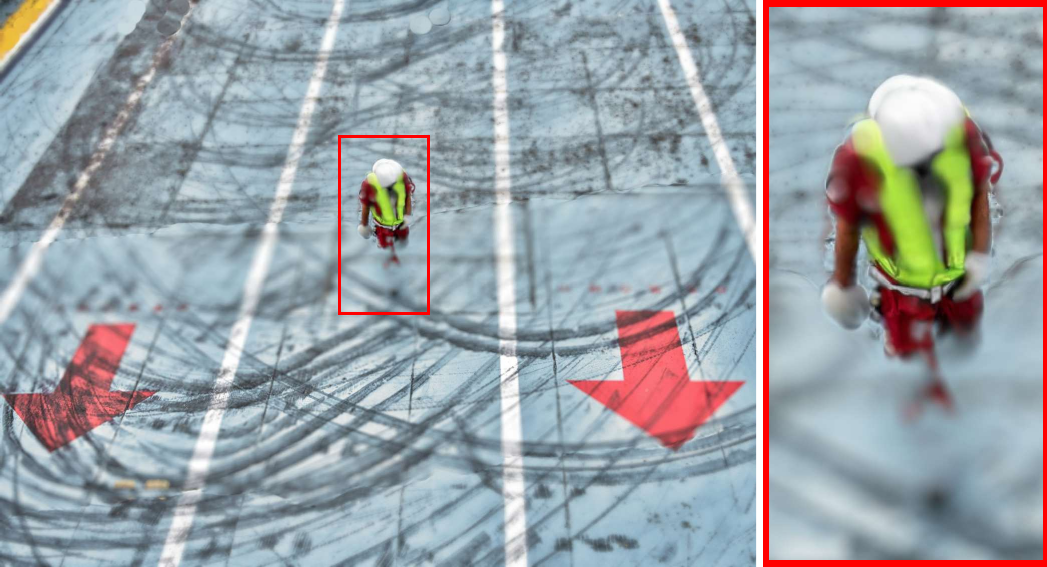}\hfill
    \includegraphics[height=8.8em]{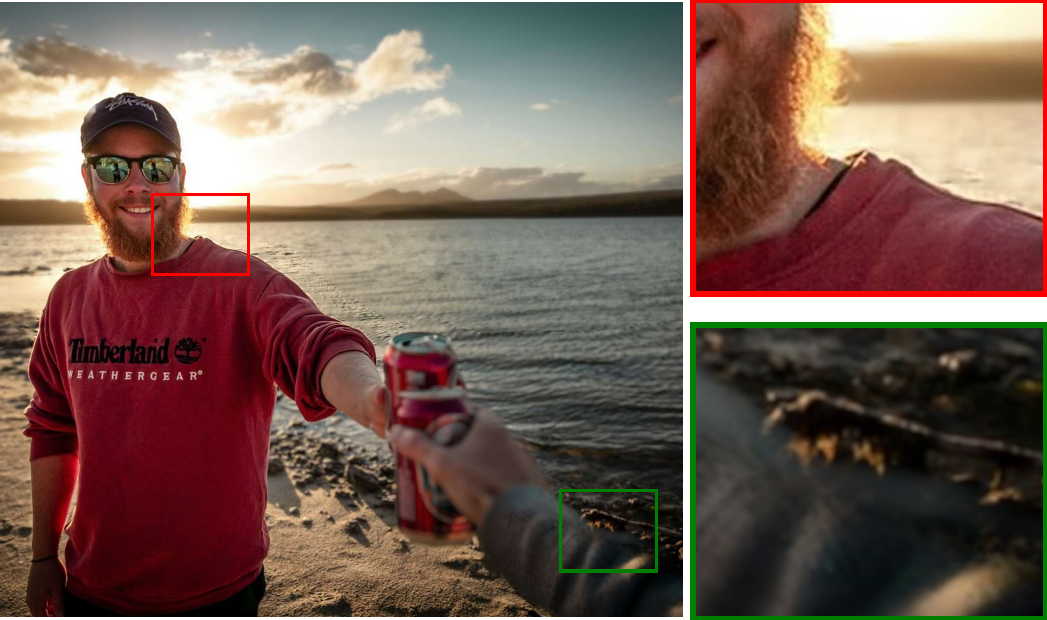}\hfill
    \includegraphics[height=8.8em]{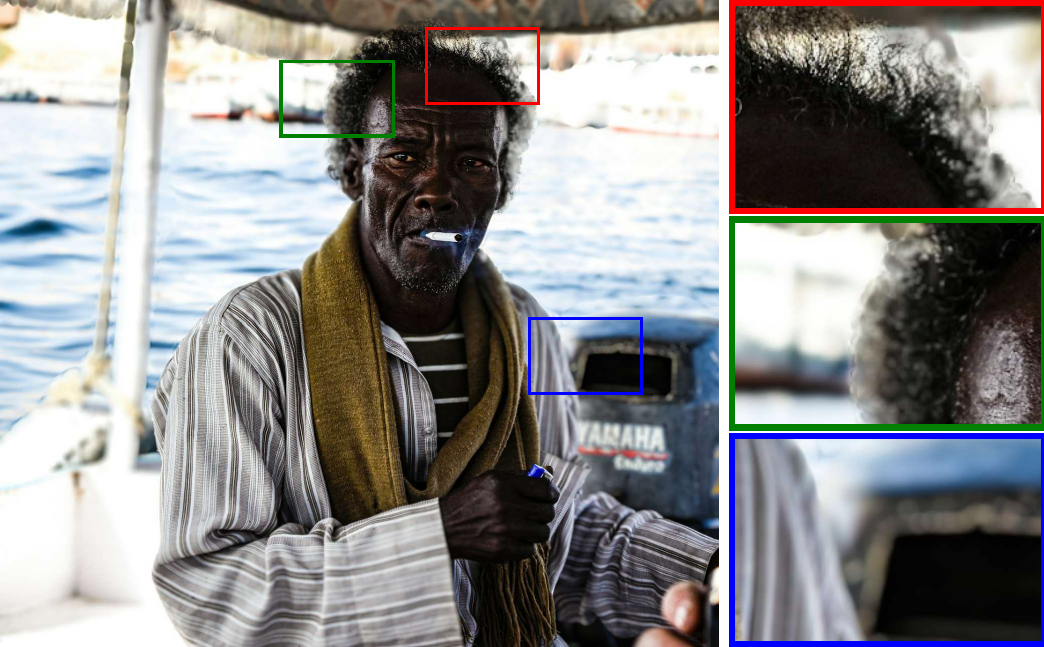}\hfill
    \hspace*{\fill}
    
    \hspace*{\fill}
    \rotatebox{90}{\quad\quad\ \ \sffamily  Dr.~Bokeh}\hfill
    \includegraphics[height=8.8em]{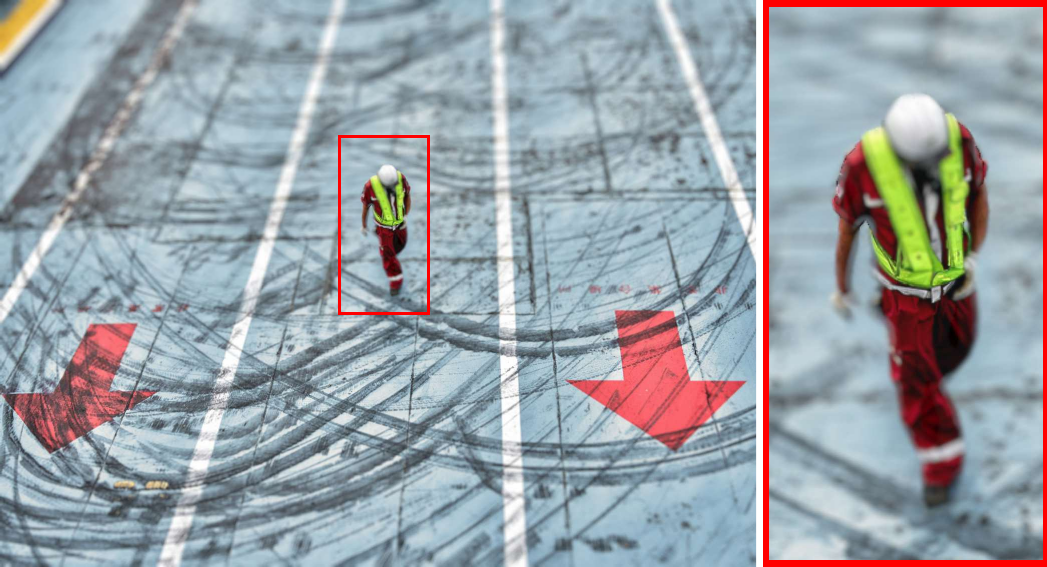}\hfill
    \includegraphics[height=8.8em]{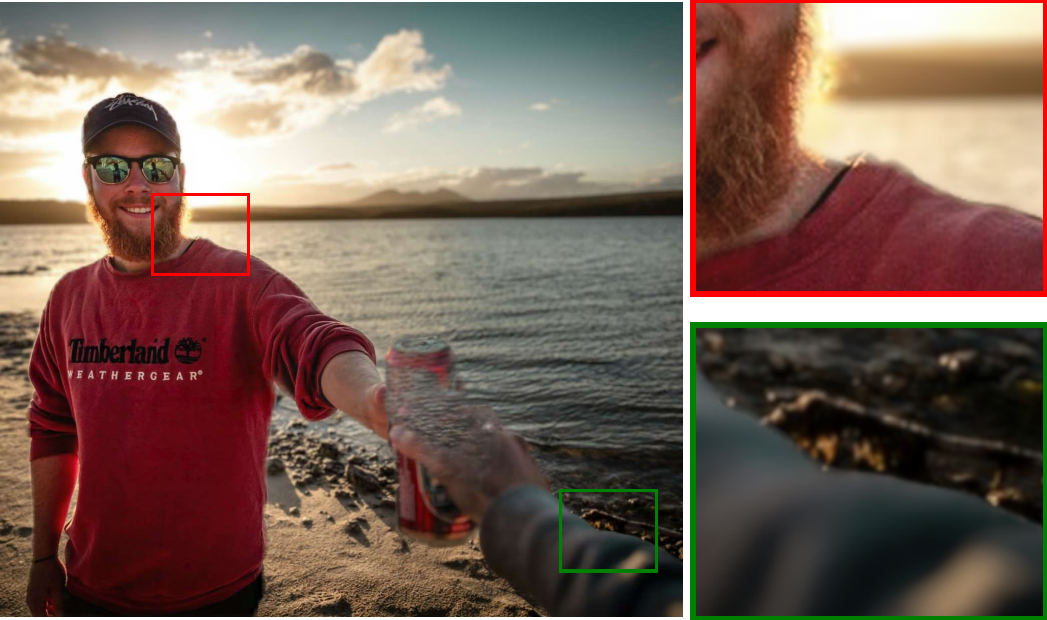}\hfill
    \includegraphics[height=8.8em]{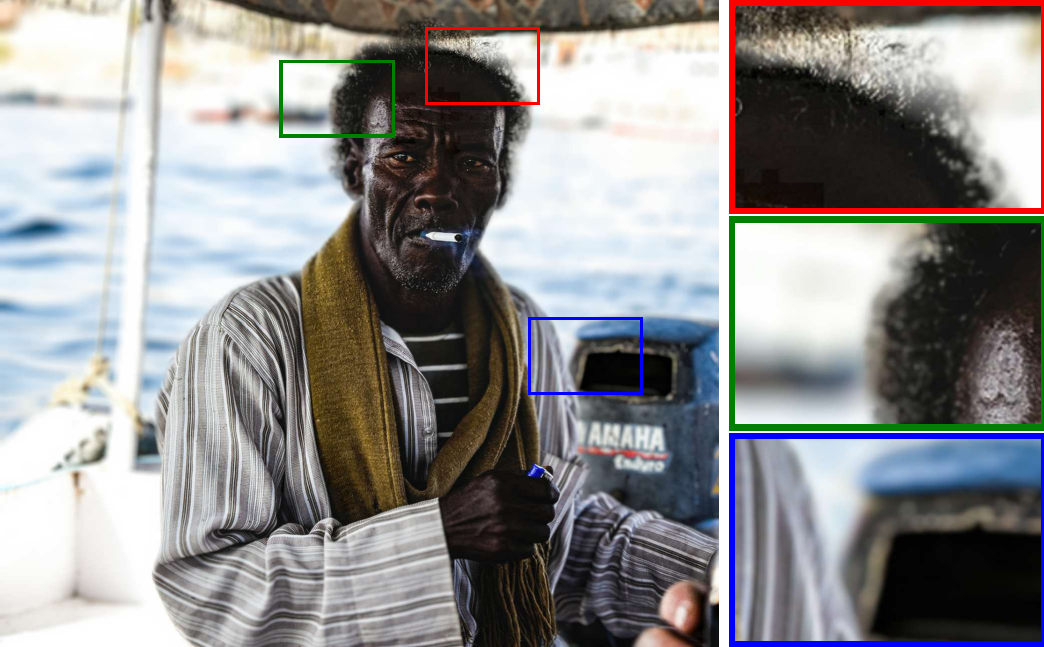}\hfill
    \hspace*{\fill}
    
    \vspace{-0.5em}
    \caption{More qualitative comparisons of BokehDiff with BokehMe~\cite{peng2022bokehme}, MPIB~\cite{peng2022mpib}, and Dr.~Bokeh~\cite{sheng2024dr}. Calculated from disparity, the defocus map is shared across the methods to be compared. The defocus map is for reference only, with whiter regions for more lens blur, but is subjected to error caused by depth estimation.}
    \label{fig:more_comparison1}
    \vspace{-1em}
\end{figure*}

First we demonstrate some more comparisons in \cref{fig:more_comparison1}. In the first example, BokehDiff successfully focuses on the person, and creates a progressive blurrness for the ground behind and before the person. In comparison, BokehMe~\cite{peng2022bokehme} over-blurs the person's helmet and hands due to the erroneous disparity estimation. MPIB~\cite{peng2022mpib} fails to produce the progressive blur, while Dr.~Bokeh~\cite{sheng2024dr} creates unnatural split near the boundary of the person.

In the second example, all the baselines over-blurs the thin end of the man's beard, while BokehDiff keeps the focused detail intact. As for the blurry foreground, BokehDiff creates a physically correct and beautiful semi-transparent blur near the unfocused edge of the sleeves. In comparison, the baselines either create a hard edge (BokehMe~\cite{peng2022bokehme} and Dr.~Bokeh~\cite{sheng2024dr}) or over-blur the boundary (MPIB~\cite{peng2022mpib}).

The third example is another case where BokehDiff outperforms previous methods at depth discontinuities. Most of the hair of the person should be focused, but the baselines over-blur the part near the edge, while BokehDiff manages to keep the fine details in focus. The transition to out-of-focus area is also smooth and natural.

\begin{figure*}[h]
    \centering
    \vspace{-1em}
    \hspace*{\fill}
    \rotatebox{90}{\quad\quad\quad  \sffamily Defocus}\hfill
    \includegraphics[height=9.9em]{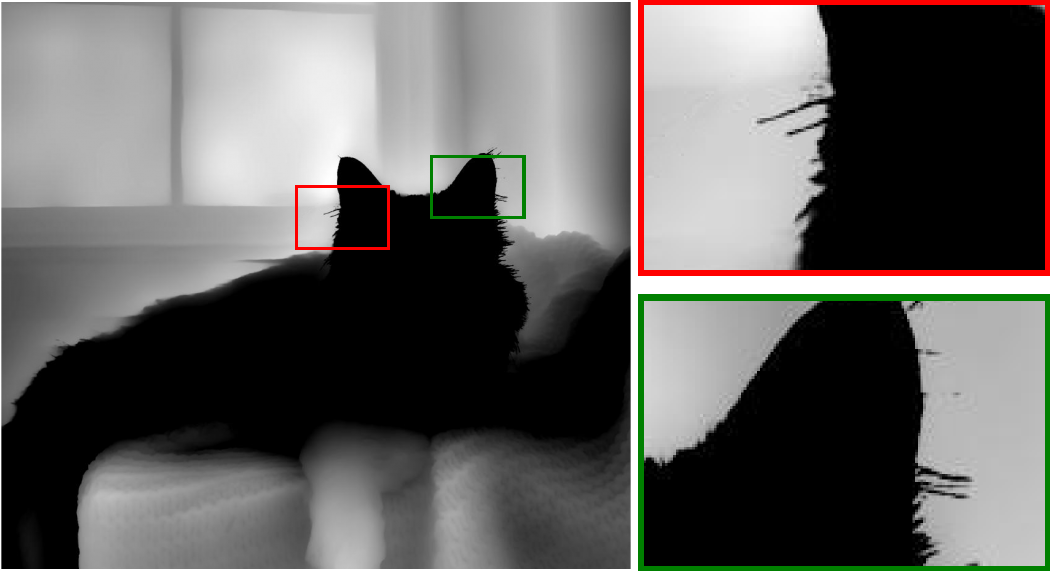}\hfill
    \includegraphics[height=9.9em]{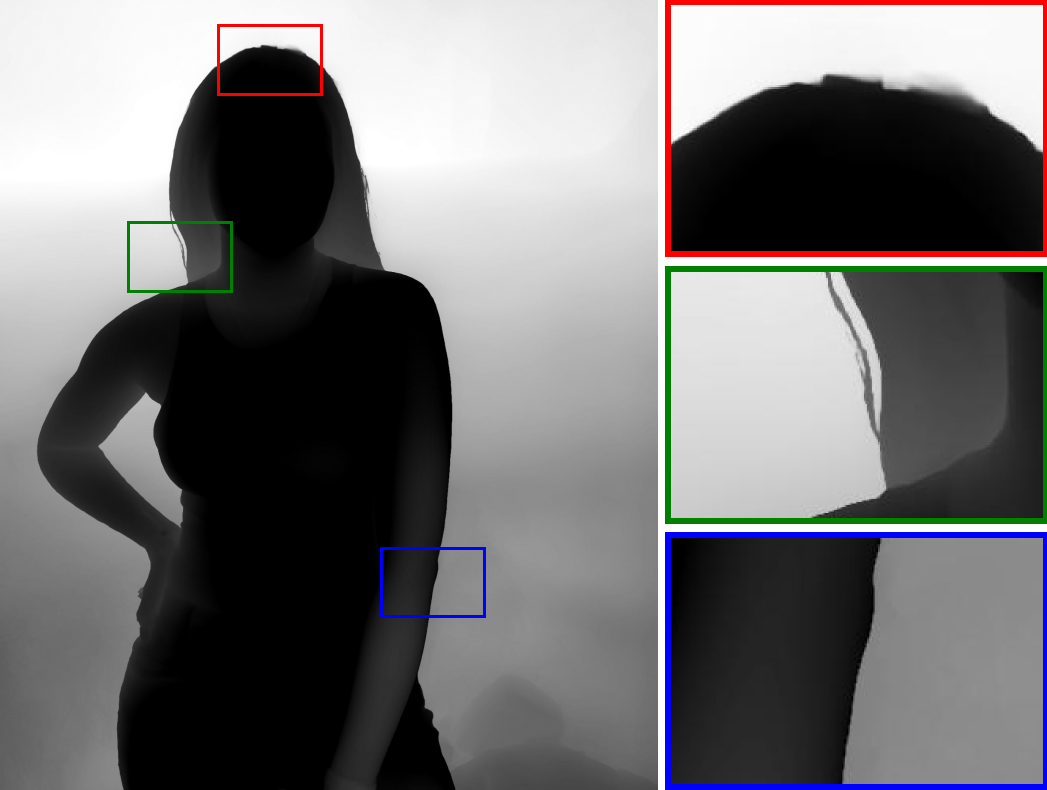}\hfill
    \includegraphics[height=9.9em]{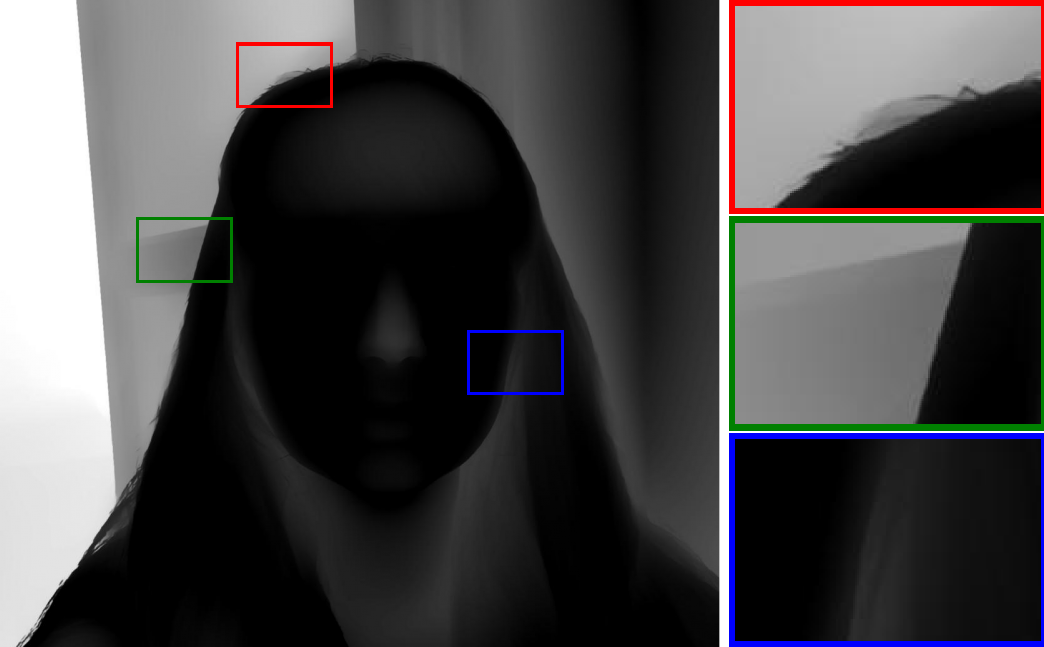}\hfill
    \hspace*{\fill}
    
\hspace*{\fill}
    \rotatebox{90}{\quad\quad \  \sffamily All-in-focus}\hfill
    \includegraphics[height=9.9em]{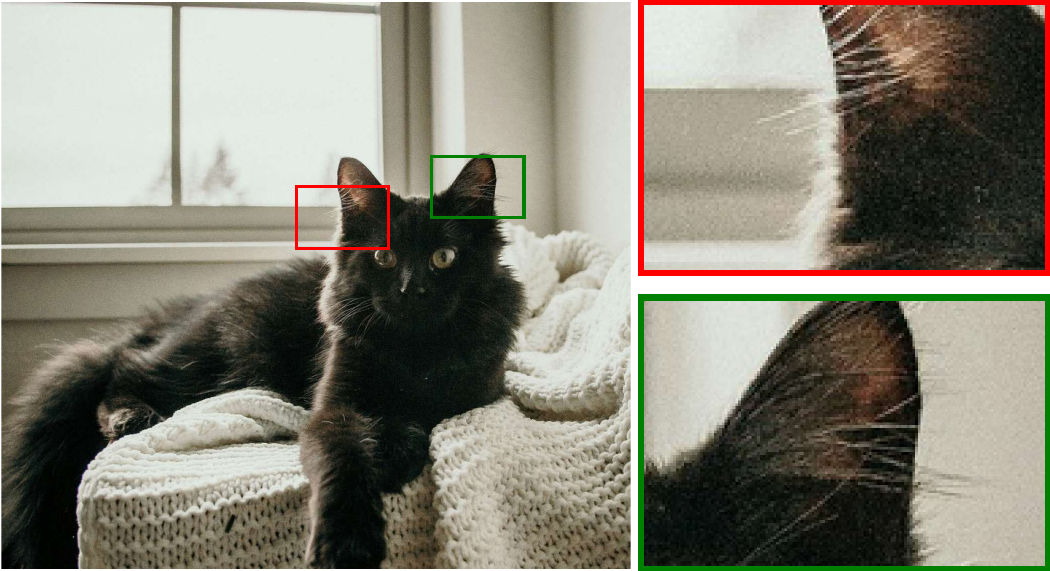}\hfill
    \includegraphics[height=9.9em]{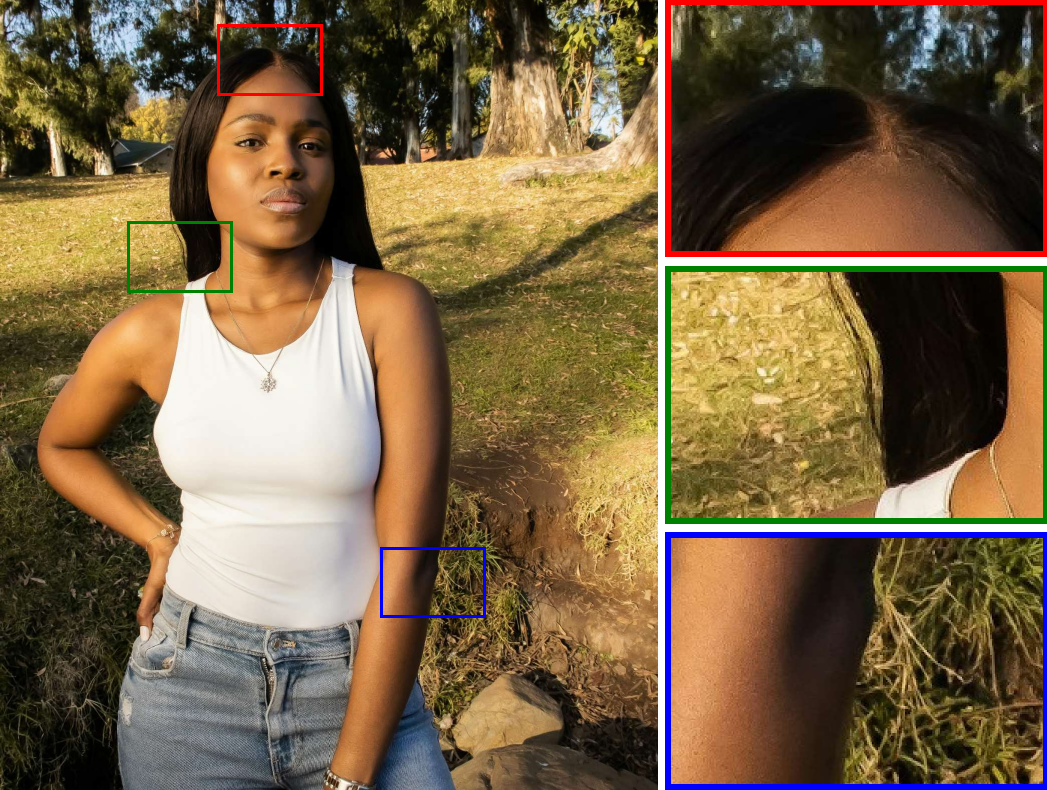}\hfill
    \includegraphics[height=9.9em]{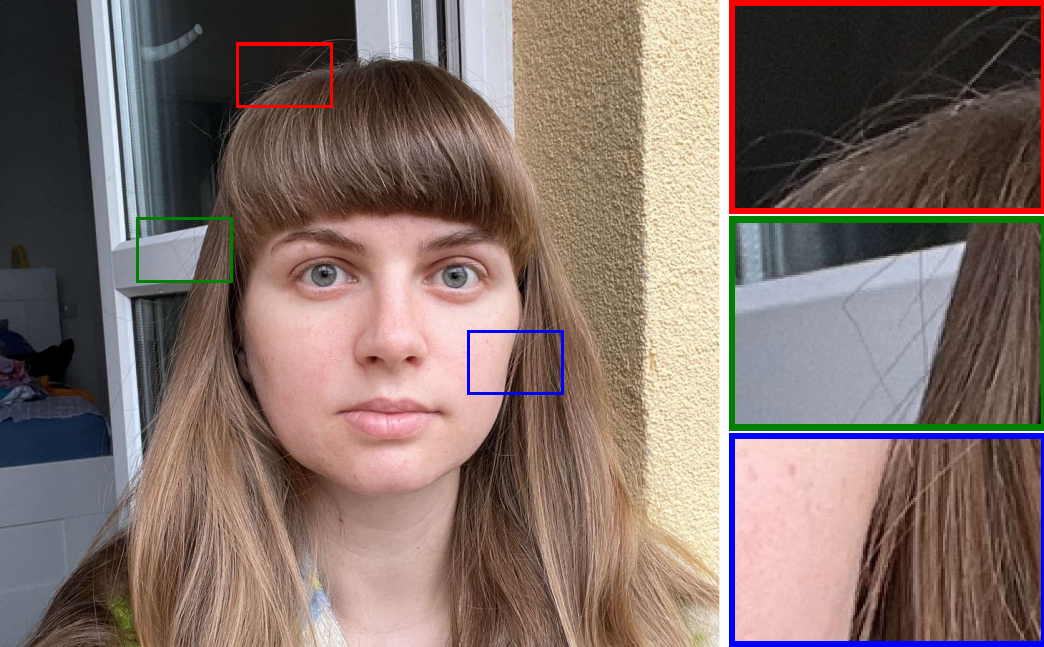}\hfill
    \hspace*{\fill}
    
    \hspace*{\fill}
    \rotatebox{90}{\quad\quad\ \ \sffamily BokehDiff}\hfill
    \includegraphics[height=9.9em]{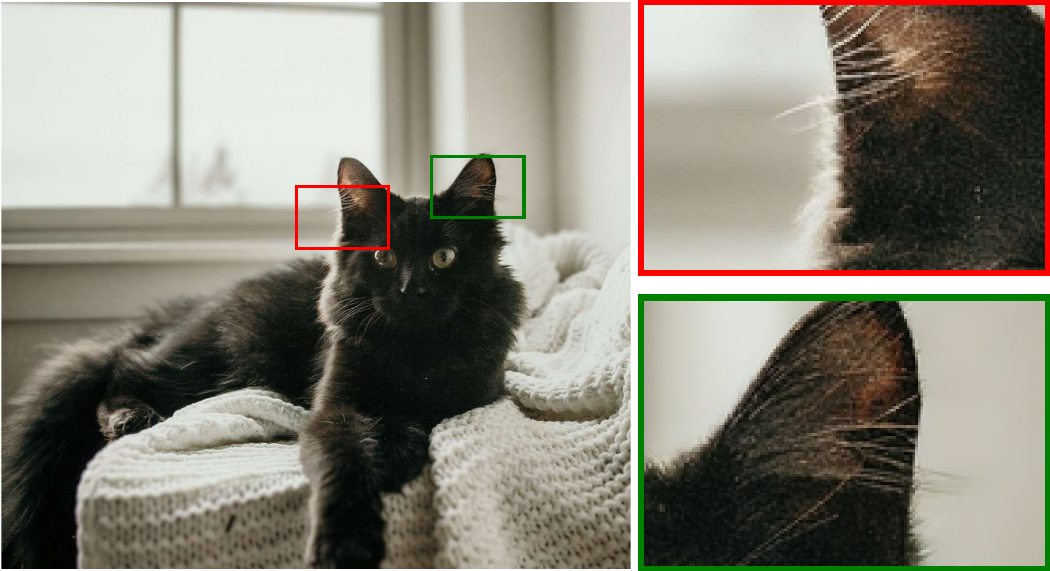}\hfill
    \includegraphics[height=9.9em]{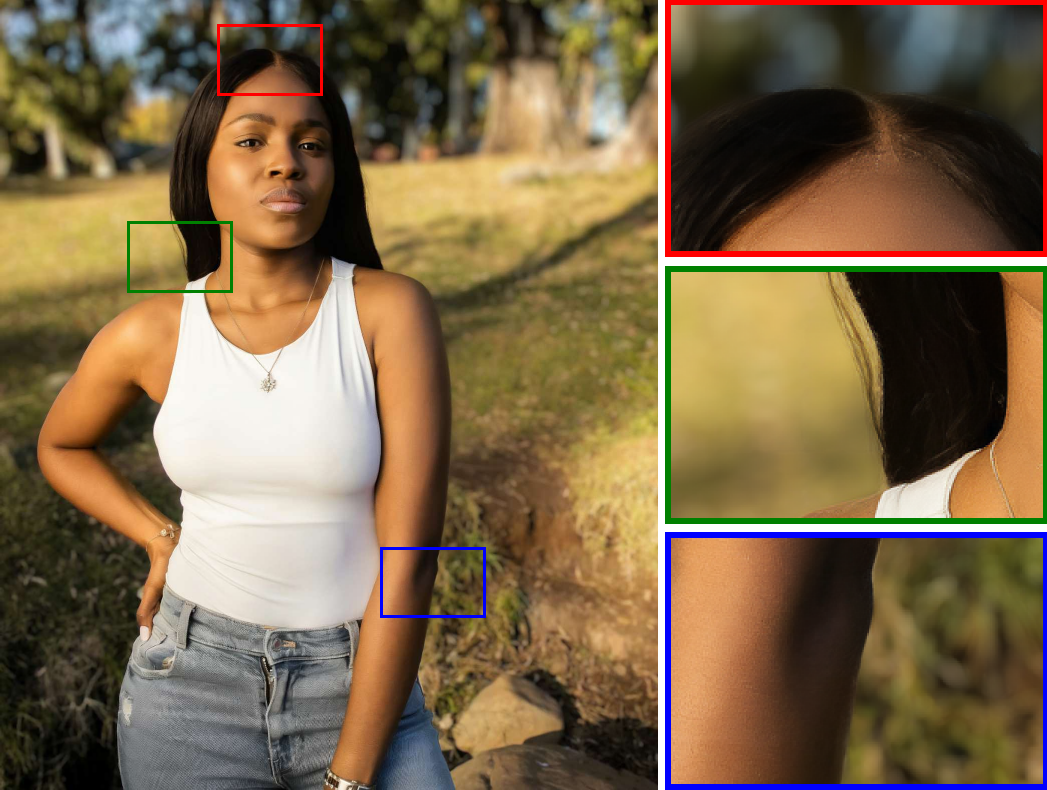}\hfill
    \includegraphics[height=9.9em]{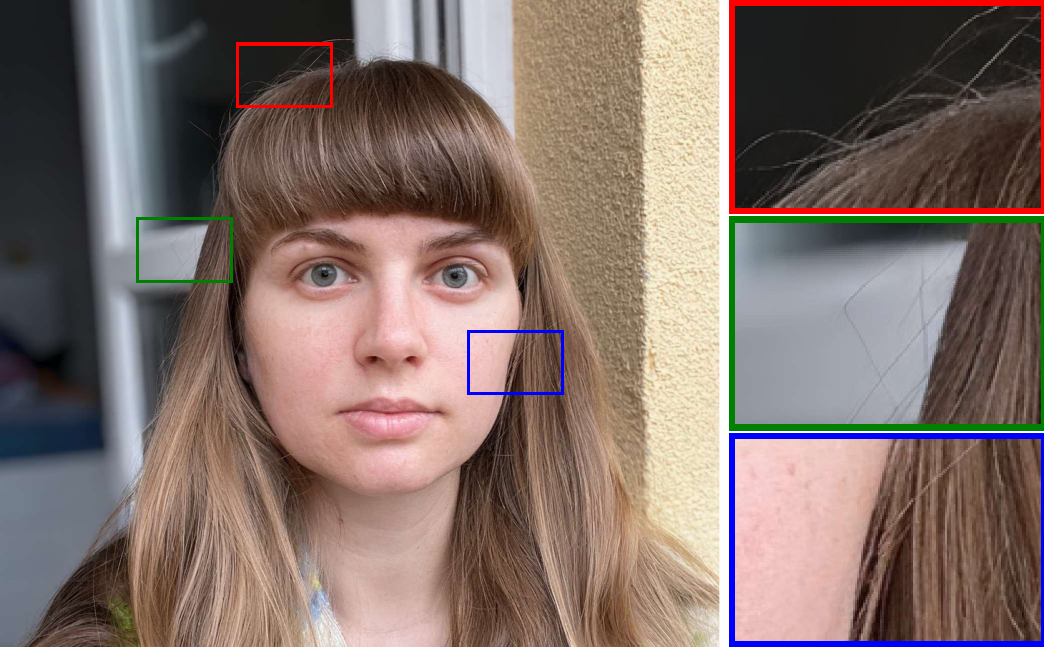}\hfill
    \hspace*{\fill}

    \hspace*{\fill}
    \rotatebox{90}{\quad\quad\  \sffamily BokehMe}\hfill
    \includegraphics[height=9.9em]{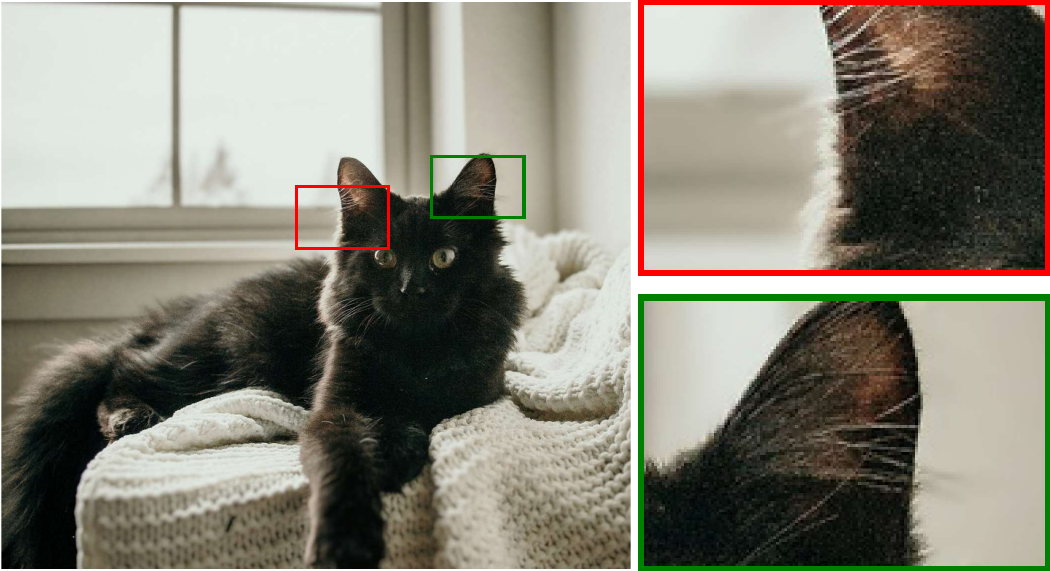}\hfill
    \includegraphics[height=9.9em]{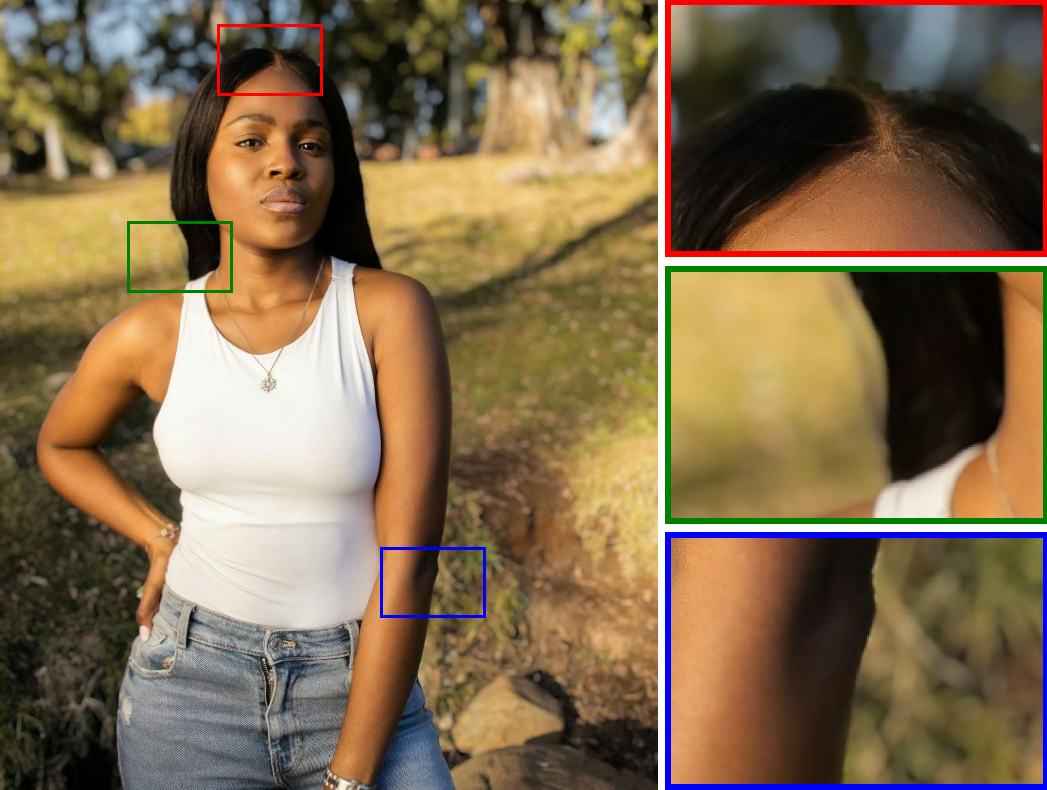}\hfill
    \includegraphics[height=9.9em]{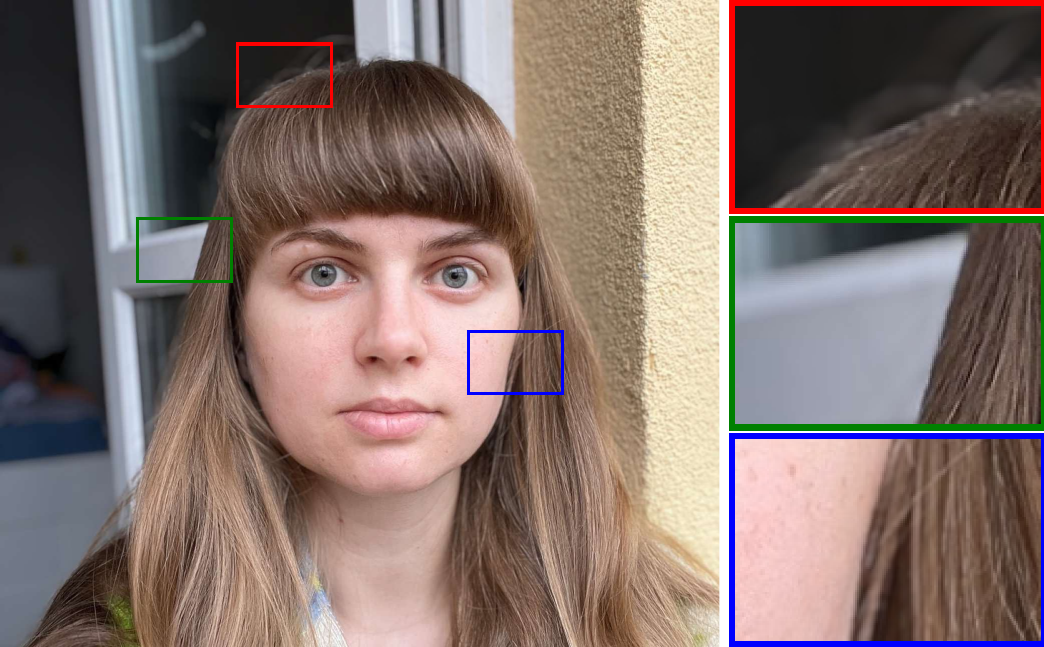}\hfill
    \hspace*{\fill}

     \hspace*{\fill}
    \rotatebox{90}{\quad\quad\quad\ \sffamily MPIB}\hfill
    \includegraphics[height=9.9em]{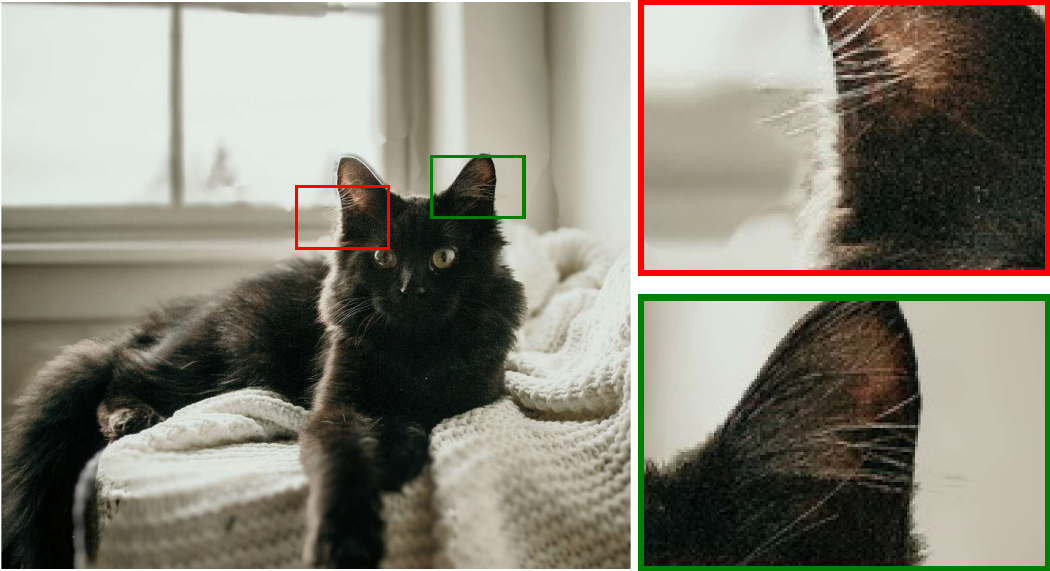}\hfill
    \includegraphics[height=9.9em]{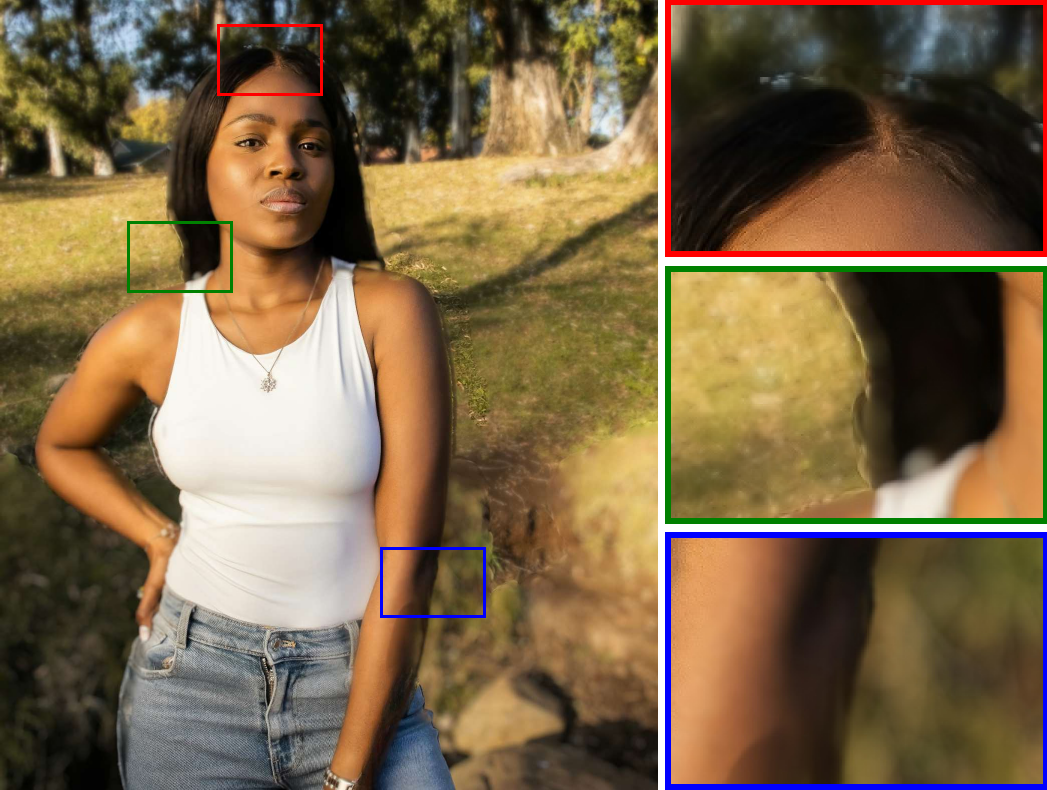}\hfill
    \includegraphics[height=9.9em]{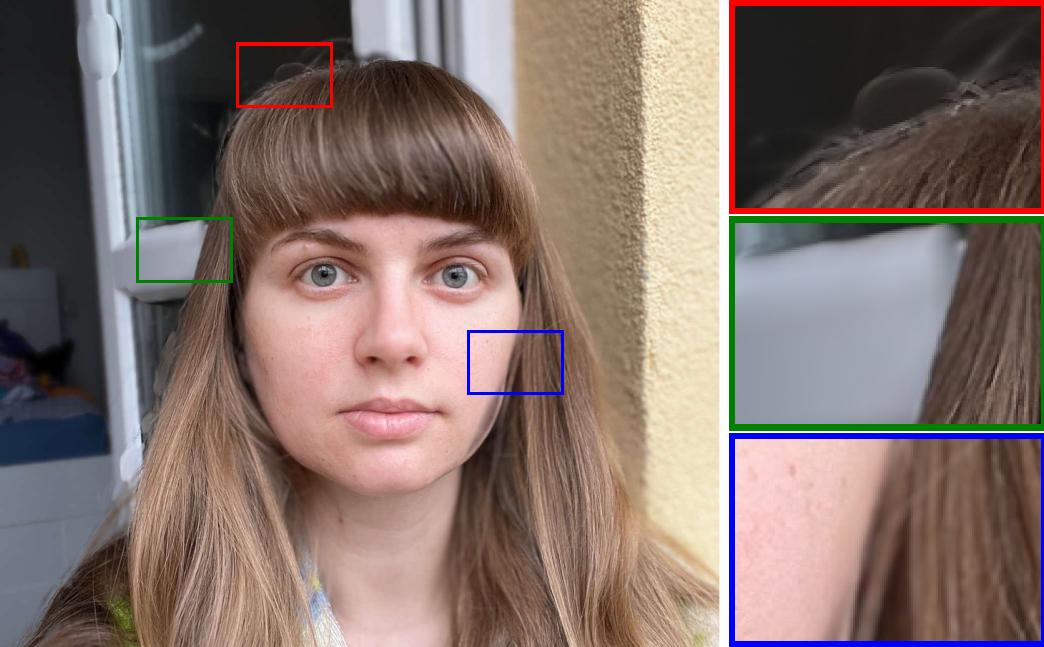}\hfill
    \hspace*{\fill}
    
    \hspace*{\fill}
    \rotatebox{90}{\quad\quad\ \ \sffamily  Dr.~Bokeh}\hfill
    \includegraphics[height=9.9em]{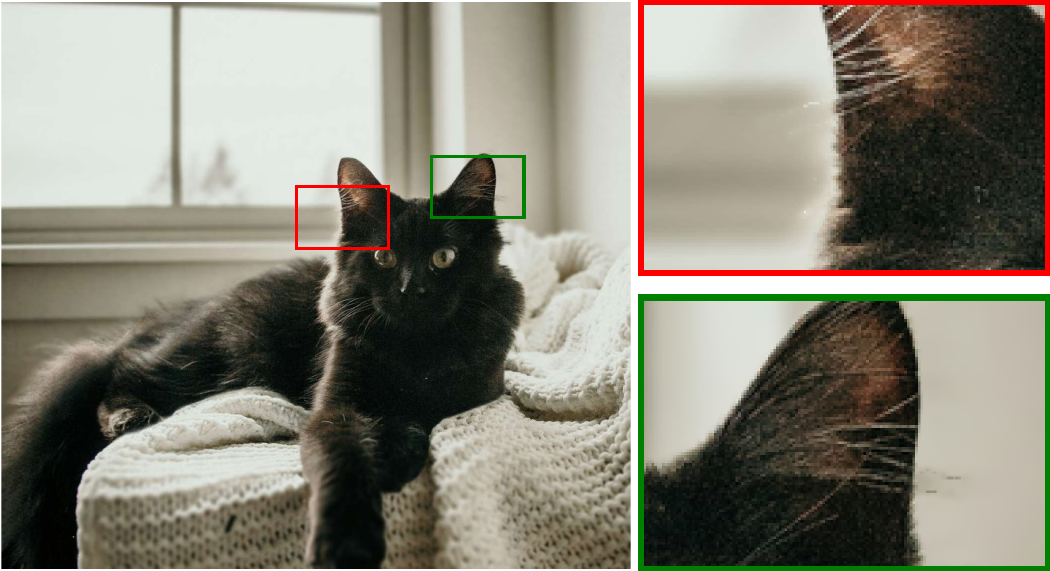}\hfill
    \includegraphics[height=9.9em]{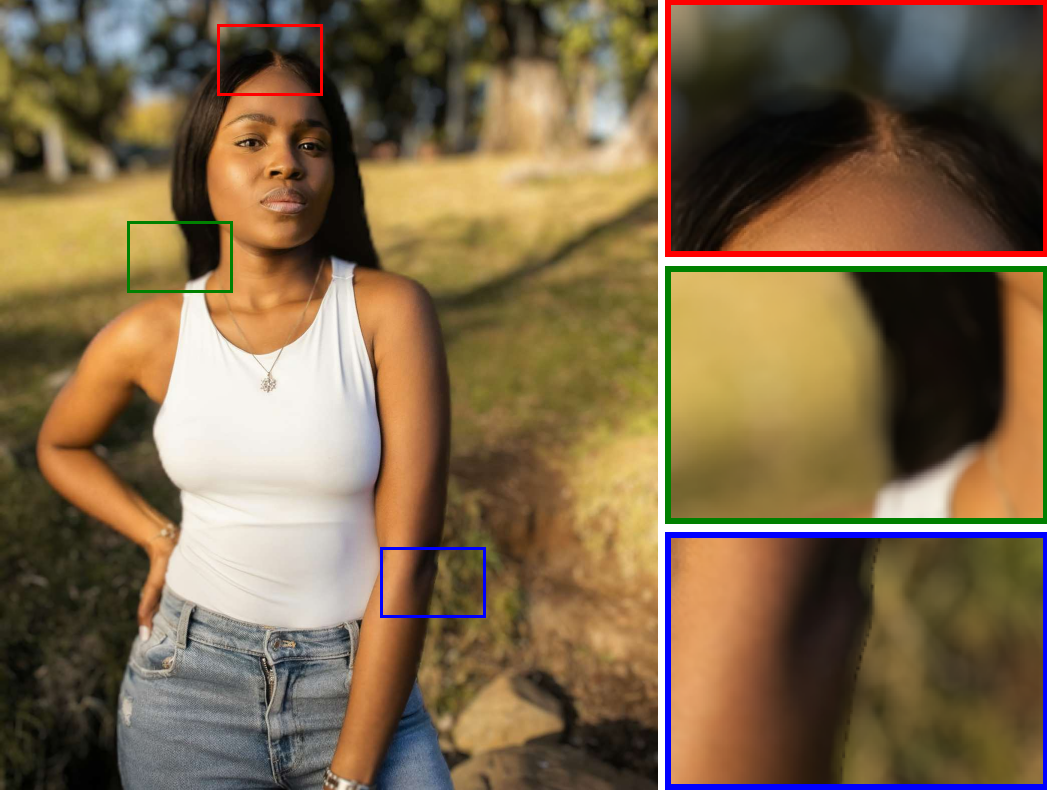}\hfill
    \includegraphics[height=9.9em]{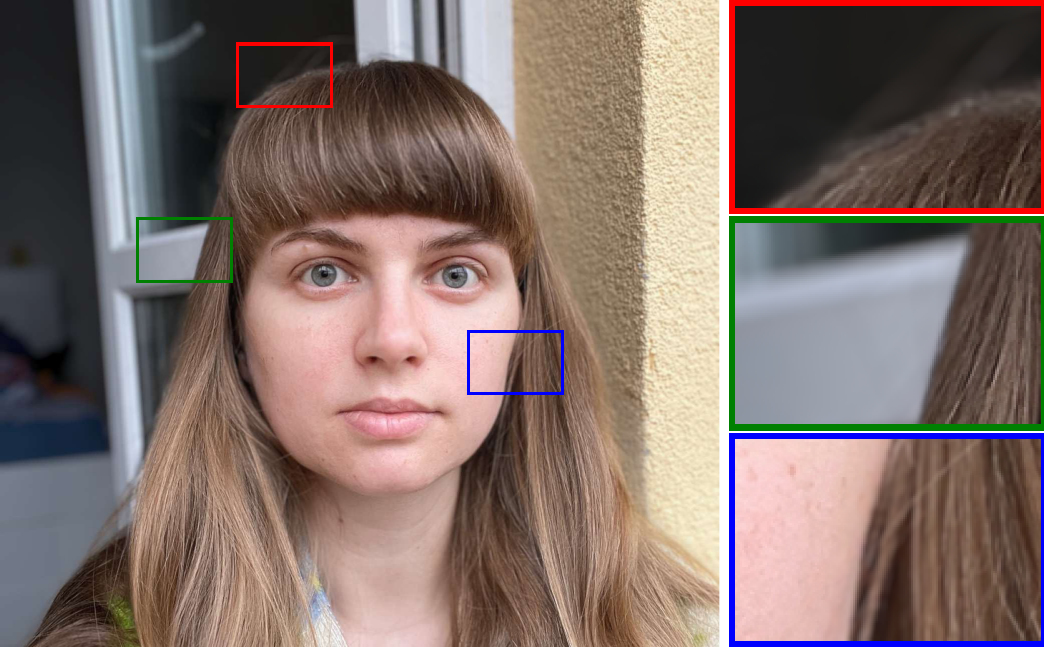}\hfill
    \hspace*{\fill}
    
    \vspace{-0.5em}
    \caption{More qualitative comparisons of BokehDiff with BokehMe~\cite{peng2022bokehme}, MPIB~\cite{peng2022mpib}, and Dr.~Bokeh~\cite{sheng2024dr}. Calculated from disparity, the defocus map is shared across the methods to be compared. The defocus map is for reference only, with whiter regions for more lens blur, but is subjected to error caused by depth estimation.}
    \label{fig:more_comparison2}
    \vspace{-1em}
\end{figure*}

We continue the demonstration of results in \cref{fig:more_comparison2}. In the first example BokehDiff keeps the thin details of the cat's fur while blurring the window behind them, while the other methods show different degrees of artifacts.
In the second example, BokehDiff manages to create a progressive blur as the defocus increases, while keeping the focused foreground intact, even in such area as the hair seam and the elbow where the background is messy. In comparison, the baselines follow the inaccurate depth estimation result, and create bumps near the hair, and unnatural zig-zags near the elbow. The hair in the green box is also blurred by mistake.

The third example also demonstrates the effectiveness of BokehDiff in generating realisitic lens blur for intricate structures. The progressive blur in the background also validates that BokehDiff follows the image formation model.

\begin{figure*}[t]
    \centering
    \vspace{-1em}
    \hspace*{\fill}
    \rotatebox{90}{\quad\quad\quad  \sffamily Defocus}\hfill
    \includegraphics[height=9.9em]{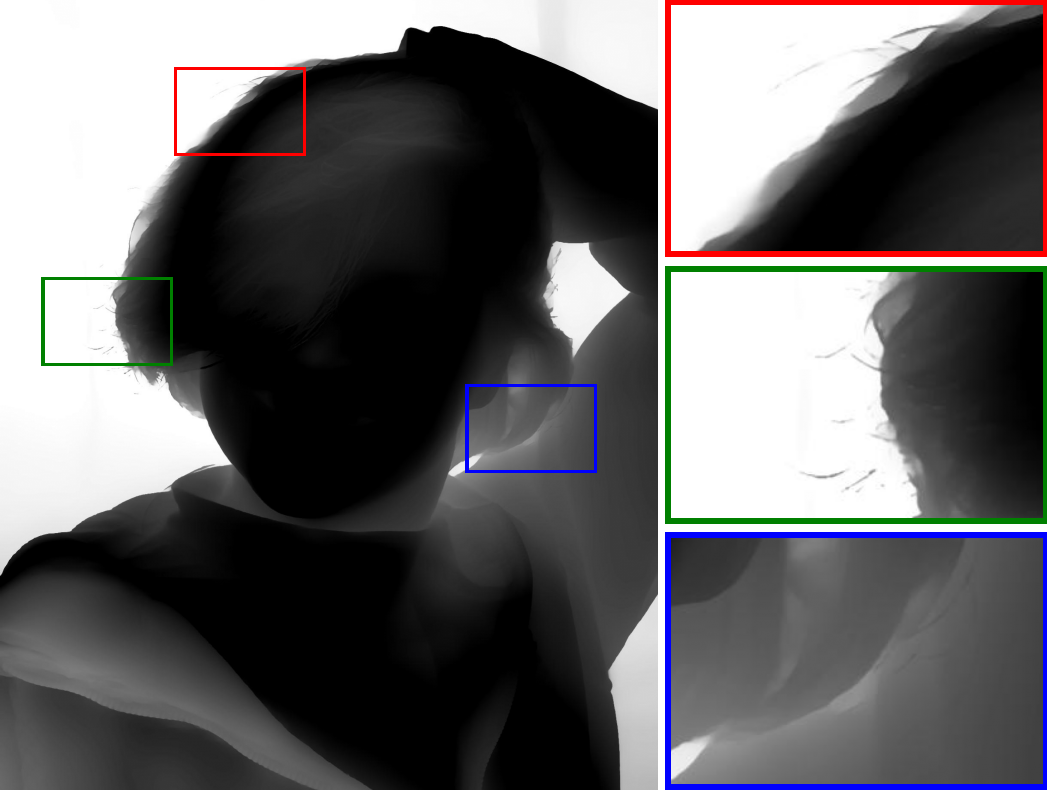}\hfill
    \includegraphics[height=9.9em]{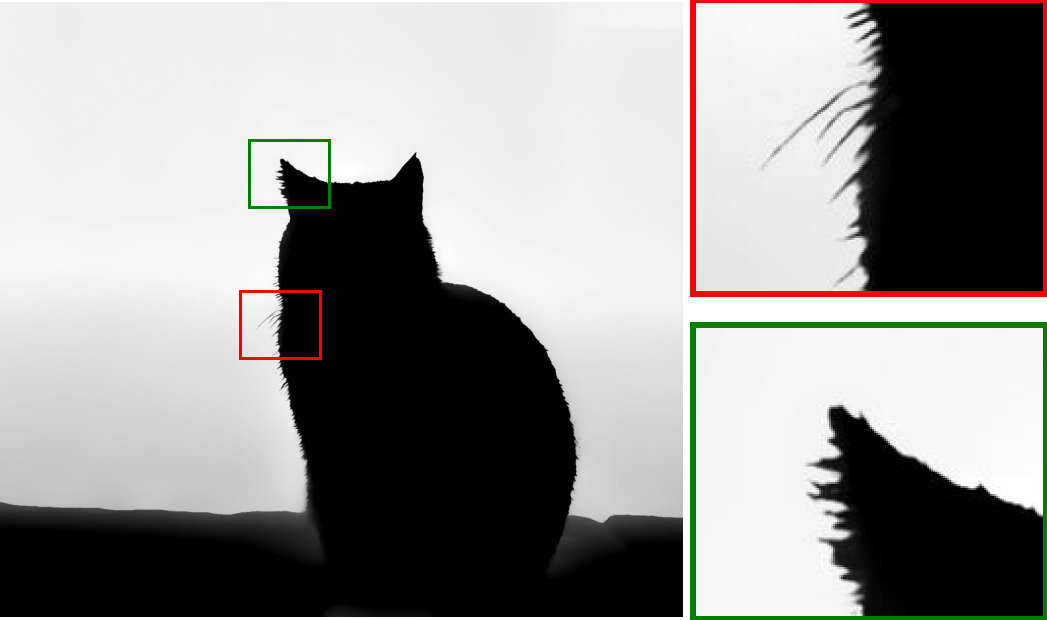}\hfill
    \includegraphics[height=9.9em]{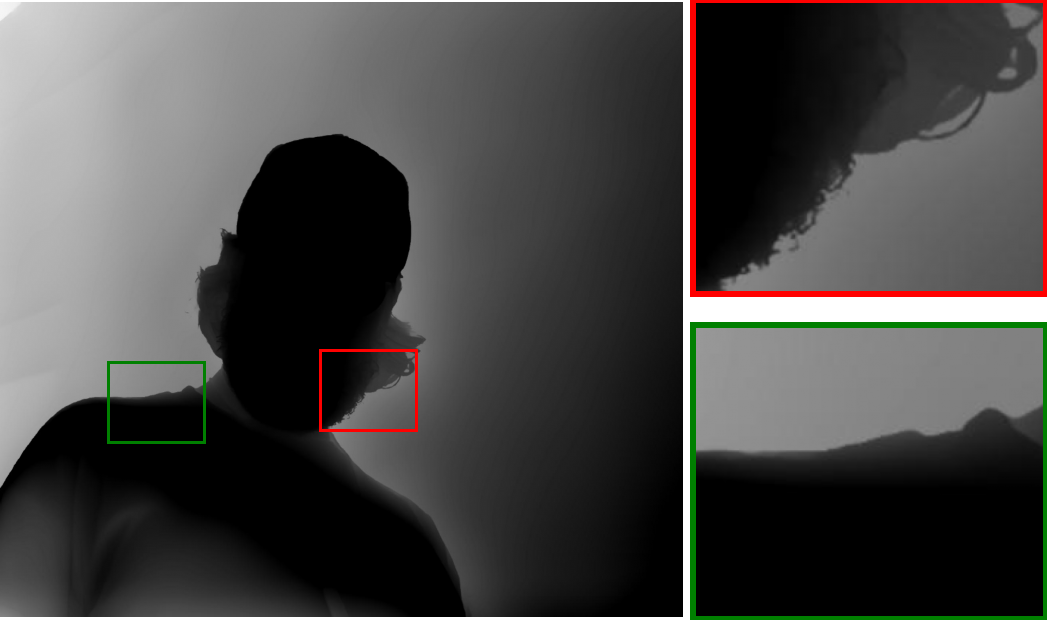}\hfill
    \hspace*{\fill}
    
\hspace*{\fill}
    \rotatebox{90}{\quad\quad \  \sffamily All-in-focus}\hfill
    \includegraphics[height=9.9em]{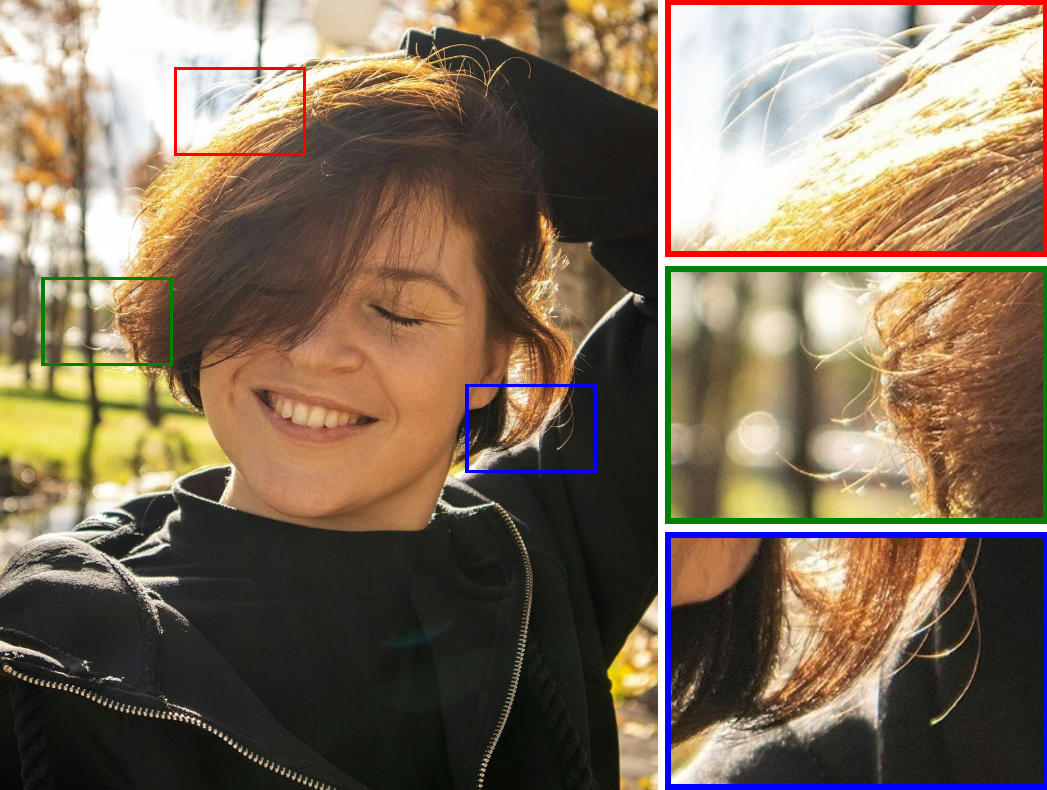}\hfill
    \includegraphics[height=9.9em]{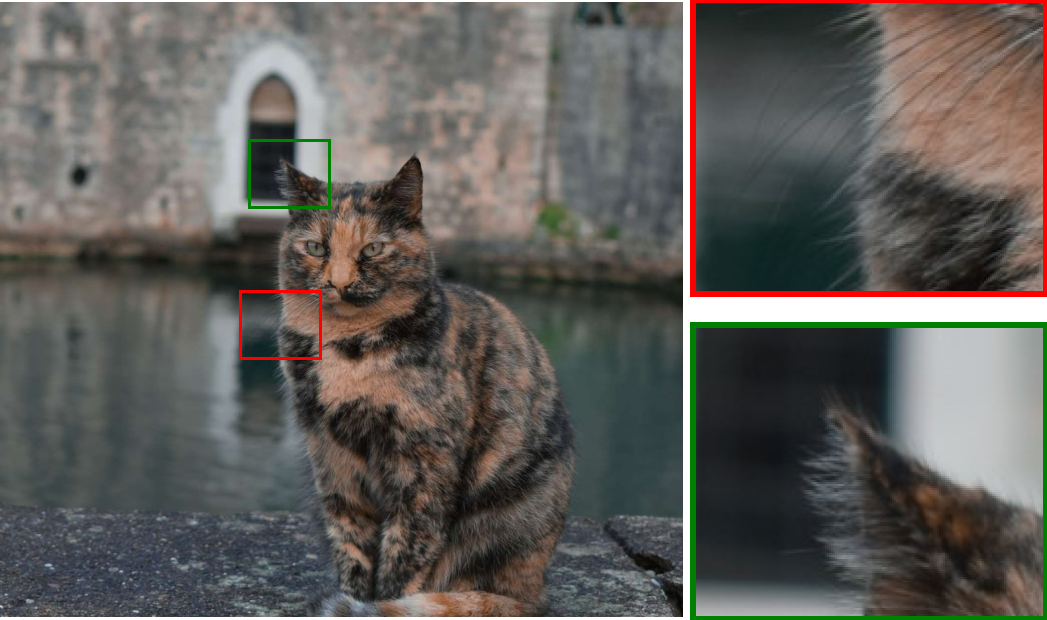}\hfill
    \includegraphics[height=9.9em]{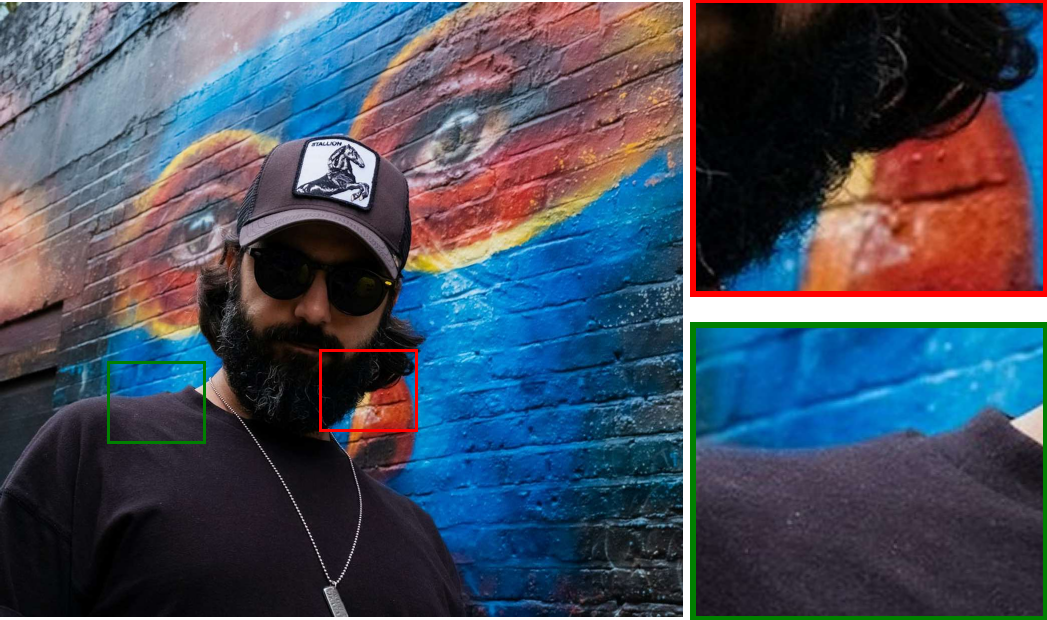}\hfill
    \hspace*{\fill}
    
    \hspace*{\fill}
    \rotatebox{90}{\quad\quad\ \ \sffamily BokehDiff}\hfill
    \includegraphics[height=9.9em]{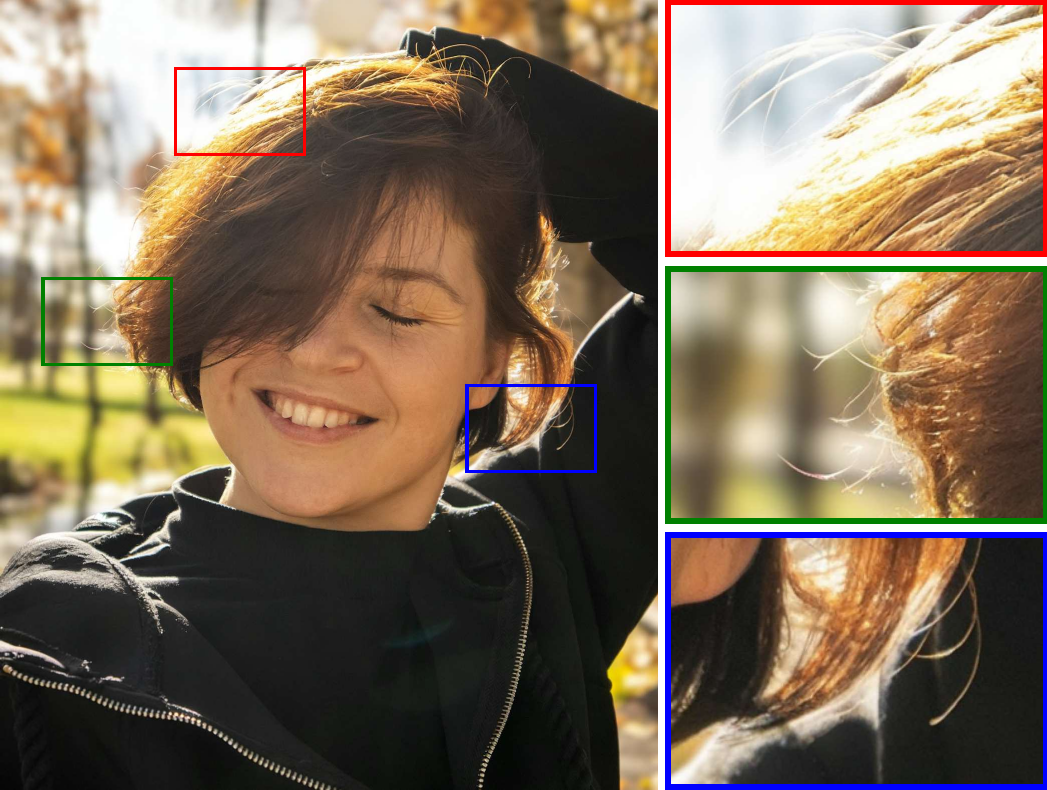}\hfill
    \includegraphics[height=9.9em]{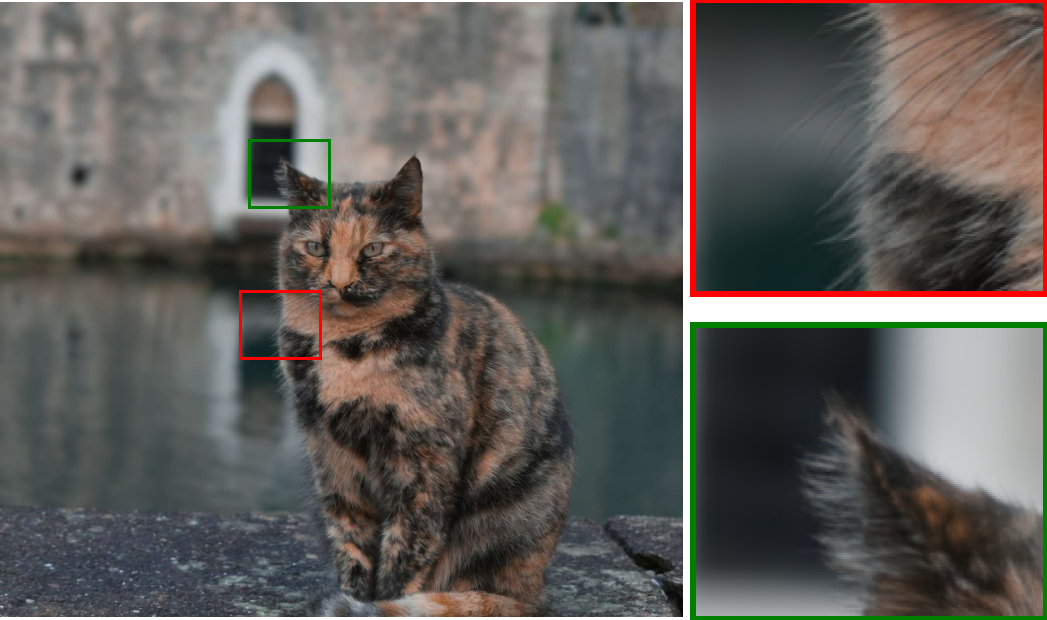}\hfill
    \includegraphics[height=9.9em]{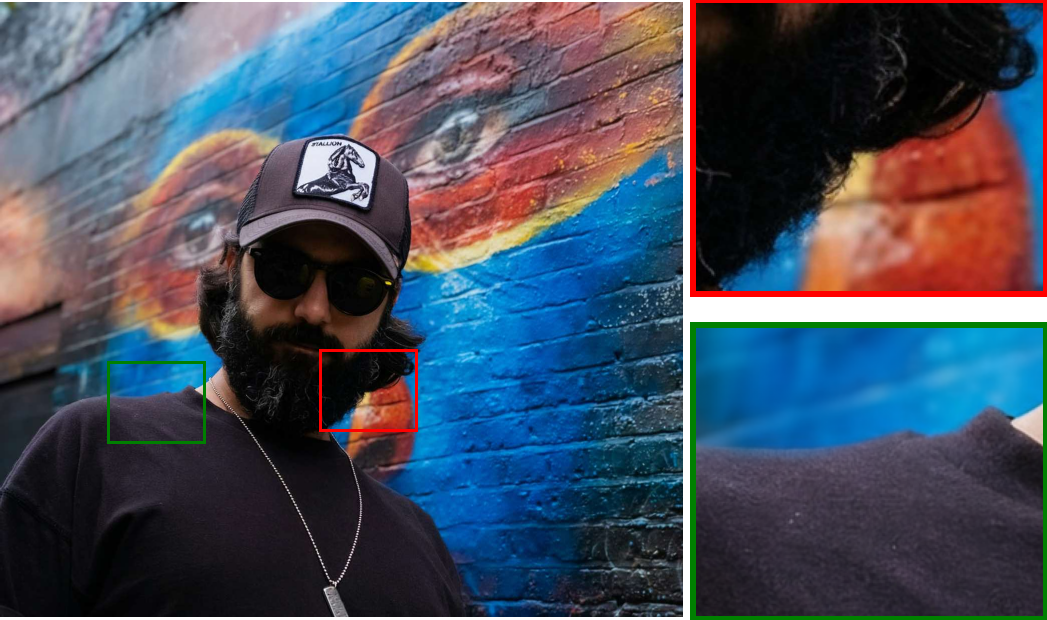}\hfill
    \hspace*{\fill}

    \hspace*{\fill}
    \rotatebox{90}{\quad\quad\  \sffamily BokehMe}\hfill
    \includegraphics[height=9.9em]{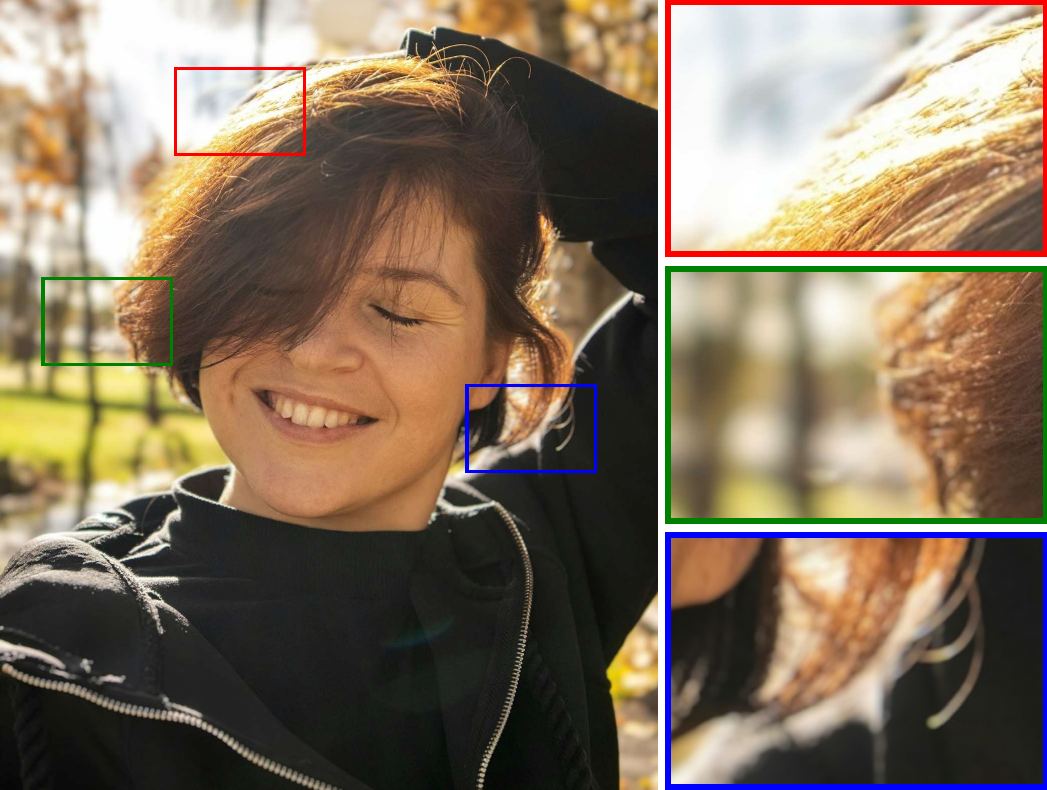}\hfill
    \includegraphics[height=9.9em]{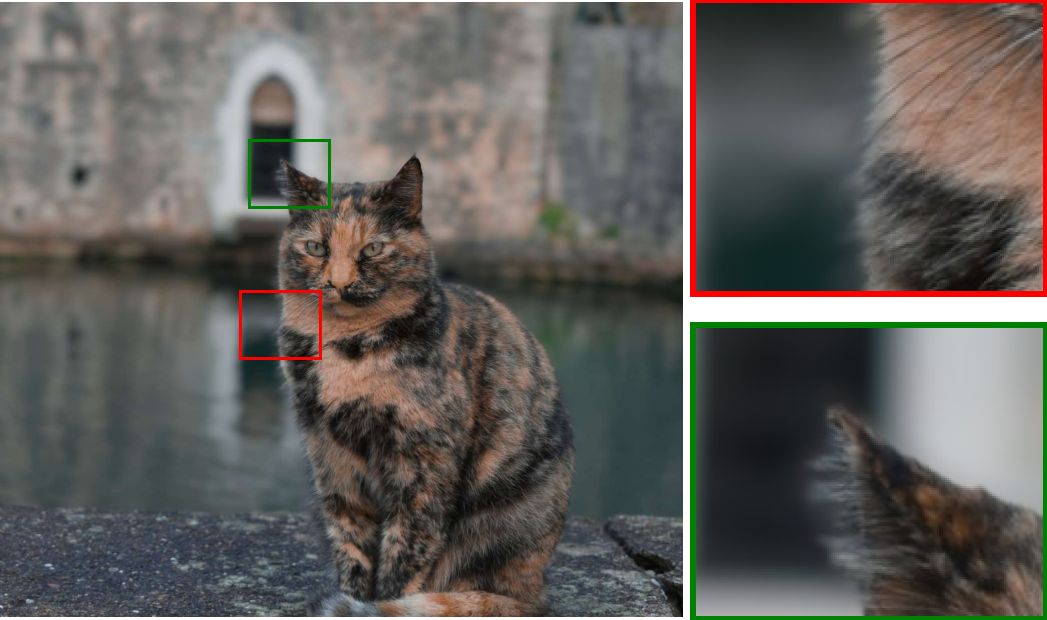}\hfill
    \includegraphics[height=9.9em]{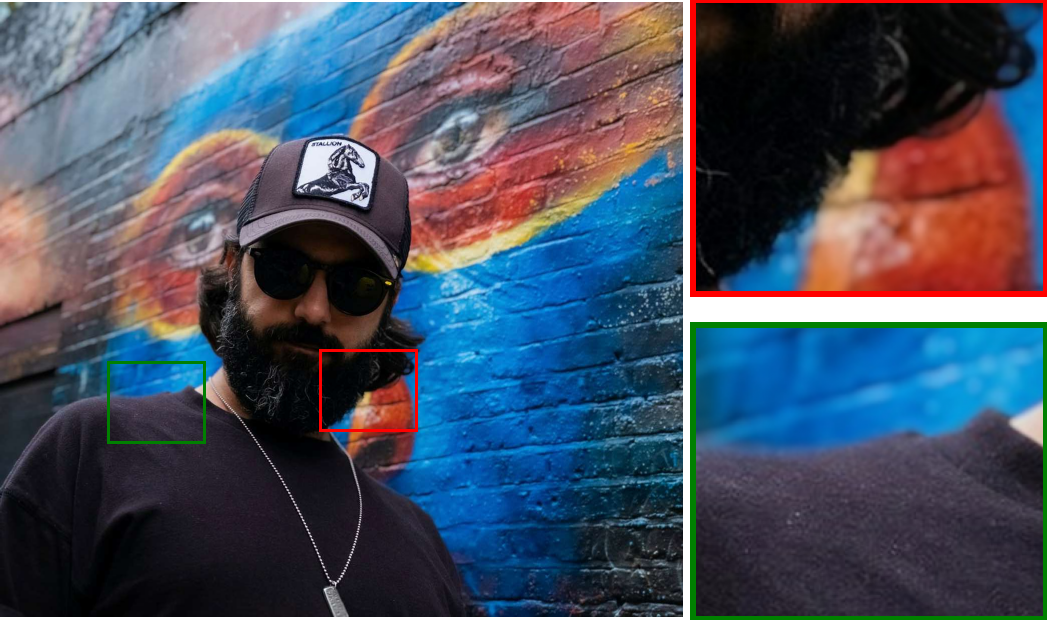}\hfill
    \hspace*{\fill}

     \hspace*{\fill}
    \rotatebox{90}{\quad\quad\quad\ \sffamily MPIB}\hfill
    \includegraphics[height=9.9em]{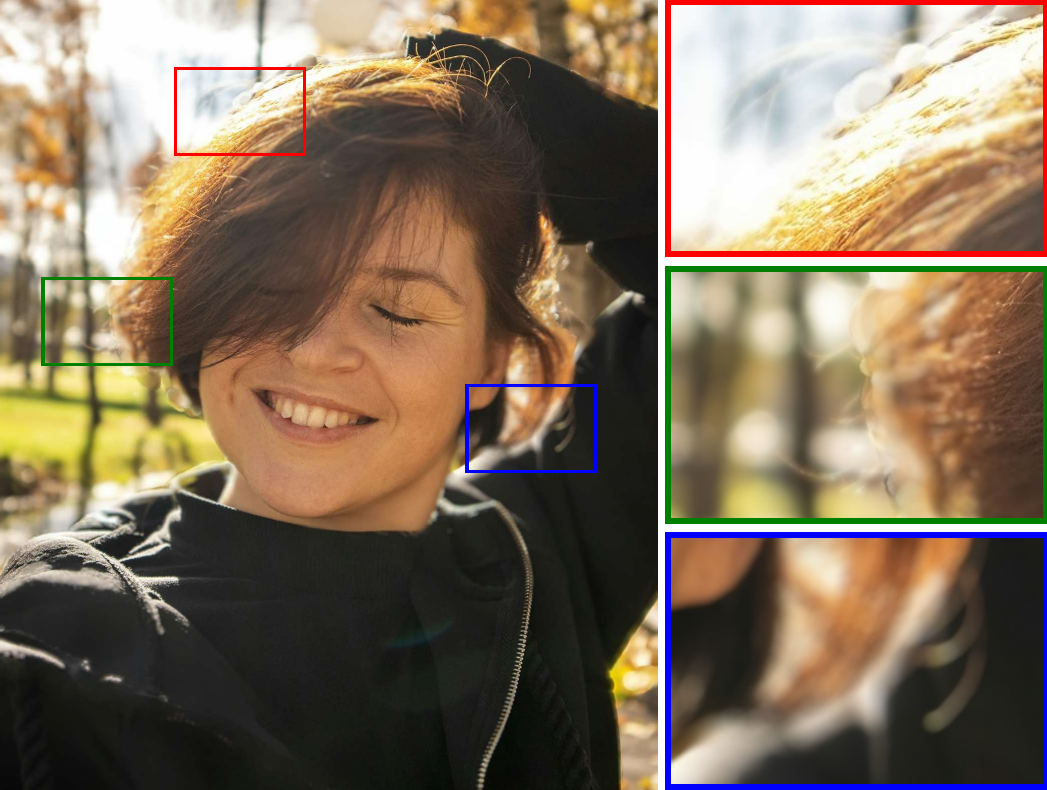}\hfill
    \includegraphics[height=9.9em]{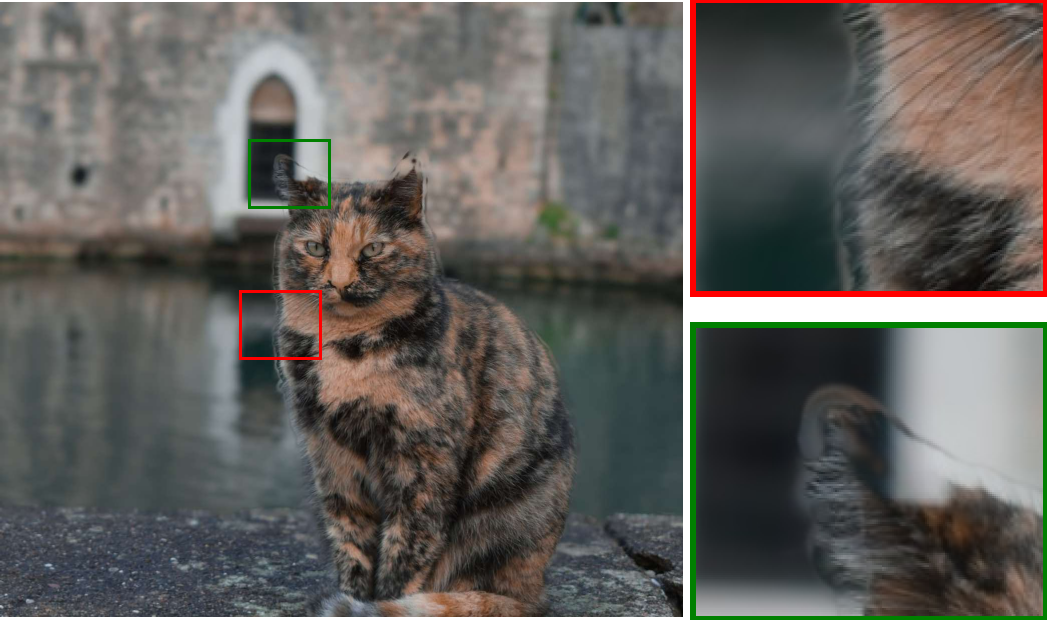}\hfill
    \includegraphics[height=9.9em]{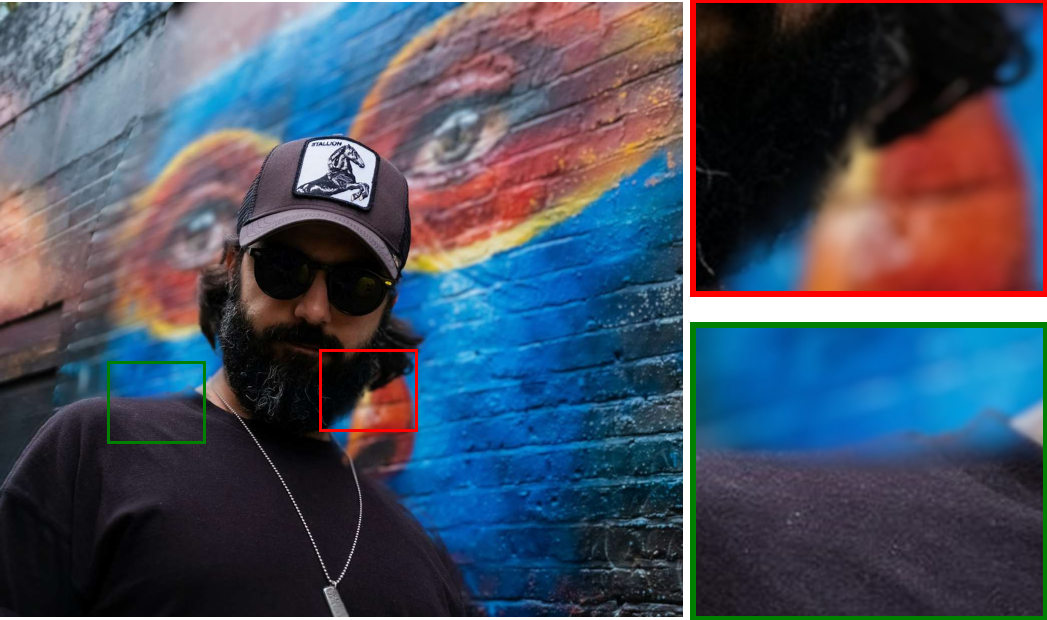}\hfill
    \hspace*{\fill}
    
    \hspace*{\fill}
    \rotatebox{90}{\quad\quad\ \ \sffamily  Dr.~Bokeh}\hfill
    \includegraphics[height=9.9em]{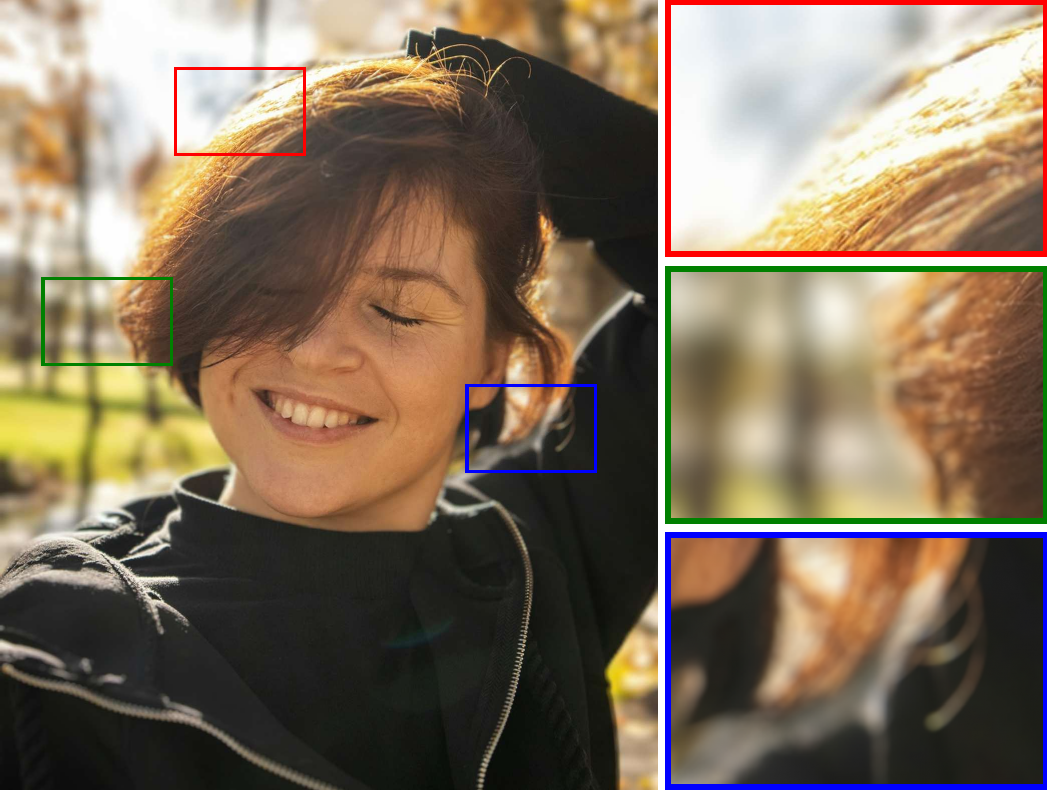}\hfill
    \includegraphics[height=9.9em]{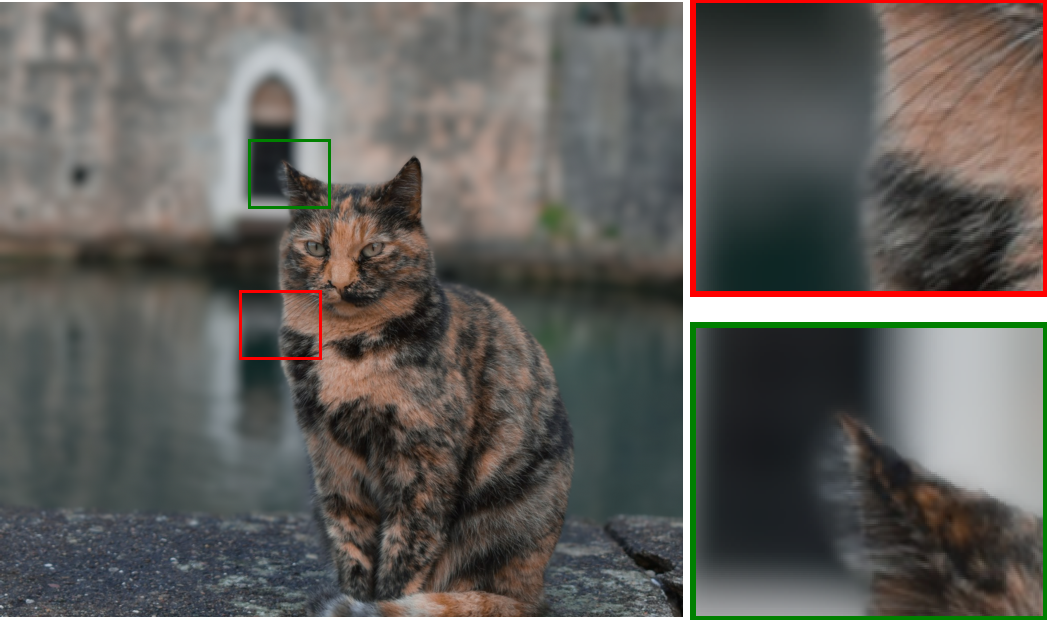}\hfill
    \includegraphics[height=9.9em]{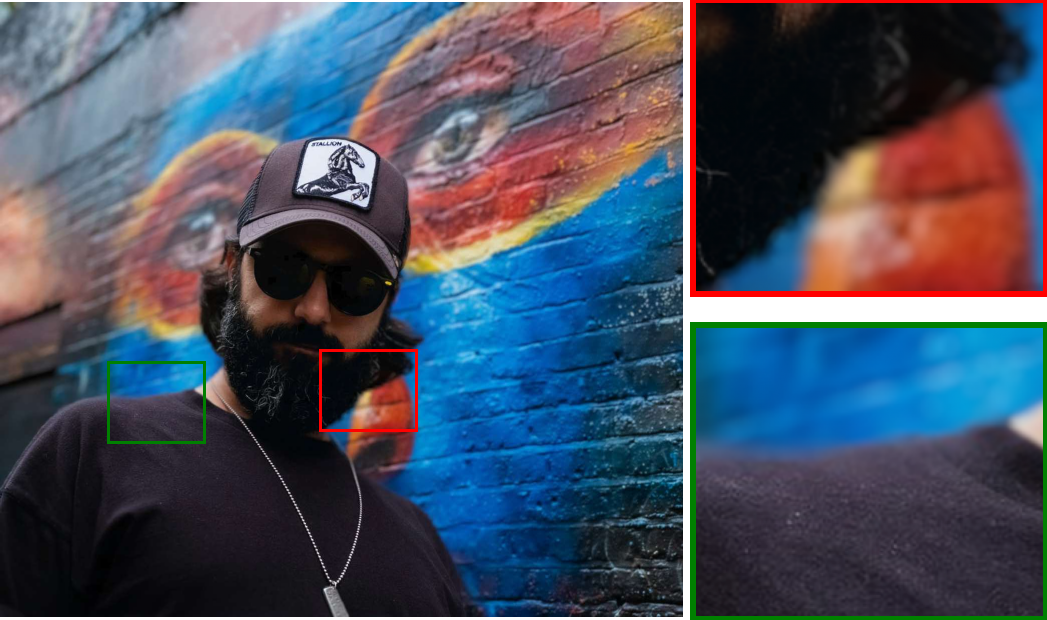}\hfill
    \hspace*{\fill}
    
    \vspace{-0.5em}
    \caption{More qualitative comparisons of BokehDiff with BokehMe~\cite{peng2022bokehme}, MPIB~\cite{peng2022mpib}, and Dr.~Bokeh~\cite{sheng2024dr}. Calculated from disparity, the defocus map is shared across the methods to be compared. The defocus map is for reference only, with whiter regions for more lens blur, but is subjected to error caused by depth estimation.}
    \label{fig:more_comparison3}
    \vspace{-1em}
\end{figure*}

In \cref{fig:more_comparison3}, as shown from the first two examples, the proposed method also work on images shot with a wide aperture, and further blurs the blurred background while keeping the foreground in focus. Both the hair streaks of the person and the furs of the cat are effectively kept. In the third example, BokehDiff also show the ability to synthesize progressive blurriness, while keeping the person's beard and the focused T-shirt. In comparison, the baselines cannot preserve the intricate details, as well as the regions where depth estimation methods go wrong, such as the hair of the person in the first example, the fur near the cat's ear in the second example, and the T-shirt edge in the third example. 

To sum up, given all the demonstrated results, we conclude that the results rendered by BokehDiff are both physically reasonable and visually pleasant.

\section{Visual Results of Ablation Study}
\begin{figure*}[t]
\centering
\subfloat[All-in-focus]{\includegraphics[width=0.192\linewidth]{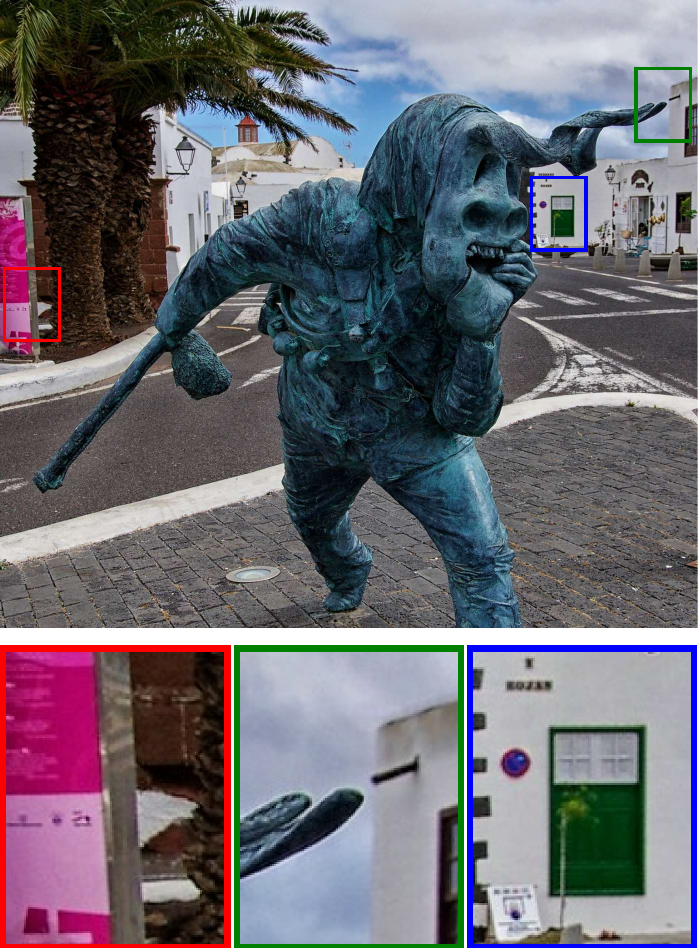}}
\subfloat[W/o CoC]{\includegraphics[width=0.192\linewidth]{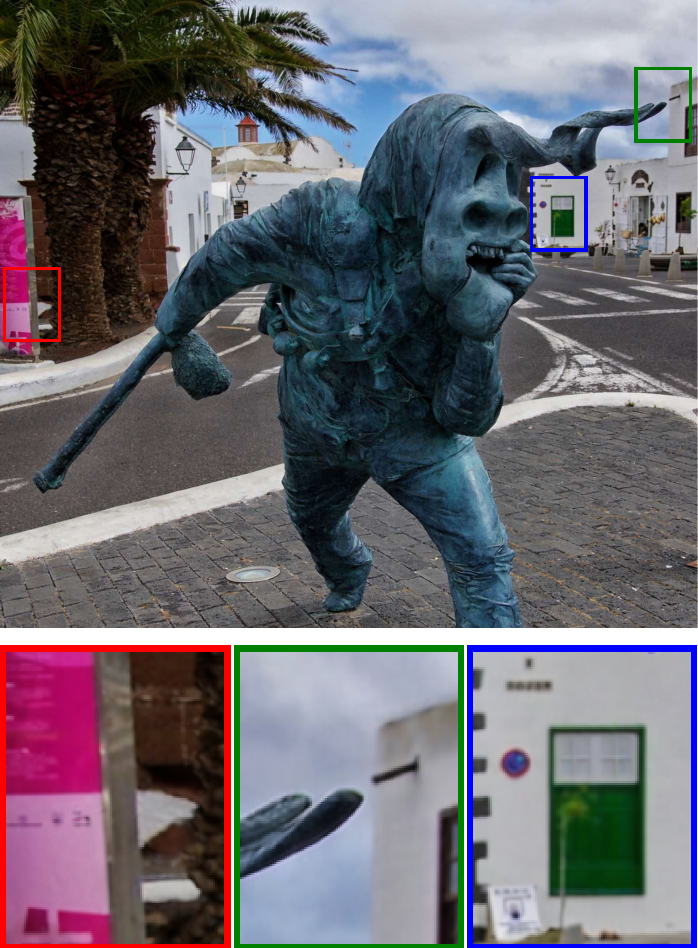}}
\subfloat[W/o occlusion]{\includegraphics[width=0.192\linewidth]{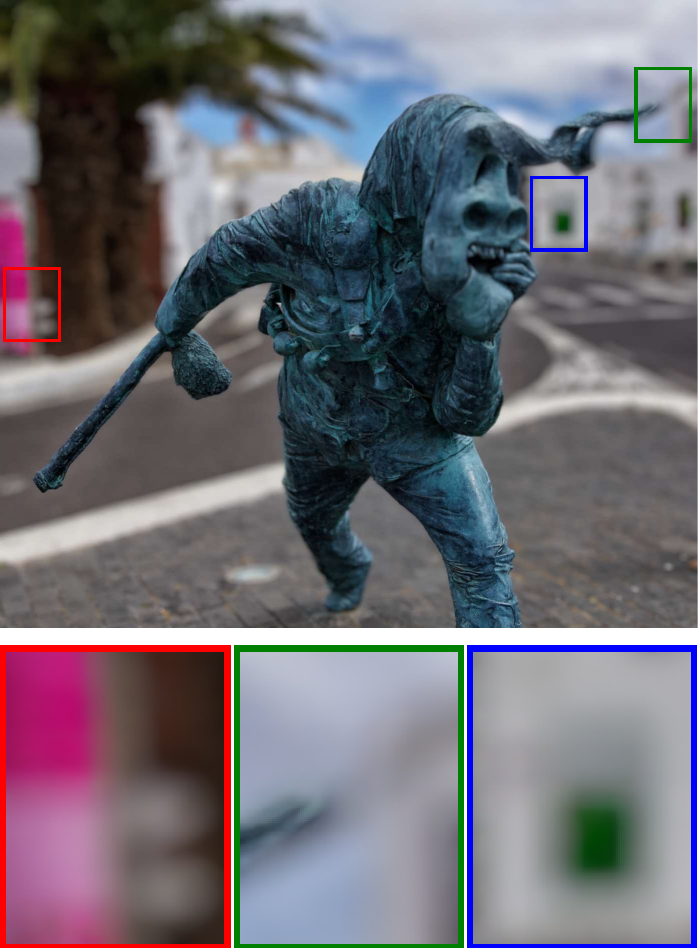}}
\subfloat[W/o SoftmaxQ]{\includegraphics[width=0.192\linewidth]{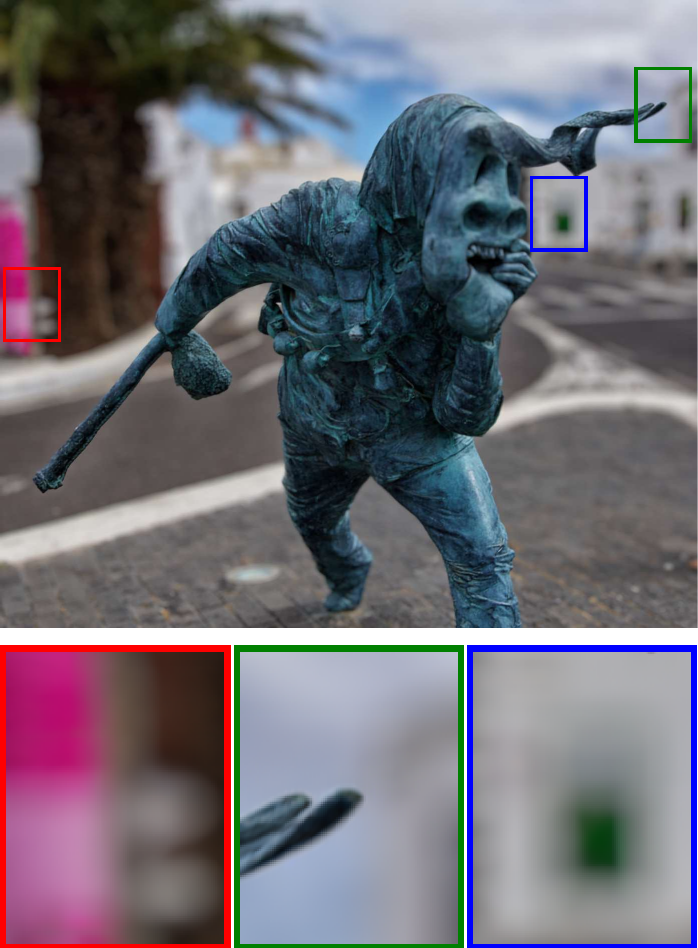}}
\subfloat[Complete Model]{\includegraphics[width=0.192\linewidth]{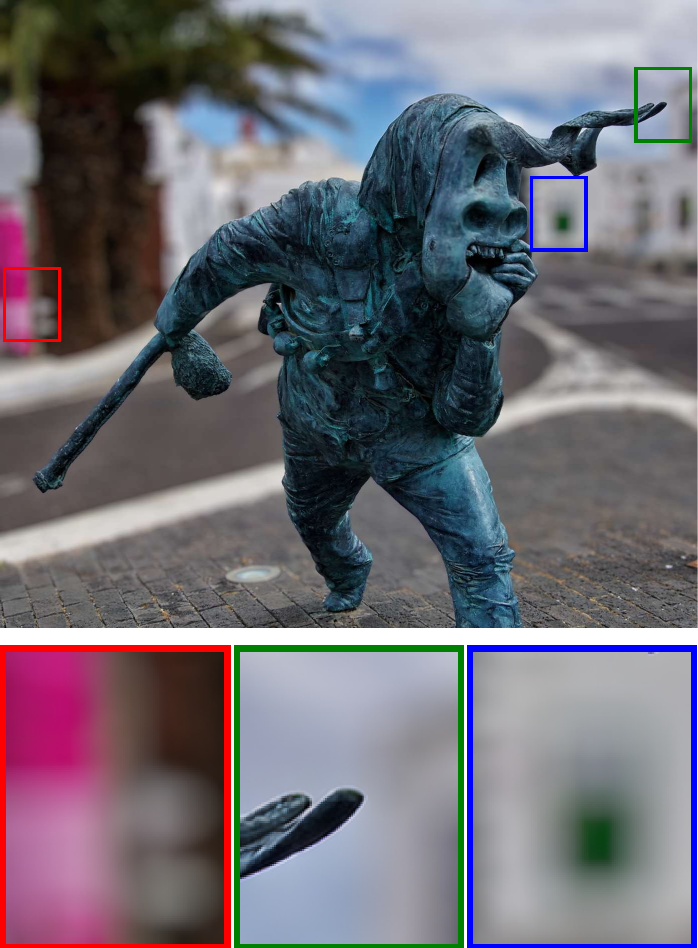}}\hfill\\
\caption{Visual comparisons of the ablation study. The setting of ``SoftmaxQ'',
``CoC'', and ``occlusion'' are short for the energy-conserved normalization, circle of confusion constraint, and self-occlusion respectively.
}\label{fig:ablation_vis}
\end{figure*}
We present some visual results to further support the ablation study in \cref{fig:ablation_vis}.

As the PISA module is designed to bring in constraints related to physics, an incomplete version cannot model the image formation model by design, and thus can only resort to learning from the data distribution. 

We first try removing the circle-of-confusion constraint, and the result in \cref{fig:ablation_vis}(b) looks very similar to the all-in-focus input, indicating that the PISA module determines how much the blurriness should be. 

In \cref{fig:ablation_vis}(c), removing the self-occlusion from the PISA module blurs the foreground that should be in focus, which is caused by the ignorance of keeping the foreground before the background, since self-occlusion is removed.

As for the energy-conserved normalization, since the energy no longer follows the physics intuition, the results look blurry even for the focused region in the green box of \cref{fig:ablation_vis}(d). 

\section{Future Works}
We conclude the paper with the some ideas for thought, hoping that BokehDiff inspire more interesting works.

The realm of lens blur rendering still lacks a metric to measure the ``photorealism''. With paired data, LPIPS is found to be the most sensitive to wrong blurring pattern~\cite{zhang2018unreasonable}. However, it is easy for human vision system to see the lens blur is synthesized, even without ground truth. The prior behind such phenomenon is intriguing, and requires further analysis. 

For example, will it be possible to train a discriminator that is able to focus on the low-level details that renders the image to be ``fake'' to human eyes, as an important reference-free metric?
If so, can we iterate a generator over the discriminator, to yield even more realistic images?

In addition, photorealistic video lens blur rendering is also an interesting follow-up thread, with its unique challenges such as consistency. With the proposed data synthesis pipeline, it will be easier to train a similar video bokeh rendering method, but this idea is beyond the scope of the paper, and deserves a paper of its own.

As for the model design, the PISA module requires more investigation. The biggest difference is that it changes the dimension on which to perform normalization. It is not self-evident to scale the default normalization to larger models (such as DiT~\cite{peebles2023scalable}).

{
    \small
    \bibliographystyle{ieeenat_fullname}
    \bibliography{main}

\begin{thebibliography}{66}
\providecommand{\natexlab}[1]{#1}
\providecommand{\url}[1]{\texttt{#1}}
\expandafter\ifx\csname urlstyle\endcsname\relax
  \providecommand{\doi}[1]{doi: #1}\else
  \providecommand{\doi}{doi: \begingroup \urlstyle{rm}\Url}\fi

\bibitem[Avrahami et~al.(2022)Avrahami, Lischinski, and Fried]{avrahami2022blended}
Omri Avrahami, Dani Lischinski, and Ohad Fried.
\newblock Blended diffusion for text-driven editing of natural images.
\newblock In \emph{Proc. of Computer Vision and Pattern Recognition}, 2022.

\bibitem[Avrahami et~al.(2023)Avrahami, Fried, and Lischinski]{avrahami2023blended}
Omri Avrahami, Ohad Fried, and Dani Lischinski.
\newblock Blended latent diffusion.
\newblock \emph{ACM Transactions on Graphics}, 42\penalty0 (4), 2023.

\bibitem[Bansal et~al.(2023)Bansal, Borgnia, Chu, Li, Kazemi, Huang, Goldblum, Geiping, and Goldstein]{bansal2024cold}
Arpit Bansal, Eitan Borgnia, Hong-Min Chu, Jie Li, Hamid Kazemi, Furong Huang, Micah Goldblum, Jonas Geiping, and Tom Goldstein.
\newblock Cold diffusion: Inverting arbitrary image transforms without noise.
\newblock In \emph{Proc. of Advances in Neural Information Processing Systems}, 2023.

\bibitem[{Black Forest Labs}(2024)]{flux}
{Black Forest Labs}.
\newblock Flux.1-schnell, 2024.
\newblock \url{https://huggingface.co/black-forest-labs/FLUX.1-schnell}, Last accessed on 2024-10-31.

\bibitem[Brooks et~al.(2023)Brooks, Holynski, and Efros]{brooks2023instructpix2pix}
Tim Brooks, Aleksander Holynski, and Alexei~A Efros.
\newblock Instructpix2pix: Learning to follow image editing instructions.
\newblock In \emph{Proc. of Computer Vision and Pattern Recognition}, 2023.

\bibitem[Busam et~al.(2019)Busam, Hog, McDonagh, and Slabaugh]{busam2019sterefo}
Benjamin Busam, Matthieu Hog, Steven McDonagh, and Gregory Slabaugh.
\newblock {SteReFo}: Efficient image refocusing with stereo vision.
\newblock In \emph{Proc. of International Conference on Computer Vision Workshops}, 2019.

\bibitem[Cao et~al.(2023)Cao, Wang, Qi, Shan, Qie, and Zheng]{cao2023masactrl}
Mingdeng Cao, Xintao Wang, Zhongang Qi, Ying Shan, Xiaohu Qie, and Yinqiang Zheng.
\newblock {MasaCtrl}: Tuning-free mutual self-attention control for consistent image synthesis and editing.
\newblock In \emph{Proc. of International Conference on Computer Vision}, 2023.

\bibitem[Chen et~al.(2023{\natexlab{a}})Chen, Zhang, Yang, and Liu]{chen2023multi}
Gang Chen, Guipeng Zhang, Zhenguo Yang, and Wenyin Liu.
\newblock Multi-scale patch-gan with edge detection for image inpainting.
\newblock \emph{Applied Intelligence}, 53\penalty0 (4), 2023{\natexlab{a}}.

\bibitem[Chen et~al.(2023{\natexlab{b}})Chen, Wang, Zhou, Qiao, and Dong]{chen2023activating}
Xiangyu Chen, Xintao Wang, Jiantao Zhou, Yu Qiao, and Chao Dong.
\newblock Activating more pixels in image super-resolution transformer.
\newblock In \emph{Proc. of Computer Vision and Pattern Recognition}, 2023{\natexlab{b}}.

\bibitem[Chesser et~al.(2020)Chesser, Carbone, and Zahid]{unsplash}
Luke Chesser, Timothy Carbone, and Ali Zahid.
\newblock Unsplash full dataset 1.2.2, 2020.
\newblock \url{unsplash.com/data}, Last accessed on 2024-10-31.

\bibitem[Conde et~al.(2023)Conde, Kolmet, Seizinger, Bishop, Timofte, Kong, Zhang, Wu, Wang, Peng, Pan, Liu, Luo, Sun, Shen, Cao, Xian, Liu, Chen, Yang, Liu, Jing, Mi, Wang, Yang, Lian, Lai, Zhang, Hoang, Yazdani, Monga, Luo, Gustafsson, Zhao, Sj\"olund, Sch\"on, Zhao, Chen, Xu, and Niu]{NTIRE_2023_CVPR}
Marcos~V. Conde, Manuel Kolmet, Tim Seizinger, Tom~E. Bishop, Radu Timofte, Xiangyu Kong, Dafeng Zhang, Jinlong Wu, Fan Wang, Juewen Peng, Zhiyu Pan, Chengxin Liu, Xianrui Luo, Huiqiang Sun, Liao Shen, Zhiguo Cao, Ke Xian, Chaowei Liu, Zigeng Chen, Xingyi Yang, Songhua Liu, Yongcheng Jing, Michael~Bi Mi, Xinchao Wang, Zhihao Yang, Wenyi Lian, Siyuan Lai, Haichuan Zhang, Trung Hoang, Amirsaeed Yazdani, Vishal Monga, Ziwei Luo, Fredrik~K. Gustafsson, Zheng Zhao, Jens Sj\"olund, Thomas~B. Sch\"on, Yuxuan Zhao, Baoliang Chen, Yiqing Xu, and JiXiang Niu.
\newblock Lens-to-lens bokeh effect transformation. ntire 2023 challenge report.
\newblock In \emph{Proc. of Computer Vision and Pattern Recognition Workshops}, 2023.

\bibitem[Ding et~al.(2020)Ding, Ma, Wang, and Simoncelli]{ding2020image}
Keyan Ding, Kede Ma, Shiqi Wang, and Eero~P Simoncelli.
\newblock Image quality assessment: Unifying structure and texture similarity.
\newblock \emph{IEEE Transactions on Pattern Analysis and Machine Intelligence}, 44\penalty0 (5), 2020.

\bibitem[Dutta et~al.(2021)Dutta, Das, Shah, and Tiwari]{dutta2021stacked}
Saikat Dutta, Sourya~Dipta Das, Nisarg~A Shah, and Anil~Kumar Tiwari.
\newblock Stacked deep multi-scale hierarchical network for fast bokeh effect rendering from a single image.
\newblock In \emph{Proc. of Computer Vision and Pattern Recognition Workshops}, 2021.

\bibitem[Epstein et~al.(2023)Epstein, Jabri, Poole, Efros, and Holynski]{epstein2023diffusion}
Dave Epstein, Allan Jabri, Ben Poole, Alexei Efros, and Aleksander Holynski.
\newblock Diffusion self-guidance for controllable image generation.
\newblock In \emph{Proc. of Advances in Neural Information Processing Systems}, 2023.

\bibitem[Hang et~al.(2023)Hang, Gu, Li, Bao, Chen, Hu, Geng, and Guo]{hang2023minsnr}
Tiankai Hang, Shuyang Gu, Chen Li, Jianmin Bao, Dong Chen, Han Hu, Xin Geng, and Baining Guo.
\newblock Efficient diffusion training via min-snr weighting strategy.
\newblock In \emph{Proc. of International Conference on Computer Vision}, 2023.

\bibitem[He et~al.(2024)He, Li, Yin, Liang, Li, Zhou, Liu, Liu, and Chen]{he2024lotus}
Jing He, Haodong Li, Wei Yin, Yixun Liang, Leheng Li, Kaiqiang Zhou, Hongbo Liu, Bingbing Liu, and Ying-Cong Chen.
\newblock Lotus: Diffusion-based visual foundation model for high-quality dense prediction.
\newblock \emph{arXiv preprint arXiv:2409.18124}, 2024.

\bibitem[Ho et~al.(2020)Ho, Jain, and Abbeel]{ho2020ddpm}
Jonathan Ho, Ajay Jain, and Pieter Abbeel.
\newblock Denoising diffusion probabilistic models.
\newblock In \emph{Proc. of Advances in Neural Information Processing Systems}, 2020.

\bibitem[Hu et~al.(2021)Hu, Shen, Wallis, Allen-Zhu, Li, Wang, Wang, and Chen]{hu2021lora}
Edward~J Hu, Yelong Shen, Phillip Wallis, Zeyuan Allen-Zhu, Yuanzhi Li, Shean Wang, Lu Wang, and Weizhu Chen.
\newblock {LoRA}: Low-rank adaptation of large language models.
\newblock \emph{arXiv preprint arXiv:2106.09685}, 2021.

\bibitem[Hussain et~al.(2022)Hussain, Zaki, and Subramanian]{hussain2022global}
Md~Shamim Hussain, Mohammed~J Zaki, and Dharmashankar Subramanian.
\newblock Global self-attention as a replacement for graph convolution.
\newblock In \emph{Conference on Knowledge Discovery and Data Mining}, 2022.

\bibitem[Ignatov et~al.(2020{\natexlab{a}})Ignatov, Patel, and Timofte]{ignatov2020rendering}
Andrey Ignatov, Jagruti Patel, and Radu Timofte.
\newblock Rendering natural camera bokeh effect with deep learning.
\newblock In \emph{Proc. of Computer Vision and Pattern Recognition Workshops}, 2020{\natexlab{a}}.

\bibitem[Ignatov et~al.(2020{\natexlab{b}})Ignatov, Timofte, Qian, Qiao, Lin, Guo, Li, Leng, Cheng, Peng, et~al.]{ignatov2020aim}
Andrey Ignatov, Radu Timofte, Ming Qian, Congyu Qiao, Jiamin Lin, Zhenyu Guo, Chenghua Li, Cong Leng, Jian Cheng, Juewen Peng, et~al.
\newblock Aim 2020 challenge on rendering realistic bokeh.
\newblock In \emph{Proc. of European Conference on Computer Vision Workshops}, 2020{\natexlab{b}}.

\bibitem[Jiang et~al.(2024)Jiang, Mao, Pan, Han, and Zhang]{jiang2024scedit}
Zeyinzi Jiang, Chaojie Mao, Yulin Pan, Zhen Han, and Jingfeng Zhang.
\newblock {SCEdit}: Efficient and controllable image diffusion generation via skip connection editing.
\newblock In \emph{Proc. of Computer Vision and Pattern Recognition}, 2024.

\bibitem[Kapitanov et~al.(2023)Kapitanov, Kvanchiani, and Sofia]{EasyPortrait}
Alexander Kapitanov, Karina Kvanchiani, and Kirillova Sofia.
\newblock {EasyPortrait} - face parsing and portrait segmentation dataset.
\newblock \emph{arXiv preprint arXiv:2304.13509}, 2023.

\bibitem[Kraus and Strengert(2007)]{kraus2007depth}
Martin Kraus and Magnus Strengert.
\newblock Depth-of-field rendering by pyramidal image processing.
\newblock In \emph{Computer Graphics Forum}, 2007.

\bibitem[Lee et~al.(2008)Lee, Kim, and Choi]{lee2008real}
Sungkil Lee, Gerard~Jounghyun Kim, and Seungmoon Choi.
\newblock Real-time depth-of-field rendering using point splatting on per-pixel layers.
\newblock In \emph{Computer Graphics Forum}, 2008.

\bibitem[Lee et~al.(2010)Lee, Eisemann, and Seidel]{lee2010real}
Sungkil Lee, Elmar Eisemann, and Hans-Peter Seidel.
\newblock Real-time lens blur effects and focus control.
\newblock \emph{ACM Transactions on Graphics}, 29\penalty0 (4), 2010.

\bibitem[Li et~al.(2022)Li, Lin, Zhou, Qi, Wang, and Jia]{li2022mat}
Wenbo Li, Zhe Lin, Kun Zhou, Lu Qi, Yi Wang, and Jiaya Jia.
\newblock {MAT}: Mask-aware transformer for large hole image inpainting.
\newblock In \emph{Proc. of Computer Vision and Pattern Recognition}, 2022.

\bibitem[Lijun et~al.(2018)Lijun, Xiaohui, Jianming, Oliver, Zhe, Chih-Yao, Sarah, and Huchuan]{deeplens2018}
Wang Lijun, Shen Xiaohui, Zhang Jianming, Wang Oliver, Lin Zhe, Hsieh Chih-Yao, Kong Sarah, and Lu Huchuan.
\newblock {DeepLens}: Shallow depth of field from a single image.
\newblock \emph{ACM Transactions on Graphics}, 37\penalty0 (6), 2018.

\bibitem[Lin et~al.(2024)Lin, Mo, Klingher, Mu, and Zhou]{lin2024ctrl}
Kuan~Heng Lin, Sicheng Mo, Ben Klingher, Fangzhou Mu, and Bolei Zhou.
\newblock {Ctrl-X}: Controlling structure and appearance for text-to-image generation without guidance.
\newblock \emph{arXiv preprint arXiv:2406.07540}, 2024.

\bibitem[Liu et~al.(2022)Liu, Mao, Wu, Feichtenhofer, Darrell, and Xie]{convnext}
Zhuang Liu, Hanzi Mao, Chao-Yuan Wu, Christoph Feichtenhofer, Trevor Darrell, and Saining Xie.
\newblock A {ConvNet} for the 2020s.
\newblock In \emph{Proc. of Computer Vision and Pattern Recognition}, 2022.

\bibitem[Loshchilov and Hutter(2019)]{adamw}
Ilya Loshchilov and Frank Hutter.
\newblock Decoupled weight decay regularization.
\newblock In \emph{Proc. of International Conference on Learning Representations}, 2019.

\bibitem[Lu et~al.(2022)Lu, Li, Liu, Huang, Zhang, and Zeng]{lu2022transformer}
Zhisheng Lu, Juncheng Li, Hong Liu, Chaoyan Huang, Linlin Zhang, and Tieyong Zeng.
\newblock Transformer for single image super-resolution.
\newblock In \emph{Proc. of Computer Vision and Pattern Recognition}, 2022.

\bibitem[Mandl et~al.(2024)Mandl, Mori, Mohr, Peng, Langlotz, Schmalstieg, and Kalkofen]{Mandl2024NeuralBokeh}
David Mandl, Shohei Mori, Peter Mohr, Yifan Peng, Tobias Langlotz, Dieter Schmalstieg, and Denis Kalkofen.
\newblock Neural bokeh: Learning lens blur for computational videography and out-of-focus mixed reality.
\newblock In \emph{IEEE Conference on Virtual Reality and 3D User Interfaces}, 2024.

\bibitem[Meng et~al.(2021)Meng, He, Song, Song, Wu, Zhu, and Ermon]{meng2021sdedit}
Chenlin Meng, Yutong He, Yang Song, Jiaming Song, Jiajun Wu, Jun-Yan Zhu, and Stefano Ermon.
\newblock {SDEdit}: Guided image synthesis and editing with stochastic differential equations.
\newblock \emph{arXiv preprint arXiv:2108.01073}, 2021.

\bibitem[Nazeri et~al.(2019)Nazeri, Ng, Joseph, Qureshi, and Ebrahimi]{nazeri2019edgeconnect}
Kamyar Nazeri, Eric Ng, Tony Joseph, Faisal Qureshi, and Mehran Ebrahimi.
\newblock Edgeconnect: Structure guided image inpainting using edge prediction.
\newblock In \emph{Proc. of International Conference on Computer Vision Workshops}, 2019.

\bibitem[Parmar et~al.(2024)Parmar, Park, Narasimhan, and Zhu]{parmar2024img2imgturbo}
Gaurav Parmar, Taesung Park, Srinivasa Narasimhan, and Jun-Yan Zhu.
\newblock One-step image translation with text-to-image models.
\newblock \emph{arXiv preprint arXiv:2403.12036}, 2024.

\bibitem[Peebles and Xie(2023)]{peebles2023scalable}
William Peebles and Saining Xie.
\newblock Scalable diffusion models with transformers.
\newblock In \emph{Proc. of International Conference on Computer Vision}, 2023.

\bibitem[Peng et~al.(2022{\natexlab{a}})Peng, Cao, Luo, Lu, Xian, and Zhang]{peng2022bokehme}
Juewen Peng, Zhiguo Cao, Xianrui Luo, Hao Lu, Ke Xian, and Jianming Zhang.
\newblock {BokehMe}: When neural rendering meets classical rendering.
\newblock In \emph{Proc. of Computer Vision and Pattern Recognition}, 2022{\natexlab{a}}.

\bibitem[Peng et~al.(2022{\natexlab{b}})Peng, Zhang, Luo, Lu, Xian, and Cao]{peng2022mpib}
Juewen Peng, Jianming Zhang, Xianrui Luo, Hao Lu, Ke Xian, and Zhiguo Cao.
\newblock {MPIB}: An {MPI}-based bokeh rendering framework for realistic partial occlusion effects.
\newblock In \emph{Proc. of European Conference on Computer Vision}, 2022{\natexlab{b}}.

\bibitem[Peng et~al.(2023)Peng, Pan, Liu, Luo, Sun, Shen, Xian, and Cao]{selective2023peng}
Juewen Peng, Zhiyu Pan, Chengxin Liu, Xianrui Luo, Huiqiang Sun, Liao Shen, Ke Xian, and Zhiguo Cao.
\newblock Selective bokeh effect transformation.
\newblock In \emph{Proc. of Computer Vision and Pattern Recognition Workshops}, 2023.

\bibitem[Peng et~al.(2024)Peng, Cao, Luo, Xian, Tang, Zhang, and Lin]{peng2024bokehmepp}
Juewen Peng, Zhiguo Cao, Xianrui Luo, Ke Xian, Wenfeng Tang, Jianming Zhang, and Guosheng Lin.
\newblock {BokehMe++}: Harmonious fusion of classical and neural rendering for versatile bokeh creation.
\newblock \emph{IEEE Transactions on Pattern Analysis and Machine Intelligence}, 2024.

\bibitem[Podell et~al.(2023)Podell, English, Lacey, Blattmann, Dockhorn, M{\"u}ller, Penna, and Rombach]{podell2023sdxl}
Dustin Podell, Zion English, Kyle Lacey, Andreas Blattmann, Tim Dockhorn, Jonas M{\"u}ller, Joe Penna, and Robin Rombach.
\newblock {SDXL}: Improving latent diffusion models for high-resolution image synthesis.
\newblock \emph{arXiv preprint arXiv:2307.01952}, 2023.

\bibitem[Potmesil and Chakravarty(1981)]{potmesil1981lens}
Michael Potmesil and Indranil Chakravarty.
\newblock A lens and aperture camera model for synthetic image generation.
\newblock \emph{Proc. of ACM SIGGRAPH}, 15\penalty0 (3), 1981.

\bibitem[Qian et~al.(2020)Qian, Qiao, Lin, Guo, Li, Leng, and Cheng]{qian2020bggan}
Ming Qian, Congyu Qiao, Jiamin Lin, Zhenyu Guo, Chenghua Li, Cong Leng, and Jian Cheng.
\newblock {BGGAN}: Bokeh-glass generative adversarial network for rendering realistic bokeh.
\newblock In \emph{Proc. of European Conference on Computer Vision Workshops}, 2020.

\bibitem[Seif and Androutsos(2018)]{seif2018edge}
George Seif and Dimitrios Androutsos.
\newblock Edge-based loss function for single image super-resolution.
\newblock In \emph{Proc. of International Conference on Acoustics, Speech, and Signal Processing}, 2018.

\bibitem[SG\_161222(2024)]{sg161222}
SG\_161222.
\newblock {RealVisXL} 5.0, 2024.
\newblock \url{https://huggingface.co/SG161222/RealVisXL_V5.0}, Last accessed on 2024-10-31.

\bibitem[Sheng et~al.(2024)Sheng, Yu, Ling, Cao, Zhang, Lu, Xian, Lin, and Benes]{sheng2024dr}
Yichen Sheng, Zixun Yu, Lu Ling, Zhiwen Cao, Xuaner Zhang, Xin Lu, Ke Xian, Haiting Lin, and Bedrich Benes.
\newblock {Dr. Bokeh}: Differentiable occlusion-aware bokeh rendering.
\newblock In \emph{Proc. of Computer Vision and Pattern Recognition}, 2024.

\bibitem[Song et~al.(2020)Song, Meng, and Ermon]{song2020ddim}
Jiaming Song, Chenlin Meng, and Stefano Ermon.
\newblock Denoising diffusion implicit models.
\newblock \emph{arXiv preprint arXiv:2010.02502}, 2020.

\bibitem[{Stability AI}(2024)]{sd3}
{Stability AI}.
\newblock Stable diffusion 3, 2024.
\newblock \url{https://huggingface.co/stabilityai/stable-diffusion-3-medium}, Last accessed on 2024-10-31.

\bibitem[Sun et~al.(2024)Sun, Wu, Zhang, Yong, and Zhang]{sun2401improving}
L Sun, R Wu, Z Zhang, H Yong, and L Zhang.
\newblock Improving the stability of diffusion models for content consistent super-resolution.
\newblock \emph{arXiv preprint arXiv:2401.00877}, 2024.

\bibitem[Vaidyanathan et~al.(2015)Vaidyanathan, Munkberg, Clarberg, and Salvi]{vaidyanathan2015layered}
Karthik Vaidyanathan, Jacob Munkberg, Petrik Clarberg, and Marco Salvi.
\newblock Layered light field reconstruction for defocus blur.
\newblock \emph{ACM Transactions on Graphics}, 34\penalty0 (2), 2015.

\bibitem[Wu et~al.(2024)Wu, Sun, Ma, and Zhang]{osediff}
Rongyuan Wu, Lingchen Sun, Zhiyuan Ma, and Lei Zhang.
\newblock One-step effective diffusion network for real-world image super-resolution.
\newblock \emph{arXiv preprint arXiv:2406.08177}, 2024.

\bibitem[Xiao et~al.(2018)Xiao, Kaplanyan, Fix, Chapman, and Lanman]{xiao2018deepfocus}
Lei Xiao, Anton Kaplanyan, Alexander Fix, Matthew Chapman, and Douglas Lanman.
\newblock Deepfocus: learned image synthesis for computational displays.
\newblock \emph{ACM Transactions on Graphics}, 37\penalty0 (6), 2018.

\bibitem[Xu et~al.(2024)Xu, Huang, Pan, Ma, and Chai]{xu2024inversion}
Sihan Xu, Yidong Huang, Jiayi Pan, Ziqiao Ma, and Joyce Chai.
\newblock Inversion-free image editing with language-guided diffusion models.
\newblock In \emph{Proc. of Computer Vision and Pattern Recognition}, 2024.

\bibitem[Yan et~al.(2015)Yan, Mehta, Ramamoorthi, and Durand]{yan2015fast}
Ling-Qi Yan, Soham~Uday Mehta, Ravi Ramamoorthi, and Fredo Durand.
\newblock Fast 4d sheared filtering for interactive rendering of distribution effects.
\newblock \emph{ACM Transactions on Graphics}, 35\penalty0 (1), 2015.

\bibitem[Yang et~al.(2024{\natexlab{a}})Yang, Kang, Huang, Xu, Feng, and Zhao]{depth_anything_v1}
Lihe Yang, Bingyi Kang, Zilong Huang, Xiaogang Xu, Jiashi Feng, and Hengshuang Zhao.
\newblock Depth anything: Unleashing the power of large-scale unlabeled data.
\newblock In \emph{Proc. of Computer Vision and Pattern Recognition}, 2024{\natexlab{a}}.

\bibitem[Yang et~al.(2024{\natexlab{b}})Yang, Kang, Huang, Xu, Feng, and Zhao]{yang2024depth}
Lihe Yang, Bingyi Kang, Zilong Huang, Xiaogang Xu, Jiashi Feng, and Hengshuang Zhao.
\newblock Depth anything: Unleashing the power of large-scale unlabeled data.
\newblock In \emph{Proc. of Computer Vision and Pattern Recognition}, 2024{\natexlab{b}}.

\bibitem[Yang et~al.(2024{\natexlab{c}})Yang, Kang, Huang, Zhao, Xu, Feng, and Zhao]{depth_anything_v2}
Lihe Yang, Bingyi Kang, Zilong Huang, Zhen Zhao, Xiaogang Xu, Jiashi Feng, and Hengshuang Zhao.
\newblock Depth anything v2.
\newblock \emph{arXiv:2406.09414}, 2024{\natexlab{c}}.

\bibitem[Yang et~al.(2023{\natexlab{a}})Yang, Wu, Ren, Xie, and Zhang]{yang2023pixel}
Tao Yang, Rongyuan Wu, Peiran Ren, Xuansong Xie, and Lei Zhang.
\newblock Pixel-aware stable diffusion for realistic image super-resolution and personalized stylization.
\newblock \emph{arXiv preprint arXiv:2308.14469}, 2023{\natexlab{a}}.

\bibitem[Yang et~al.(2023{\natexlab{b}})Yang, Lian, and Lai]{yang2023bokehornot}
Zhihao Yang, Wenyi Lian, and Siyuan Lai.
\newblock {BokehOrNot}: Transforming bokeh effect with image transformer and lens metadata embedding.
\newblock In \emph{Proc. of Computer Vision and Pattern Recognition}, 2023{\natexlab{b}}.

\bibitem[Yuan et~al.(2025)Yuan, Wang, Sheng, Chennuri, Zhang, and Chan]{yuan2025generative}
Yu Yuan, Xijun Wang, Yichen Sheng, Prateek Chennuri, Xingguang Zhang, and Stanley Chan.
\newblock Generative photography: Scene-consistent camera control for realistic text-to-image synthesis.
\newblock In \emph{Proc. of Computer Vision and Pattern Recognition}, 2025.

\bibitem[Zamir et~al.(2022)Zamir, Arora, Khan, Hayat, Khan, and Yang]{zamir2022restormer}
Syed~Waqas Zamir, Aditya Arora, Salman Khan, Munawar Hayat, Fahad~Shahbaz Khan, and Ming-Hsuan Yang.
\newblock Restormer: Efficient transformer for high-resolution image restoration.
\newblock In \emph{Proc. of Computer Vision and Pattern Recognition}, 2022.

\bibitem[Zhang and Agrawala(2024)]{zhang2024transparent}
Lvmin Zhang and Maneesh Agrawala.
\newblock Transparent image layer diffusion using latent transparency.
\newblock \emph{arXiv preprint arXiv:2402.17113}, 2024.

\bibitem[Zhang et~al.(2023)Zhang, Rao, and Agrawala]{zhang2023adding}
Lvmin Zhang, Anyi Rao, and Maneesh Agrawala.
\newblock Adding conditional control to text-to-image diffusion models.
\newblock In \emph{Proc. of International Conference on Computer Vision}, 2023.

\bibitem[Zhang et~al.(2018)Zhang, Isola, Efros, Shechtman, and Wang]{zhang2018unreasonable}
Richard Zhang, Phillip Isola, Alexei~A Efros, Eli Shechtman, and Oliver Wang.
\newblock The unreasonable effectiveness of deep features as a perceptual metric.
\newblock In \emph{Proc. of Computer Vision and Pattern Recognition}, 2018.

\bibitem[Zhang et~al.(2019)Zhang, Matzen, Nguyen, Yao, Zhang, and Ng]{zhang2019synthetic}
Xuaner Zhang, Kevin Matzen, Vivien Nguyen, Dillon Yao, You Zhang, and Ren Ng.
\newblock Synthetic defocus and look-ahead autofocus for casual videography.
\newblock \emph{ACM Transactions on Graphics}, 38\penalty0 (4), 2019.

\end{thebibliography}
}
\end{document}